\documentclass[floatfix,superscriptaddress,groupedaddress,preprint,amsmath,amssymb,nofootinbib]{revtex4-1}
\pdfoutput=1
\usepackage[utf8]{inputenc} 
\usepackage{graphicx}
\usepackage{subfigure}
\usepackage{dcolumn}
\usepackage{bm}
\usepackage{hyperref}
\usepackage{epstopdf} 
\usepackage[export]{adjustbox}
\usepackage[table,x11names]{xcolor}
\usepackage[numbered,framed]{mcode}

\begin{document}

\title{Sparse model selection in the highly under-sampled regime}

\author{Nicola Bulso}
\email{nicola.bulso@ntnu.no}
\affiliation{The Kavli Institute and Centre for Neural Computation, Trondheim, Norway}
\affiliation{The Abdus Salam International Centre for Theoretical Physics (ICTP), Trieste, Italy}

\author{Matteo Marsili}
\email{marsili@ictp.it}
\affiliation{The Abdus Salam International Centre for Theoretical Physics (ICTP), Trieste, Italy}

\author{Yasser Roudi}
\email{yasser.roudi@ntnu.no}
\affiliation{The Kavli Institute and Centre for Neural Computation, Trondheim, Norway}
\affiliation{Institute for Advanced Study, Princeton, USA}
\affiliation{The Abdus Salam International Centre for Theoretical Physics (ICTP), Trieste, Italy}

\begin{abstract}
We propose a method for recovering the structure of a sparse undirected graphical model when very few samples are available. The method decides about the presence or absence of bonds between pairs of variable by considering one pair at a time and using a closed form formula, analytically derived by calculating the posterior probability for every possible model explaining a two body system using Jeffreys prior. The approach does not rely on the optimization of any cost functions and consequently is much faster than existing algorithms. Despite this time and computational advantage, numerical results show that for several sparse topologies the algorithm is comparable to the best existing algorithms, and is more accurate in the presence of hidden variables. We apply this approach to the analysis of US stock market data and to neural data, in order to show its efficiency in recovering robust statistical dependencies in real data with non-stationary correlations in time and/or space.
\end{abstract}

\pacs{}
\keywords{}
                             
\maketitle

\section{\label{sec: introduction}Introduction}

Natural images and sounds, along with many other signals in the world admit sparse representations: models with only a small number of non-zero parameters suffice to uniquely identify the signal. Similarly, many real networks, such as neuronal and gene regulatory networks are sparse that is each node connects to a relatively small fraction of all possible nodes. 
In the typical situation, in these cases, the dimensionality of the data exceeds by far the number of available observations, and even when many samples are available one is hardly ever in the situation where one can regard them as being drawn independently from the same distribution. 
In addition, typically only a fraction of the relevant variables are sampled and the observed behaviour is potentially influenced by an unknown number of {\em hidden} variables. 
In these cases, typically, any inferred model is far from being an accurate description of the data generating process. Models that are very complex afford a predictive power that exceeds what can be validated statistically, whereas simpler (sparse) models are more likely to perform well out of sample. 
Therefore sparse model selection is an important tool for both sparse recovery, when the specific instance or phenomenon of the world we are investigating admit a sparse representation, and statistical modelling in the under-sampled regime, where learning more complex models instead would inevitably lead to overfitting issues.

Understanding the theoretical foundations of sparse signals and network recovery from a small set of high-dimensional observations has thus become an important area of research in the past few years. This theory is not only of important practical use, but it also is likely to shine light on fundamental aspects of biological information processing. Biological systems in their environment usually face the same problem that scientists do when trying to understand a complex system: identifying signals and relationships between them from limited noisy measurements and as fast as possible. 

In recent years, approaches based on optimizing cost functions that favor sparse representations have been shown to be very successful for sparse recovery. From earlier work on Lasso \cite{Lasso}  to more recent ones on compressed sensing \cite{compressed-sensing}, a large body of research has shown the asymptotic power of the sparsity prior in the exact or almost exact recovery of signals from a small sample of observations. 

In network reconstruction, the standard platform for studying network recovery is that of an Ising model, or a pairwise Markov Random Field as is known in the machine Learning literature. This takes the form of a distribution
\begin{equation}
P(\vec S|J,h)=\frac{\exp\left\{  \sum_{i<j} J_{ij} S_i S_j+\sum_i h_i S_i \right\}}{Z}
\label{eq:ising}
\end{equation}
\noindent 
over $n$ binary spin variables $S_i= \pm 1$, where the partition function $Z$ ensures normalisation. Given a sample $\hat S$ of $N$ observations of the vector $\vec S=(S_1,\ldots, S_n)$ of spin variables, the general problem is that of finding the interactions $J_{ij}$ and fields $h_i$ that best describe the data. In the case where the data is generated as independent random draws from $P(S|J,h)$ in Eq. \ref{eq:ising}, with unknown $J_{ij}$ and $h_i$, the problem is that of recovering these parameters. In the sparse regime, only a small number of the couplings are non-zero, so the problem amounts to recovering the network of interactions and then estimating the values of the non-zero parameters. When $\hat S$ comes from a real experiment or from observation of a complex system, Eq. \ref{eq:ising} becomes a tool that can be used to infer a putative network of statistical dependencies among the variables $S_i$.

In the general setting where sparsity is not necessarily imposed,  starting from the work on ``Boltzmann machine learning" \cite{ackley85}, many approximations for inferring the model parameters have been proposed and studied \cite{Yasser09,Cocco11,Federico12,Aurell12}.

Similarly, structure estimation has been the subject of many studies over the last decades, especially in the machine learning and statistics community. Classical approaches based on penalising cost functions by their complexity \cite{Akaike,BIC} or on constraining the search in the model space \cite{Chow&Liu1968} have been exploited and extended over the years making the way for new algorithms and ideas. In general, two main classes of algorithms can be distinguished: constrained based \cite{Tsamardinos2003,Abbeel2006,Bromberg2006,Anandkumar12} and optimisation based algorithms \cite{Lee2006,Hofling09,ravikuumar10,Schmidt2010,Lowd2014,Pensar2014,Barber2015}. Algorithms in the first class apt to recover the structure of the network employing a series of local independence tests which are appropriately combined together to yield an estimate for the structure. On the other hand, optimisation based approaches rest on the definition of a penalised cost function which integrates a more global information of the status of the network and its optimisation over the models space. The size of that space grows exponentially with at least the square of the network size making hard the optimisation already for small sizes. Therefore, beyond the issues that eventually the evaluation of ``global" cost function might arise, it is crucial devising methods for performing efficiently such an optimisation in the model space. An elegant and popular solution converts the optimisation over the space of models into an optimisation over the space of parameters typically by minimising an $\ell_1$ regularised function of the parameters of the model \cite{Lee2006,Hofling09,ravikuumar10}. In this direction an important contribution was made by Ravikumar et al \cite{ravikuumar10}, who showed that in the high-dimensional regime, defined as the asymptotic regime where the size of the system is larger than the number of samples (while both go to infinity), exact recovery of the set of non-zero bonds is possible via the optimization of the $\ell_1$ regularized pseudo-likelihood function. In the following we denote this method by PLM+$\ell_1$. Although asymptotic high-dimensional analysis ensures the reconstruction of the connectivity pattern by an $\ell_1$ regularized logistic regression, a more recent approach \cite{Decelle13} was shown to outperform the $\ell_1$ regularized approach in certain regimes with finite datasets. This algorithm by Decelle and Ricci-Tersenghi \cite{Decelle13} does not require the selection of a regularization parameter. Typically, the regularization parameter can be chosen by cross-validation, but this is not a straightforward task when very few data points are available. This recipe has two main ingredients: the pseudo-likelihood function \cite{Besag1972} and the decimation algorithm which enables a fast and efficient walk in the space of models.

All these methods mainly concentrate on an high-dimensional setting where the number of samples can be smaller than the size of the network while both growing large. Conversely, we focus our analysis on a more noisy regime driven by the fact that in many real world applications the number of samples which can be approximately considered as independently drawn by the same distribution, might be very small and the presence of hidden variables makes the data even more noisy. In what follows we present a simple local algorithm built on the idea that only very sparse representations can be statistically validate in this regime. The simplicity of the algorithm allows it to be used as an on-line tool which leads to a huge computational advantage and makes the algorithm appealing especially for large-scale applications.

\subsection*{Overview of the method and results}

In this work we study the problem of sparse network recovery from a Bayesian model selection perspective. In brief, network recovery amounts to finding which model best describes the data. Yet the number of possible pairwise graphical models grows exponentially with the (square of the) network size $n$, which makes Bayesian model selection unfeasible in practice. 
However, Bayesian model selection penalises models for their complexity, which grows with the number of parameters they depend on. This penalty is particularly severe when the number $N$ of samples is not very large (compared to the dimensionality) -- which we will call the under-sampling regime in what follows. In this regime, sparse networks are the most likely representation because they are simpler. Indeed, in the deep under-sampling regime, the most likely models are those where the network of interactions is so sparse that spins hardly interact. In other words, only sparse representations of the world can be learnt because only those representations can be validated statistically. Therefore, in this regime, it makes sense to consider each pair of spins as if they were independent of all other spins, and to decide about the presence or absence of a bond between them independently from all others. This reduces the problem to $n(n-1)/2$ simpler problems of Bayesian model selection for systems of two spins, that we treat in full detail. 

Another main ingredient of our recipe concerns the choice of the prior over parameters. In the under-sampling regime, this choice becomes relevant and that is why we use Jeffreys prior \cite{Jeffreys46}. Among the many interesting properties of such an uninformative prior, it is worth remembering its reparametrization invariance: it makes the Bayesian posterior invariant under reparametrization leading to predictions which are independent of the subjective choice of the parametrization made by the observer. Furthermore, another interesting fact comes from its geometrical interpretation. It has been shown \cite{Myung00, Balasubramanian1996} that such a prior assigns a weight to a given set of parameters according to how many ``distinguishable" distributions are indexed by them. This means that all distinguishable distributions which are embedded in the manifold spanned by the parametric family in the space of distributions, are assigned the same prior probability, e.g. a uniform prior in the space of the parametric family distributions. Finally, for exponential families, Bayesian predictions with Jeffreys prior are closely related to the Minimum Description Length principle in its refined form, giving to Jeffreys prior an universal coding interpretation \cite{Rissanen1984,Rissanen1996, Grunwald}.

As a result of the above discussion and of the following analysis, we observe that if the recovered network contains no loops, we expect this to be an accurate reconstruction whereas if it contains loops or cliques, it is expected to over-estimate the number of bonds. In the weakly interacting regime and for sparse topologies admitting few loops, we find that the method is able to classify relevant features predicting the probability of having a direct connection between a pair of nodes in a network. In the under-sampling regime this task is performed quite well and with the same accuracy of more complex optimisation based methods. This makes it possible to save computational time, thus facilitating large scale network applications. Moreover, our method has been found particularly successful in revealing the absence of a direct connection between two spins, becoming more and more accurate as the number of samples grows.
Therefore, our method can also be used as a pre-processing step to prune the set of possible interactions, before applying more complex inference methods. These results are supported by simulations and discussed in detail in Sec.\ \ref{sec: results}. In particular in this section we contrast MS with PLM+$\ell_1$ on synthetic data from systems with known topology. Our numerical results demonstrate that, when all the variables on the nodes of the graph are observed PLM+$\ell_1$ offers a superior performance, though the results of our method still afford a good performance and are computationally advantageous. For data on partially observed graphs, where the data only accounts for a fraction of the variables while the remaining ones act as hidden nodes (unknown unknowns), the procedure proposed here is both computationally advantageous and it has superior performance, specially in the limit of few samples.

On the computational side, our approach is remarkably fast, as it entails solving a two spin problem and building a look-up table, using which the decision of which bond is present or not can be taken. It is worth noticing that the decision table needs to be built only once for a given sample length and then the resulting discriminator can be applied as an on-line tool to any couple of nodes in any dataset of the same length.

This approach is particularly suited to study real data where the number of available samples is necessarily limited. For example, in the study of non-stationary data or in the presence of an external relevant variable. In these cases, taking a too large time window or neglecting the effect of the external variable induces correlations that result in effective interactions that do not reflect genuine statistical dependence. This effect can be controlled by performing inference on small time windows or by partitioning the data conditioning on the external variable. This leads to inference problems where the number of samples is very small, for which our method is ideally suited. In this situation, a model selection approach is necessary in order to correctly evince how much structure can be inferred from the data. Hence the MS approach is a valuable alternative to other approaches \cite{Ben13,Claudia15} to inference in the presence of hidden variables. We illustrate these points for the specific cases of a dataset of daily returns of $n=41$ stocks from US stocks market and a dataset on the neural activity of cells in the entorhinal cortex of a moving rat (the same data studied in Refs. \cite{ponsot} and \cite{Maria15} respectively). At odds with other methods that need longer samples, our approach is able to spot the non-stationary nature of the financial data by focusing on small time windows, and in eliminating spurious correlations arising from co-variation in the neural data by focusing on small cells. 
 

The paper is organized as follows. In the first section, we describe the model selection approach for topology reconstruction in a network of interacting variables. Then we focus on two-spins clusters, deriving the key equations of Bayesian model selection which lie at the heart of our method. The third section is aimed at exhibiting results with synthetic data drawn from an equilibrium Ising distribution and comparing them with those obtained with a pseudo-likelihood method. Finally we discuss the application of our method to real data.

\section{\label{sec: method}A model selection approach to topology reconstruction}
Suppose we have $N$ observations $\hat{S} = (\vec{S}^1,\vec{S}^2,...,\vec{S}^N)$ of a vector $\vec{S}= (S_1,S_2,...,S_n)$ of $n$ variables $S_i$. We think of $\vec S$ as a configuration of a graphical model, i.e. a model of $n$  variables whose interaction is defined by a network between the $n$ variable nodes \cite{BookMezardMontanari}. Let us consider a collection of different mathematical models, ${\cal M}_i$ each with a possibly different set of connections. Each ${\cal M}_i$ is a possible hypothesis to explain the data and the probability $P({\cal M}_i|\hat{S})$ of ${\cal M}_i$ given the data, is related to the likelihood $P(\hat{S}|{\cal M}_i)$ that the model ${\cal M}_i$ has generated the data $\hat{S}$, through Bayes theorem: 
\begin{equation}\label{eqn:Bayes}
P({\cal M}_i|\hat{S}) = \frac{P(\hat{S}|{\cal M}_i) P_0({\cal M}_i)}{\sum_j P(\hat{S}|{\cal M}_j) P_0({\cal M}_j)}.
\end{equation}
Here $ P_0({\cal M}_i)$ is the prior probability of model ${\cal M}_i$. 
Each model employs a different vector of parameters $\vec{\theta}$ whose length and properties depend on the particular model ${\cal M}_i$. The {\em a priori} knowledge on the value of $\vec \theta$ is encoded in the prior distribution $P(\vec{\theta}| {\cal M}_i)$. Hence, the likelihood $P(\hat{S}|{\cal M}_i)$ can be written as
\begin{equation}\label{eqn:1}
P(\hat{S}| {\cal M}_i) = \int d\vec{\theta} P(\hat{S}|\vec{\theta}, {\cal M}_i) P(\vec{\theta}| {\cal M}_i)
\end{equation}
where $P(\hat{S}|\vec{\theta}, {\cal M}_i)$  is the conditional probability of observing the data $\hat{S}$ given a particular choice of the parameters $\vec{\theta}$ of model ${\cal M}_i$.

We focus on systems of binary variables, or Ising spins, where each variable takes values $S_i=\pm 1$, and on models in the exponential family, i.e. 
\begin{equation}\label{eqn:equi_Ising}
P(\vec{S}|\vec{\theta}, {\cal M}_i) = \frac{e^{\sum_{j=1}^\Theta f_j(\vec{S})\theta_j}}{Z(\vec{\theta})},
\end{equation}
where the number of parameters $\Theta$ depends on the particular model ${\cal M}_i$, $f_j(\vec{S})$ are coefficients multiplying the parameters and depending on spins $\vec{S}$ and, finally, $Z(\vec{\theta})$ is the partition function. Given this and a set of independent observations $\hat{S} = (\vec{S}^1,\vec{S}^2,...,\vec{S}^N)$, the probability of observing the collected data given a particular choice of the parameters within model ${\cal M}_i$  is 
 $P(\hat{S}|\vec{\theta}, {\cal M}_i) = \prod_\mu P(\vec{S}^\mu|\vec{\theta}, {\cal M}_i) $  with $\mu = 1,2,...,N$ or

\begin{equation}\label{eqn:equi_Ising2}
 P(\hat{S}|\vec{\theta} ,{\cal M}_i) = \frac{e^{N \vec{\phi}\cdot\vec{\theta}}}{Z^N(\vec{\theta})}
\end{equation} 
where 
\begin{equation}
\phi_j = \frac{1}{N}\sum_\mu f_j(\vec{S}^\mu).
\end{equation}
and $\vec{\phi}\cdot\vec{\theta} = \sum_j \phi_j \theta_j$
The probability of observing the data under model ${\cal M}_i$ can then be written as
\begin{equation}\label{eqn:finalGeneralEq}
P(\hat{S}| {\cal M}_i) = \int d\vec{\theta}\ e^{N \vec{\phi}\cdot\vec{\theta}-N \log Z(\vec{\theta})} P(\vec{\theta}| {\cal M}_i).
\end{equation}

Our problem is to choose from these models by calculating $P({\cal M}_i|\hat{S})$ through Eq.\ \ref{eqn:Bayes}. The prior on models $P_0({\cal M}_i)$ can account for any prior belief regarding the structure of the network, for instance the degree of sparsity.

For large $N$, the integral in Eq.\ \ref{eqn:finalGeneralEq} is dominated by the maximum of $\vec{\phi}\cdot\vec{\theta}-\log Z(\vec{\theta})$ and it can be evaluated by the saddle point method (see e.g. \cite{Myung00}). This produces a leading term in $\log P(\hat{S}| {\cal M}_i)$ that is proportional to $N$, which is given by the likelihood evaluated at the saddle point $\vec\theta^*$. Besides this, a term proportional to $\frac{\Theta}{2}\log N$ also arises from the Gaussian integration over the $\Theta$ parameters. This term, which penalises models with many ($\Theta\gg 1$) parameters, is the basis of the Bayesian Information Criterium (BIC) \cite{BIC}. Finally, further constant terms, that depend on the choice of the prior $P(\vec \theta|{\cal M}_i)$ also appear. In the under-sampling regime, when $N$ is not very large, all these terms become important, and as $N$ increases one expects the most likely model to become more and more complex. 

A direct and exact application of the above scheme for deriving the network would require ranking the evidence $P({\cal M}_i|\hat{S})$ for all possible network models ${\cal M}_i$, that number to at least $2^{n(n+1)/2}$ possible models.  Apart from the complications related to calculating the partition function for a single evaluation of Eq.\ \ref{eqn:finalEq}, this exponentially large number of graphs makes it impossible to work out the above approach based on the model posterior. 

Yet, for many real life applications, and in the under-sampling regime, we expect the most likely models to correspond to sparse graphs composed of many disconnected components.
In this case, $P(\vec S|\vec \theta,{\cal M}_i)$ becomes a product of factors corresponding to each component, which are easier to handle. In the extreme limit where components are formed of isolated spins or of dimers of interacting spins, calculations can be done by considering minimal clusters of only two spins which are taken to interact with the rest of the network through some effective fields. As we will show, this approach proves to be effective when there are few data points, performing as well as $\ell_1$ regularized pseudo-likelihood methods. 
Given the simplicity of the minimal cluster of two spins, we are able to perform all calculations analytically and rank the models. Last but not least an important advantage of such an approach is that it can be parallelized up to $n(n-1)/2$ times since all pairs are assumed to be independent of each other and can be analysed in parallel.

\subsection{\label{sec: 2spins} The minimal cluster}

\begin{figure}[t!]
\centering
\includegraphics[width = 0.6\columnwidth]{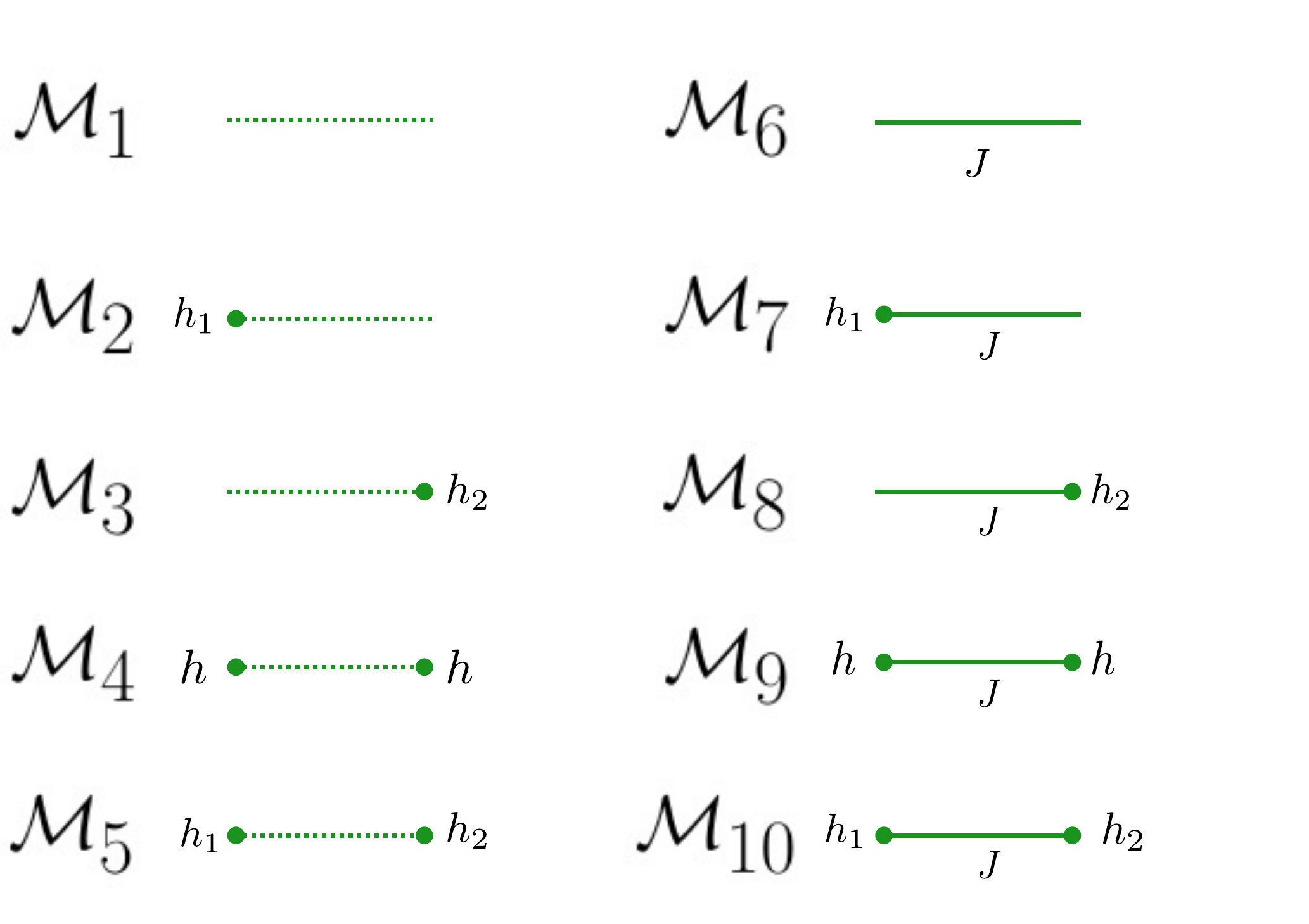}
\caption{The ten competiting models for a two spins system. The simplest one (top left of the figure) has no parameters: the two spins are independent of each other (dotted line) and from the rest of the network; the most complex model (bottom right) is the one with 3 parameters: the two spins are linked by a direct connection $J$ (solid line) and influenced by effective fields $h_1$ and $h_2$ (spots on the edges) representing external biases and other spins' influence. Models on the left do not employ a direct connection between spins whereas the other five models on the right do. The parameter $h$ is used to depict situations in which both spins are conditioned by the same bias.}
\label{fig:models}
\end{figure}

A two spin system has at most three parameters: two fields and one coupling. The number of possible models for two spins is thus 10, as shown in Fig.\ \ref{fig:models}.
The models exhibit a growing degree of complexity with parameters representing fields acting on each spin ($h$, $h_1$ and $h_2$) and a direct connection between them ($J$), increasing the number of parameters from $\Theta = 0$ (top left model in the figure) where the spins are thought as two independent spins without any field, to $\Theta = 3$ (bottom right) for spins connected by a bond and being affected by two different effective fields. The vector $\vec{\phi}$ in Eq.\ \ref{eqn:equi_Ising2} is made up of combinations of the empirical mean values of the activity of each spin ($m_1$ and $m_2$) and the empirical correlation $c_{12}$.

As for the prior $P(\vec{\theta}| {\cal M}_i)$, we have already discussed in the Introduction a number of advantages for selecting the uninformative Jeffreys prior when no informations about the distribution of parameters are available a priori:
\begin{equation}\label{eqn:Jeffrey}
P(\vec{\theta}| {\cal M}_i) = \mathcal{N}^{-1} \sqrt{\det \mathbb{J}(\vec{\theta}) }
\end{equation}
where $ \mathbb{J}$ is the Fisher Information matrix whose components are given by the following expression $\mathbb{J}_{ij}(\vec{\theta}) = \partial_{\theta_i}\partial_{\theta_j} \log{Z(\vec{\theta})}$ and $\mathcal{N}= \int d\vec{\theta} \sqrt{\det \mathbb{J}(\vec{\theta}) }$. This choice of the prior is appealing not only because it assumes a flat distribution on the space of samples, but also because it corresponds to the same information theoretic assumption for all models. This avoids introducing unintended biases among the models. 
Using the Jeffreys prior, the probability of the observed data under model ${\cal M}_i$ can be written as
\begin{equation}\label{eqn:finalEq}
P(\hat{S}| {\cal M}_i) = \mathcal{N}^{-1}\int d\vec{\theta}\,\frac{e^{N \vec{\phi}\cdot\vec{\theta}}}{Z^N(\vec{\theta})}\sqrt{\det\mathbb{J}(\vec{\theta}) }.
\end{equation}

The heart of the problem is then solving Eq.\ \ref{eqn:finalEq} for each model, and ranking the resulting probabilities in order to identify which model is the most likely in each region of the $(m_1,m_2,c_{12})$ space which is what we will do in the following.

\begin{table}[t]
\centering
\begin{tabular}{|*{9}{c|}}
\hline
$\scriptstyle {\cal M}_i$ & $\scriptstyle \Theta$ & $\scriptstyle \vec{\theta}$ & $\scriptstyle \vec{\phi}$ & $\scriptstyle Z(\vec{\theta})$ & $\scriptstyle\det\mathbb{J}(\vec{\theta})$ & $\scriptstyle\mathcal{N}$ & $\scriptstyle\delta$ & $\scriptstyle\epsilon(\vec{\theta})$ \\
\hline
$\scriptstyle {\cal M}_1$ &$\scriptstyle 0$& - & - & $\scriptstyle 4$ & $\scriptstyle 1$ & - & $\scriptstyle 1$ & $\scriptstyle 2\log 2$\\
$\scriptstyle {\cal M}_2$ &$\scriptstyle 1$ & $\scriptstyle h_1$ & $\scriptstyle m_1$ & $\scriptstyle 4\cosh(h_1)$ & $\scriptstyle (\frac{4}{Z})^2$ & $\scriptstyle\pi$ & $\scriptstyle 1$ & $\scriptstyle 2\log 2$  \\
$\scriptstyle {\cal M}_3$ &$\scriptstyle 1 $& $\scriptstyle h_2$ & $\scriptstyle m_2$ & $\scriptstyle 4\cosh(h_2)$ & $\scriptstyle (\frac{4}{Z})^2$ & $\scriptstyle\pi$ & $\scriptstyle 1$ & $\scriptstyle 2\log 2$  \\
$\scriptstyle {\cal M}_4$ &$\scriptstyle 1$& $\scriptstyle h$ & $\scriptstyle m_1 + m_2$ & $\scriptstyle 4\cosh^2(h)$ & $\scriptstyle \frac{8}{Z}$ & $\scriptstyle\sqrt{2}\pi$ & $\scriptstyle \frac{1}{2}$ & $\scriptstyle \frac{3}{2}\log 2$  \\
$\scriptstyle {\cal M}_5$ &$\scriptstyle 2$ & $\scriptstyle (h_1,h_2)$ & $\scriptstyle (m_1,m_2)$ & $\scriptstyle 4\cosh(h_1)\cosh(h_2)$ & $\scriptstyle (\frac{4}{Z})^2$ & $\scriptstyle\pi^2$ & $\scriptstyle 1$ & $\scriptstyle 2\log 2$  \\
$\scriptstyle {\cal M}_6$ &$\scriptstyle 1$ & $\scriptstyle J$ & $\scriptstyle c_{12}$ & $\scriptstyle 4\cosh(J)$ & $\scriptstyle (\frac{4}{Z})^2$ & $\scriptstyle\pi$ & $\scriptstyle 1$ & $\scriptstyle 2\log 2$   \\
$\scriptstyle {\cal M}_7$ &$\scriptstyle 2$ & $\scriptstyle (h_1,J)$ & $\scriptstyle (m_1,c_{12})$ & $\scriptstyle 4\cosh(h_1)\cosh(J)$ & $\scriptstyle (\frac{4}{Z})^2$ & $\scriptstyle\pi^2$ &$\scriptstyle 1$ & $\scriptstyle 2\log 2$   \\
$\scriptstyle {\cal M}_8$ &$\scriptstyle 2$ & $\scriptstyle (h_2,J)$ & $\scriptstyle (m_2,c_{12})$ & $\scriptstyle 4\cosh(h_2)\cosh(J)$ & $\scriptstyle (\frac{4}{Z})^2$ & $\scriptstyle\pi^2$ & $\scriptstyle 1$ & $\scriptstyle 2\log 2$   \\
$\scriptstyle {\cal M}_9$ &$\scriptstyle 2$ & $\scriptstyle (h,J)$ & $\scriptstyle (m_1+m_2,c_{12})$ & $\scriptstyle 4\cosh^2(h)\cosh(J) \,+$ & $\scriptstyle \frac{2^7e^J}{Z^3}$ & $\scriptstyle 2\pi$ & $\scriptstyle \frac{3}{2}$ & $\scriptstyle \frac{7}{2}\log 2 +\frac{J}{2}$   \\
&  &  &  & $\scriptstyle + \, 4\sinh^2(h)\sinh(J)$ & &  & &\\
$\scriptstyle {\cal M}_{10}$ &$\scriptstyle 3$& $\scriptstyle (h_1,h_2,J)$ & $\scriptstyle (m_1,m_2,c_{12})$ & $\scriptstyle 4\cosh(h_1)\cosh(h_2)\cosh(J) \,+$ & $\scriptstyle (\frac{4}{Z})^4$ & $\scriptstyle\pi^2$ & $\scriptstyle 2$ & $\scriptstyle 4\log 2$ \\
&  &  &  & $\scriptstyle + \, 4\sinh(h_1)\sinh(h_2)\sinh(J)$ & & & & \\
\hline
\end{tabular}\caption{The ten competing models for a two spin system}
\label{table:models}
\end{table}
We first note that the determinant of the Fisher Information matrix is related to the partition function in the following way (see Tab.\ \ref{table:models})

\begin{equation}\label{eqn:Fisher}
\frac{1}{2}\log{\det\mathbb{J(\vec{\theta}})} = -\delta \log{Z(\vec{\theta})} + \epsilon(\vec{\theta}).
\end{equation} 
The values of the coefficients $\delta$ and $\epsilon(\vec{\theta})$ depend on the particular model as shown in Tab.\ \ref{table:models}: $\delta$ is a constant while epsilon is at most linear in $\vec{\theta}$. Consequently, performing the saddle point approximation and the Gaussian integral around the saddle, the integral in Eq.\ (\ref{eqn:finalEq}) becomes
\begin{equation}
\int d\vec{\theta}\,e^{N \Psi (\vec{\theta})} \sim \sqrt{\frac{(2\pi)^\Theta}{N^\Theta |\det\mathbb{H}(\vec{\theta}^\star)|}}e^{N \Psi (\vec{\theta}^\star)}
\end{equation}
where the argument of the exponential is
\begin{equation}\label{eqn:PsiwithDelta}
\Psi (\vec{\theta}) = \vec{\phi}\cdot\vec{\theta} - \left(1 + \frac{\delta}{N}\right)\log{Z(\vec{\theta})} + \frac{\epsilon(\vec{\theta})}{N}
\end{equation}
and $\mathbb{H}_{ij}(\vec{\theta}^\star) = \partial_{\theta_i}\partial_{\theta_j} \Psi(\vec{\theta})|_{\vec{\theta} = \vec{\theta}^\star}$ is the Hessian matrix calculated at $\vec{\theta}^\star$ which is the solution of the saddle point equations
\begin{equation}\label{eqn:saddlepoint}
\phi_i -  \left(1 + \frac{\delta}{N}\right)\partial_{\theta_i}\log{Z(\vec{\theta}^\star)} + \frac{\partial_{\theta_i}\epsilon(\vec{\theta}^\star)}{N} = 0.
\end{equation}
Using Eq.\ \ref{eqn:PsiwithDelta}, it is not hard to see that the determinant of the Hessian matrix is related to the determinant of the Fisher Information at the saddle point by the simple relation $ \det\mathbb{H}(\vec{\theta}^\star) = - \left(1 + \frac{\delta}{N}\right)^\Theta \det\mathbb{J}(\vec{\theta}^\star) $.

Putting all the previous results together, we can estimate the conditional probability $P(\hat{S}| {\cal M}_i)$ as
\begin{equation}\label{eqn:finalfinalEq}
P(\hat{S}| {\cal M}_i) = \sqrt{\frac{(2\pi)^\Theta}{\mathcal{N}^2 N^\Theta \left(1 + \frac{\delta}{N}\right)^\Theta}}\frac{e^{N \vec{\phi}\cdot\vec{\theta}^\star}}{Z^N(\vec{\theta}^\star)}.
\end{equation}

As anticipated, the resulting expression in Eq.\ \ref{eqn:finalfinalEq}  is made up of two terms: the maximum likelihood one which counts for the goodness of fit, and a complexity cost which depends on the dimensionality of the models $\Theta$ and on the shape of the prior

\begin{equation}\label{eqn:complexity}
\begin{array}{l}
\displaystyle {\rm Complexity\ Cost} = e^{-C} \vspace{2mm}\\
\displaystyle C \simeq \frac{\Theta}{2}\log{\frac{N}{2\pi}} + \log\int d\vec{\theta} \sqrt{\det \mathbb{J}(\vec{\theta}) }
\end{array}
\end{equation}

Similar to the Bayesian Information Criterion \cite{BIC}, this complexity cost consists of a term proportional to $\log N$ that arises from the dimensionality of the model and a second term accounting for the geometric complexity pertaining to the number of distinguishable probability distribution encoded in a parametric family distribution\cite{Myung00}. The criterion described here matches exactly the Minimum Description Length principle\cite{Rissanen1996} and therefore the maximisation of this cost function would ensure the greatest compression of the data description. As one can notice from the results obtained so far, the corrections related to the complexity term are $O(\log{N})$ and consequently they are evident when $\log{N}/N$ is not negligible. For $N \rightarrow \infty$ one recovers MLE of a two spins system \cite{Yasser09}.

\subsection{\label{sec: parameterspace} The selected model}
As described in the previous section, the probability $P(\hat{S}| {\cal M}_i)$ is fully determined, through Eq.\ \ref{eqn:finalfinalEq} from the measured statistics $\{m_1,m_2,c_{12}\}$. In this section we discuss how the space of observations $\{m_1,m_2,c_{12}\}$ is partitioned among the models: for each point in this space, we rank the posterior probability of each model to see which is the most likely model. It is worth noting that not all the points in this space are achievable in the limit $N \rightarrow \infty$, therefore we will discard un-physical points lying outside the tetrahedron identified by the following double inequality: $-1 + |m_1 + m_2| \leq c_{12} \leq 1 - |m1 - m2|$.

\begin{figure}[t!]
\centering
\subfigure{\includegraphics[width = 0.49\columnwidth]{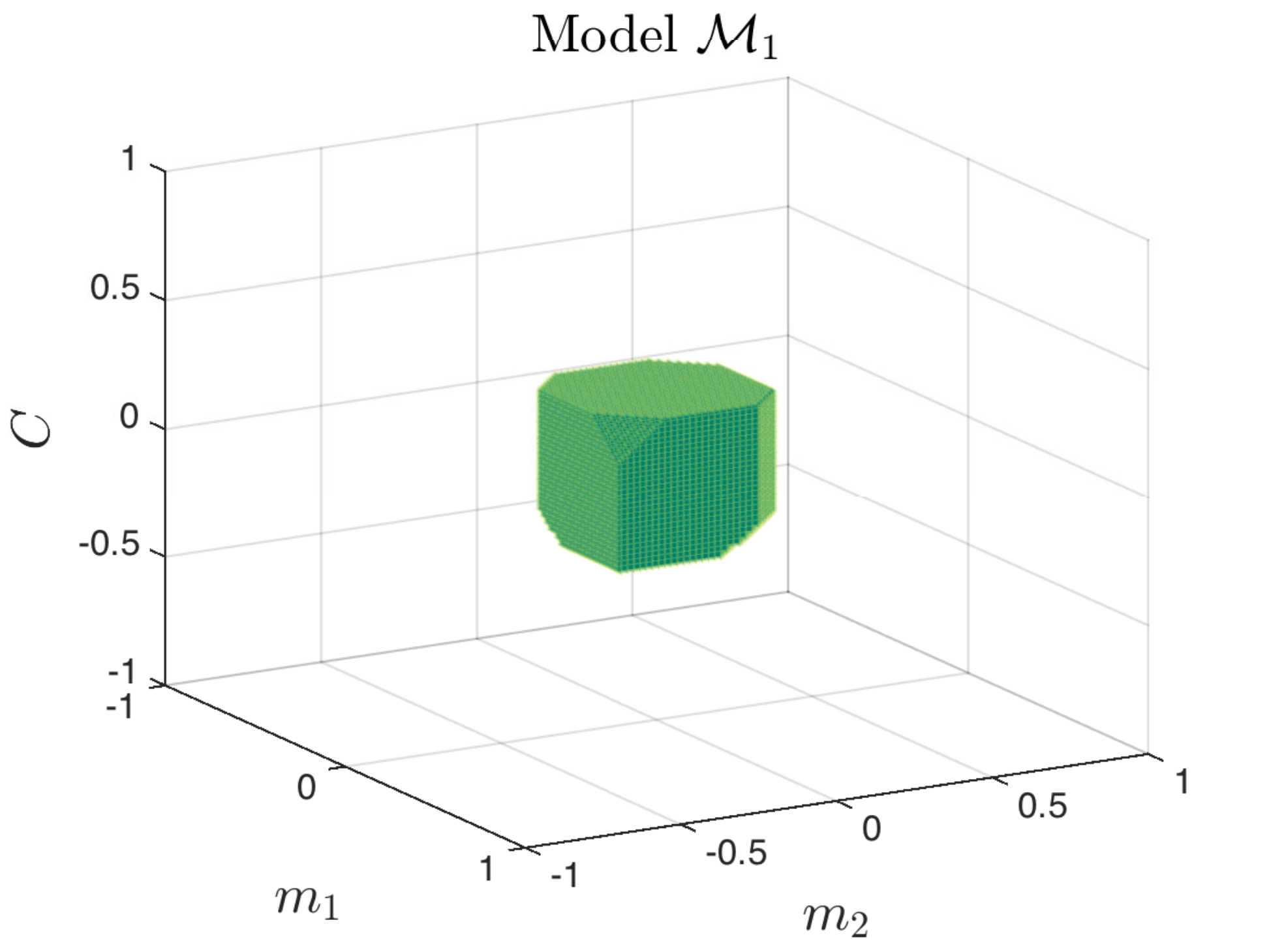}}
\subfigure{\includegraphics[width = 0.49\columnwidth]{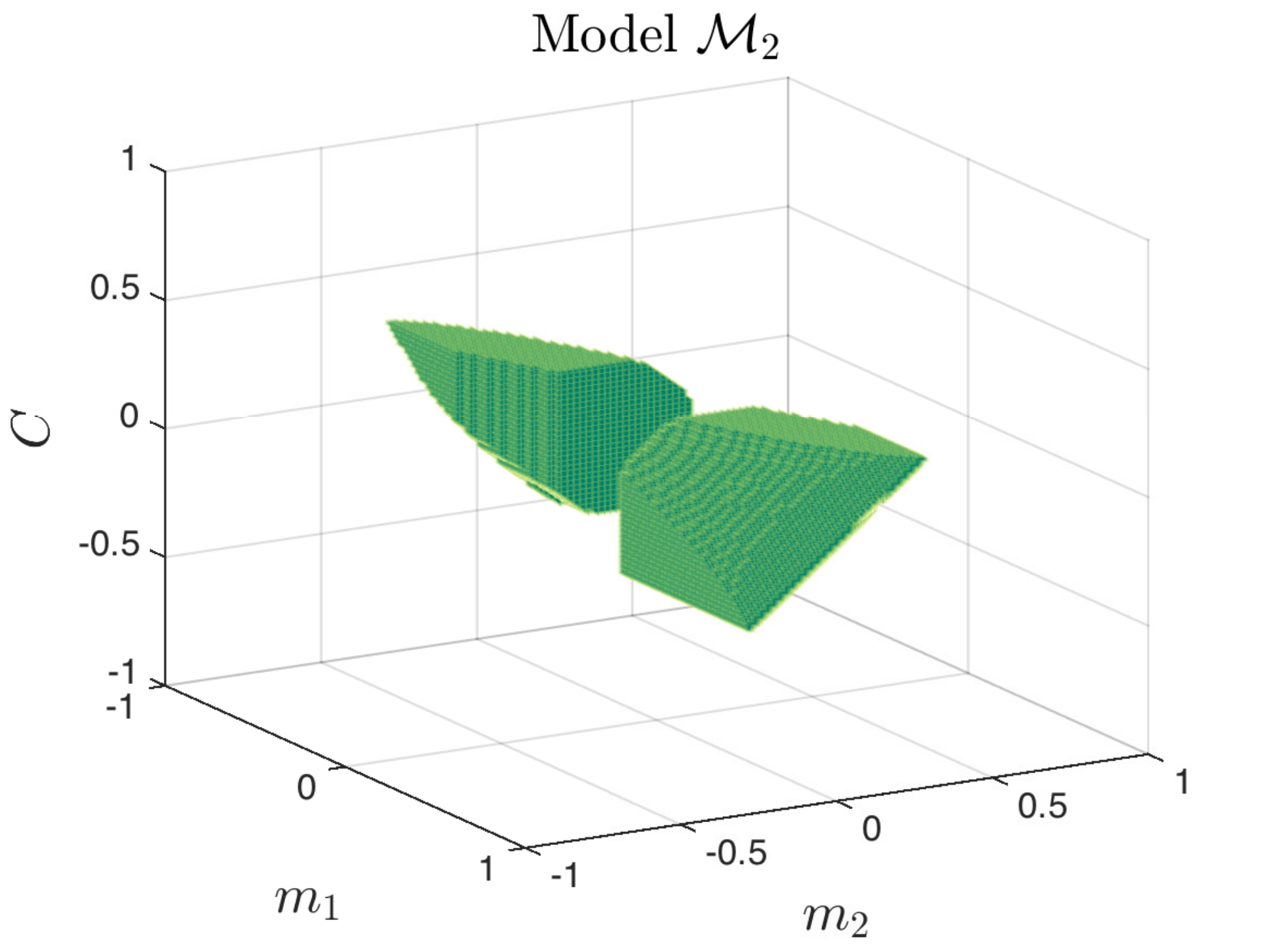}}
\subfigure{\includegraphics[width = 0.49\columnwidth]{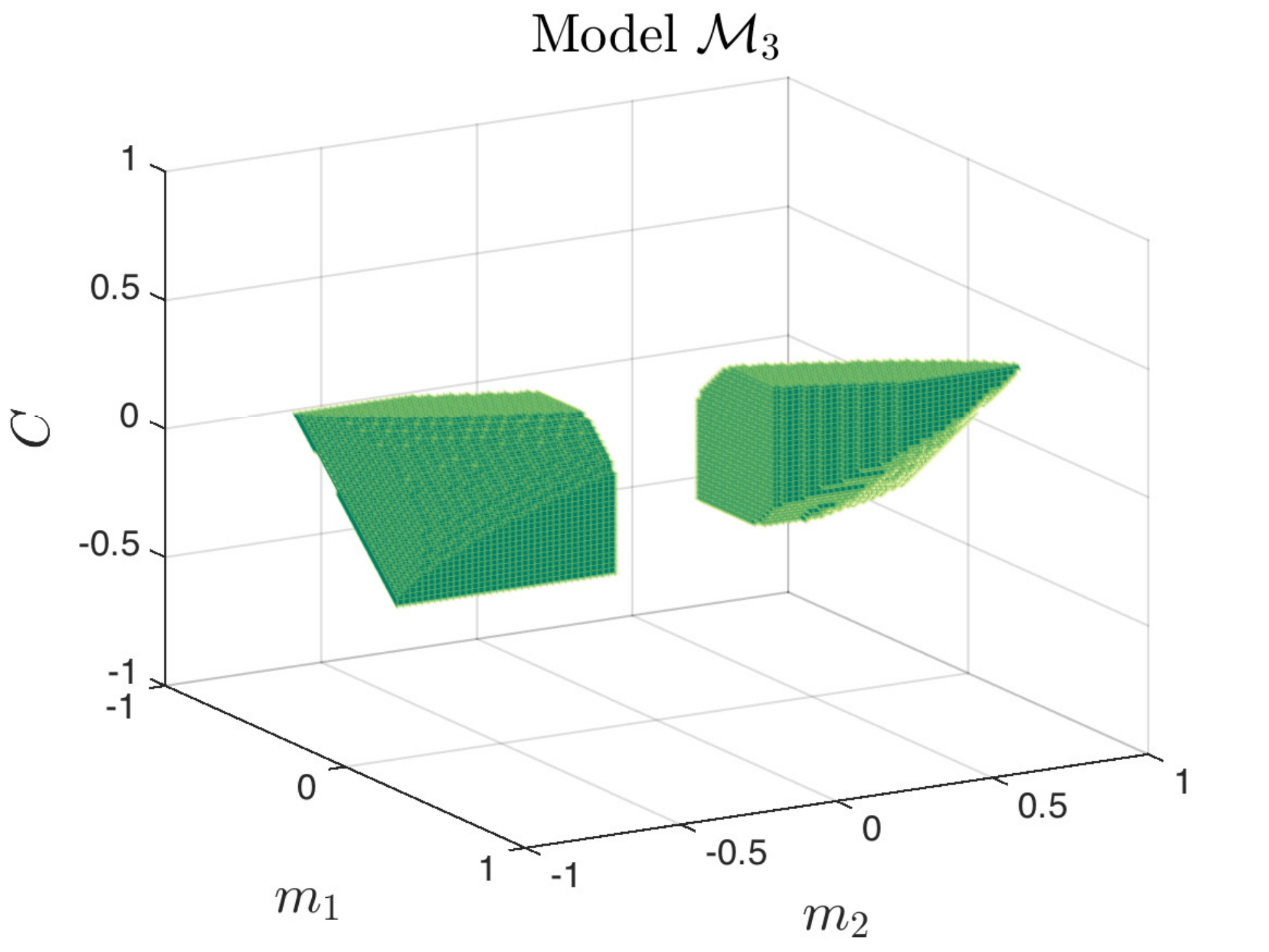}}
\subfigure{\includegraphics[width = 0.49\columnwidth]{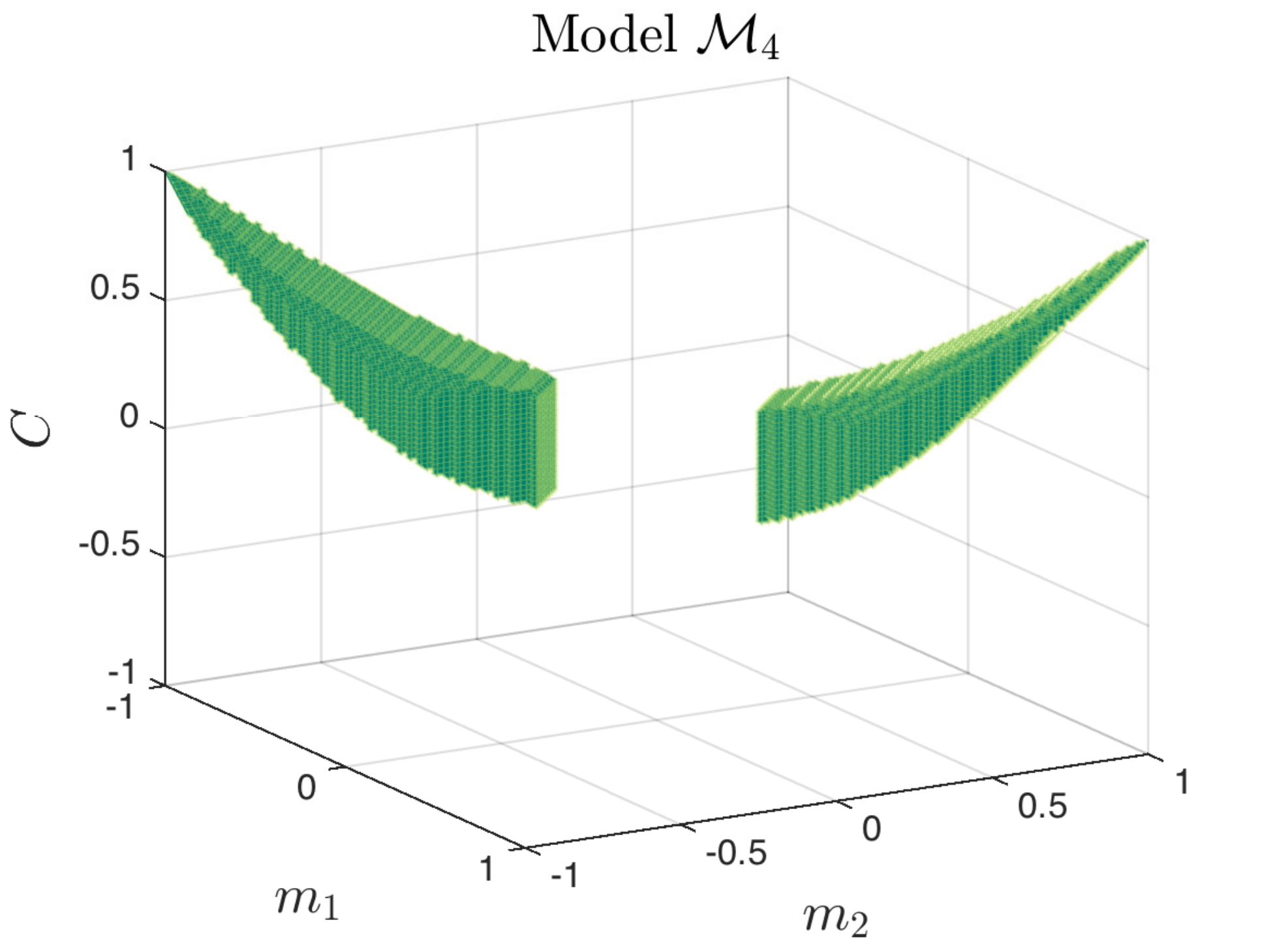}}
\subfigure{\includegraphics[width = 0.49\columnwidth]{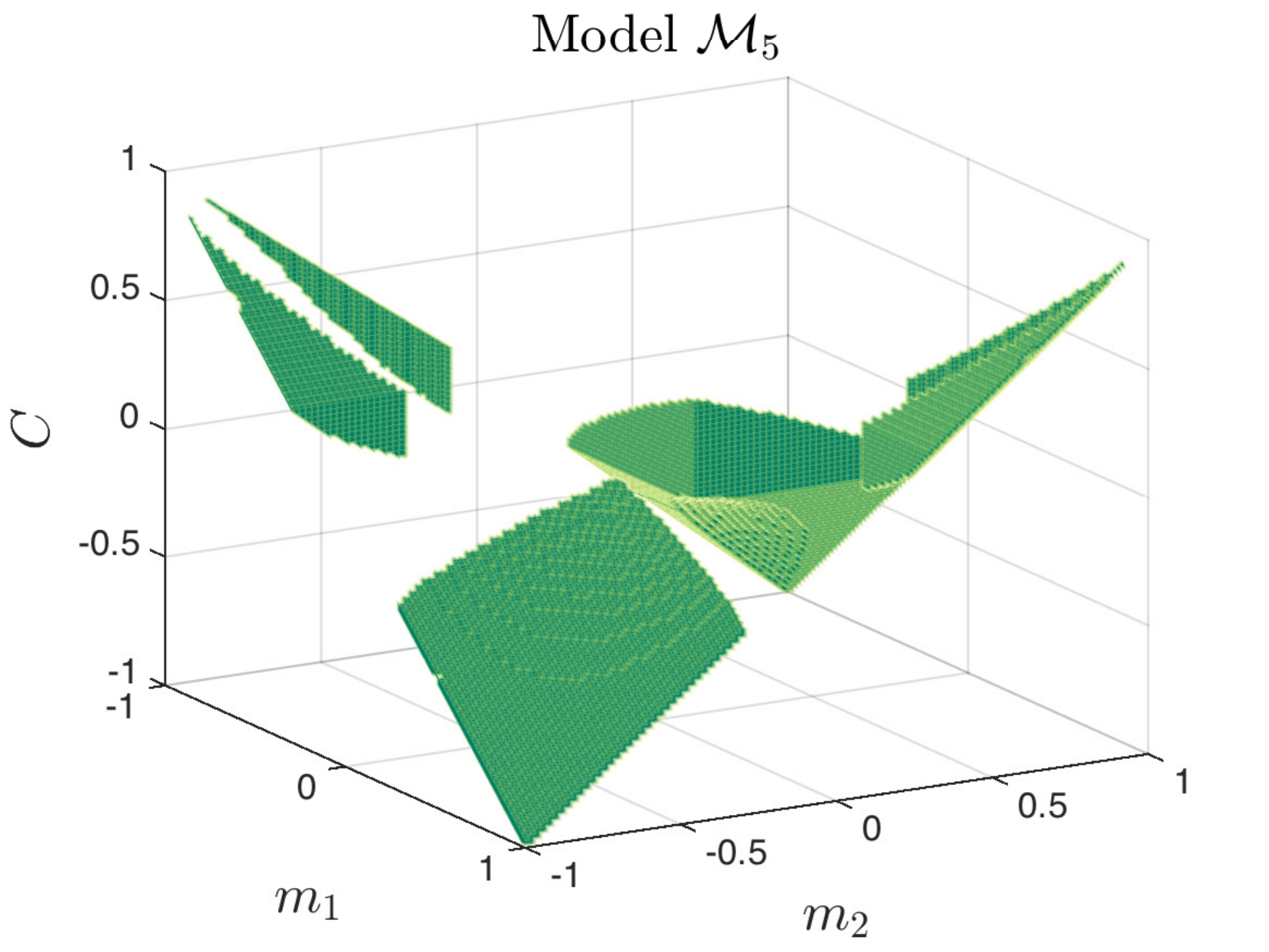}}
\subfigure{\includegraphics[width = 0.49\columnwidth]{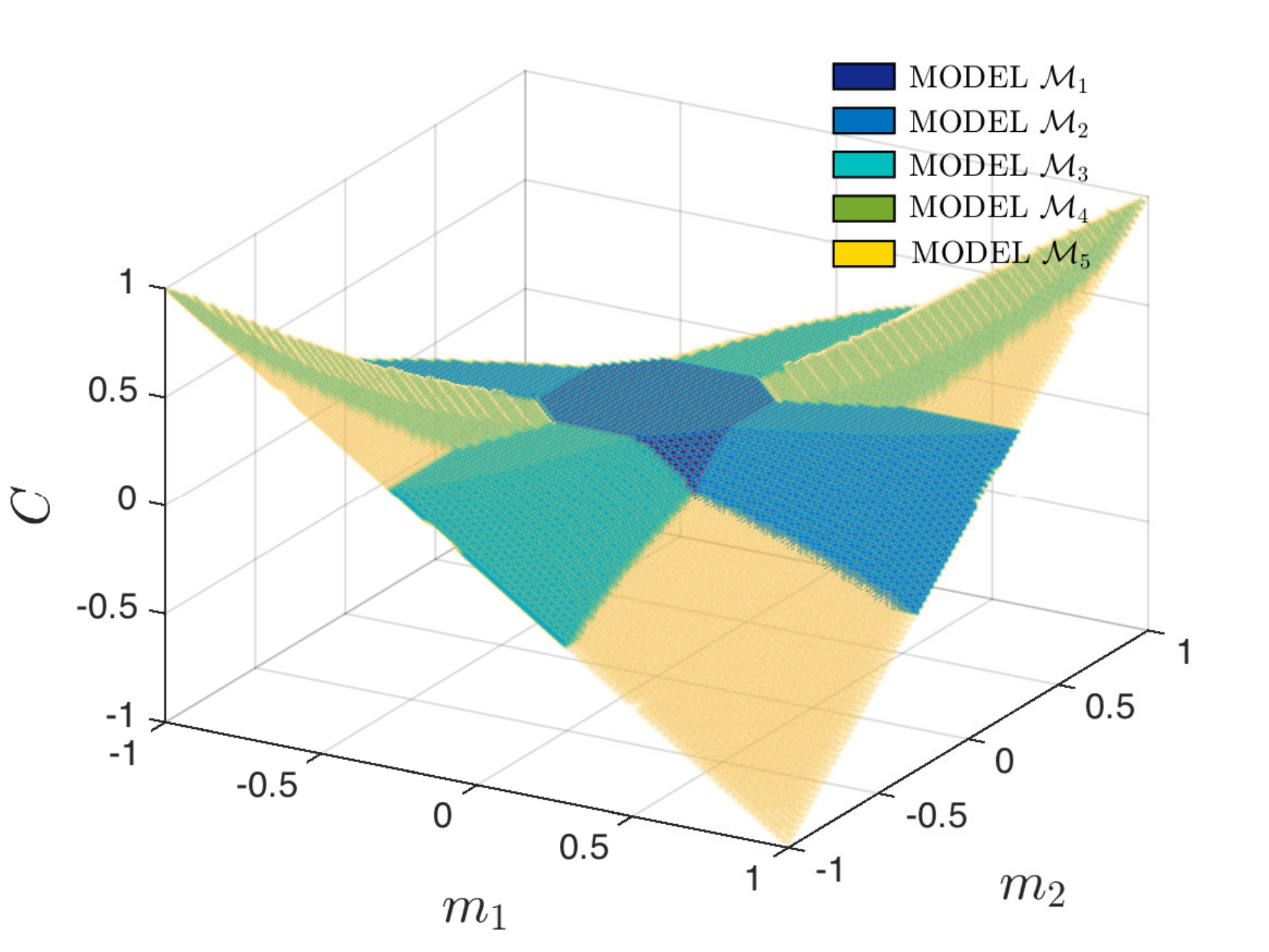}}
\caption{The figures show the regions of the space $\{m_1,m_2,c_{12}\}$ where the first five models in Tab.\ \ref{table:models} are the most likely according to $P({\cal M}_i|\hat{S})$ given $N = 50$ measurements. The region of the space covered by all five models is shown in the last figures, with different colors for different models, and it encloses the surface $c_{12} = m_1m_2$ which represents the limit of independent spins for $N \rightarrow \infty$}.
\label{fig:nbModels50}
\end{figure}

\begin{figure}[t!]
\centering
\subfigure{\includegraphics[width = 0.49\columnwidth]{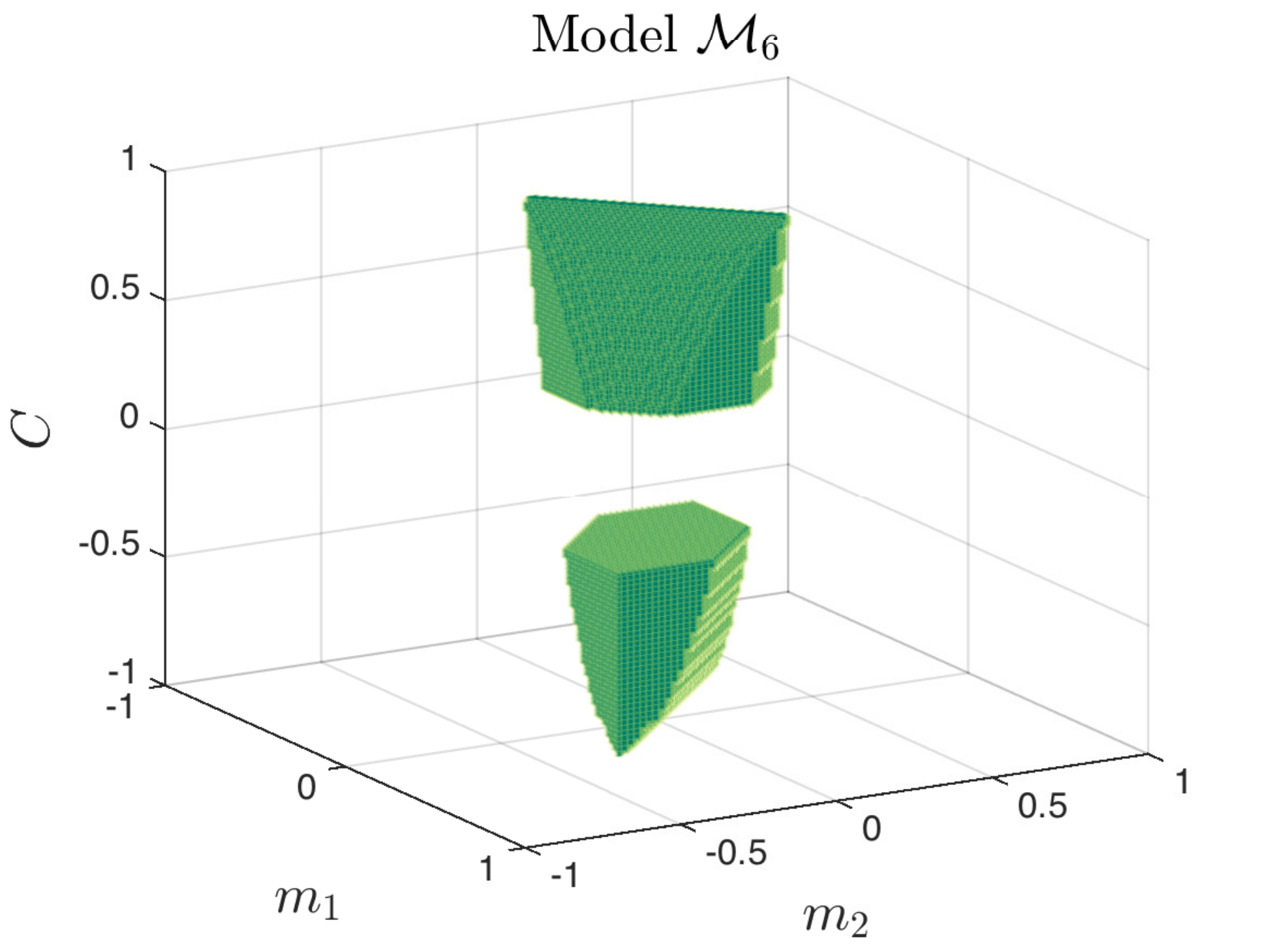}}
\subfigure{\includegraphics[width = 0.49\columnwidth]{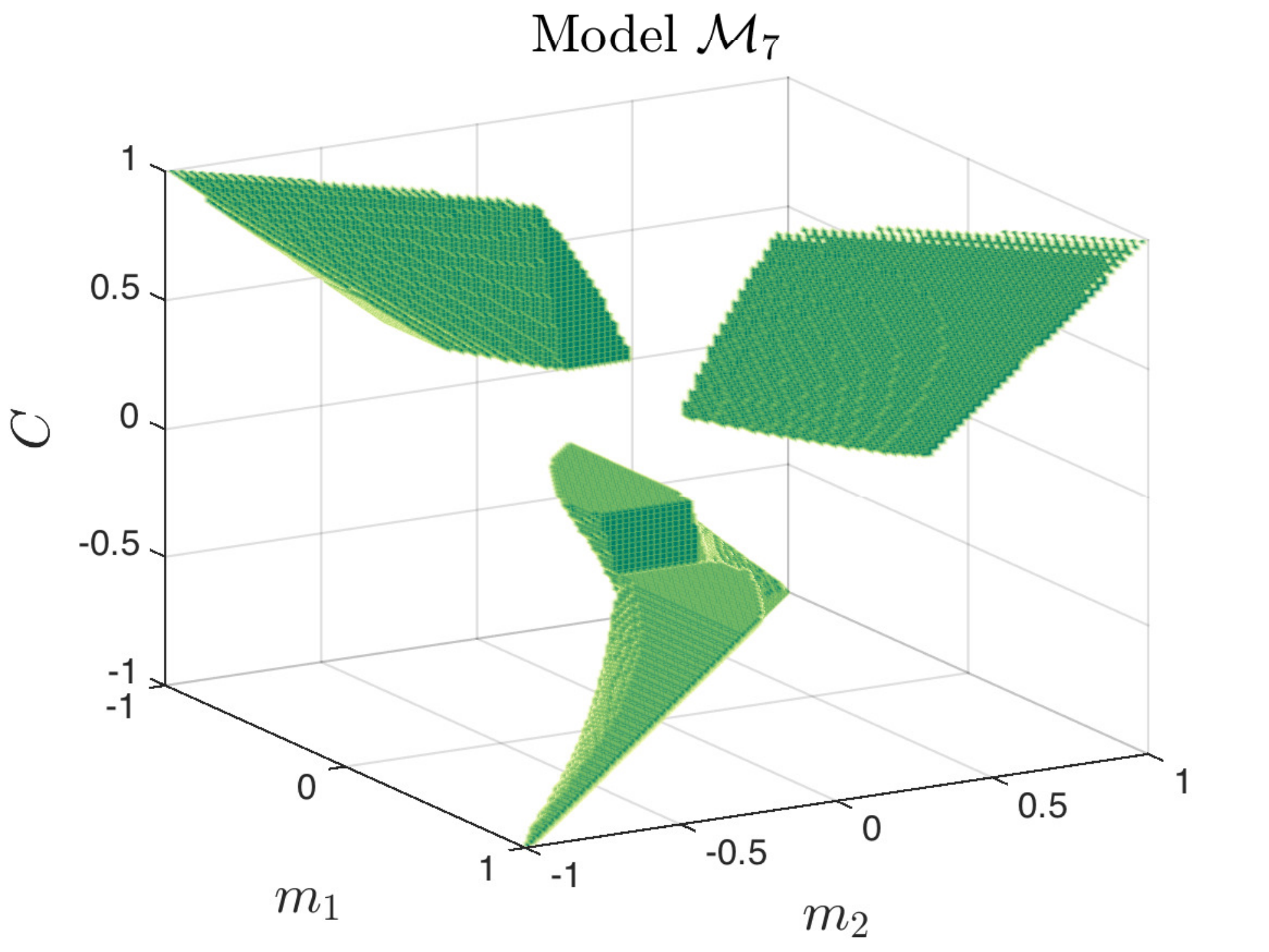}}
\subfigure{\includegraphics[width = 0.49\columnwidth]{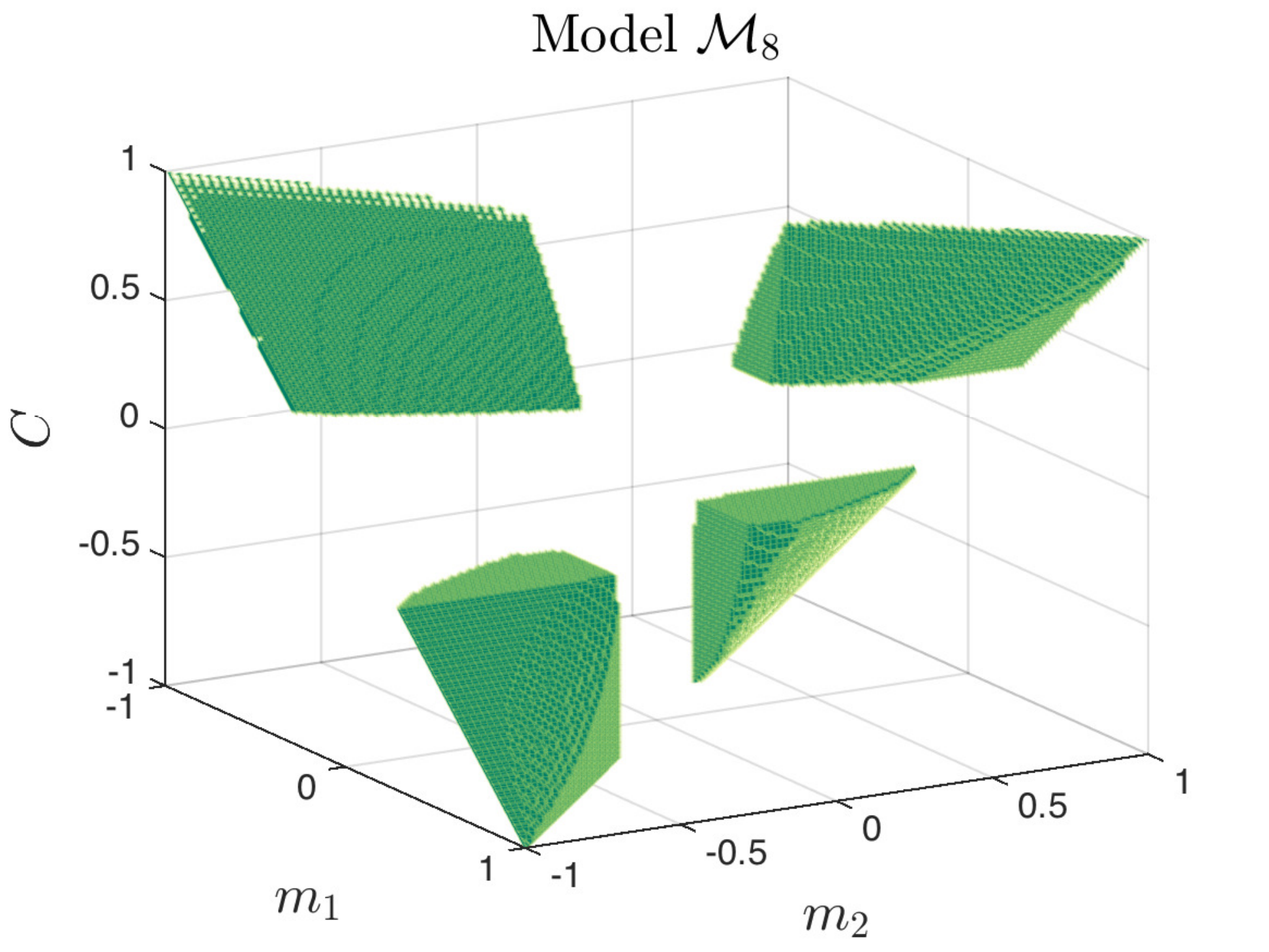}}
\subfigure{\includegraphics[width = 0.49\columnwidth]{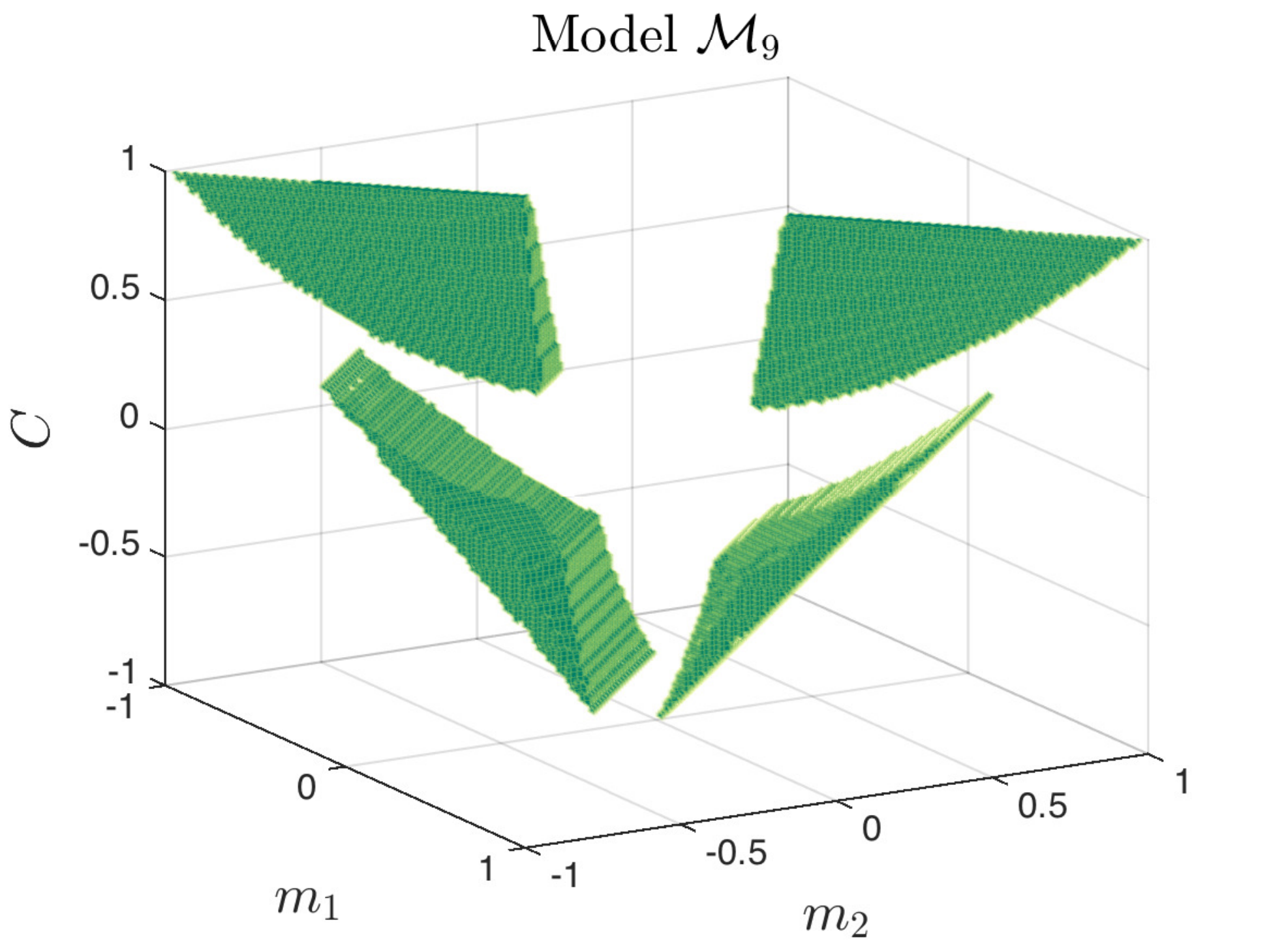}}
\subfigure{\includegraphics[width = 0.49\columnwidth]{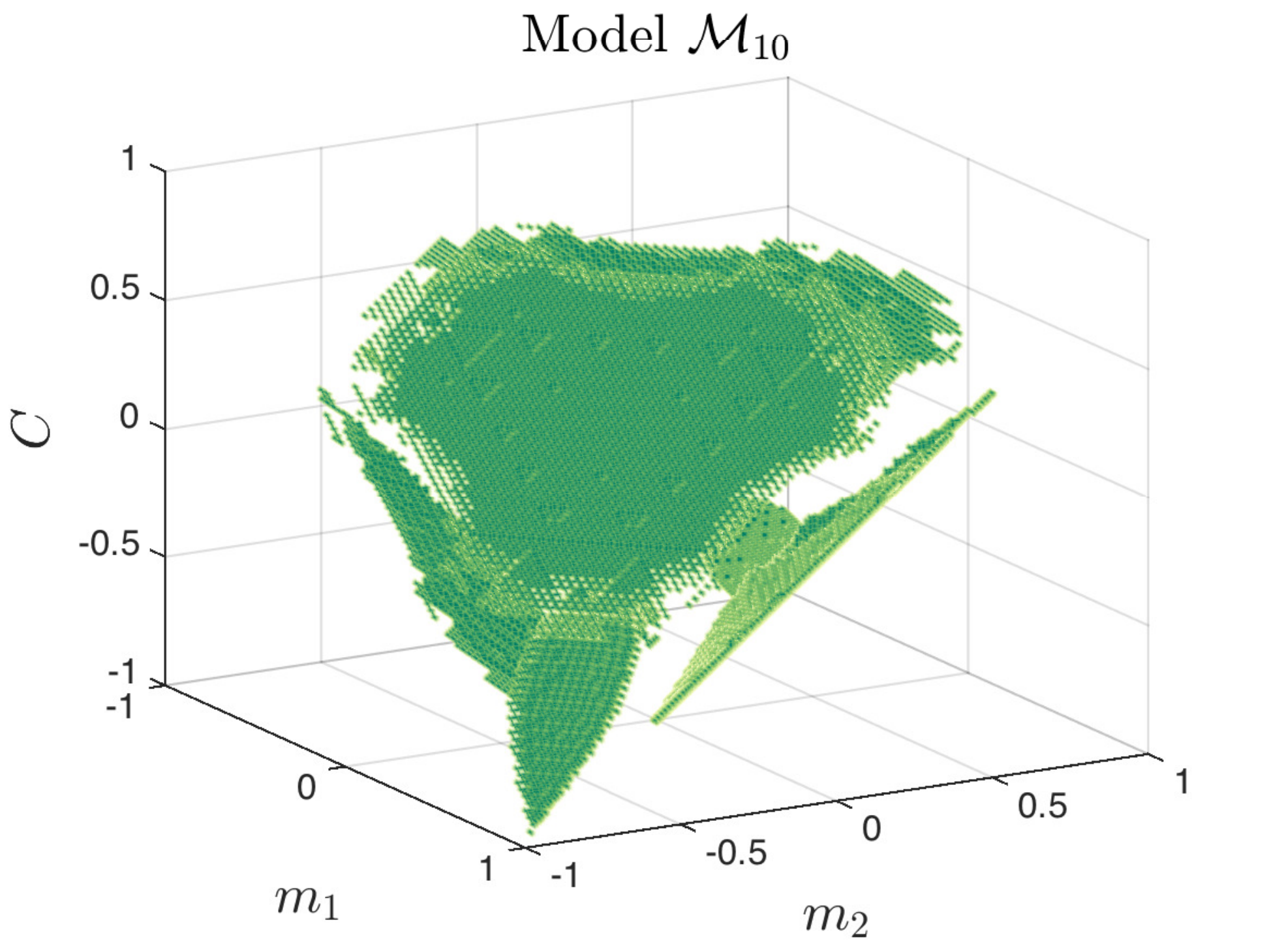}}
\subfigure{\includegraphics[width = 0.49\columnwidth]{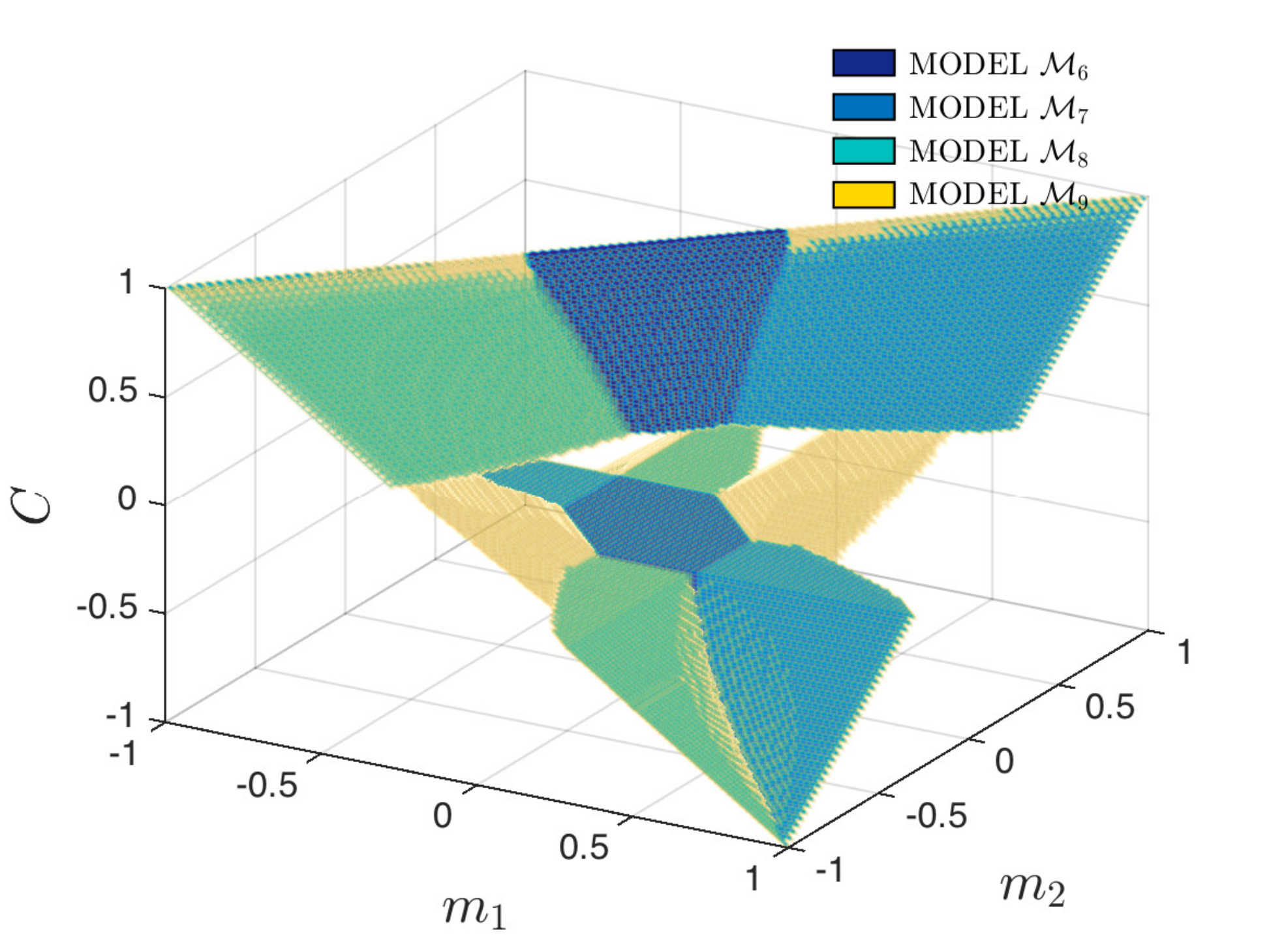}}
\caption{The figures show the regions of the space $\{m_1,m_2,c_{12}\}$ where the last five models in Tab.\ \ref{table:models} are the most likely according to $P({\cal M}_i|\hat{S})$ given $N = 50$ measurements. All these models employ the parameter $J$ meaning that the spins are thought to be directly interacting here in contrast with the first five models. The last figure summarizes how the $\{m_1,m_2,c_{12}\}$ is partitioned among these models (Model ${\cal M}_{10}$ has not been included in the last figure for reasons of clarity).}
\label{fig:bModels50}
\end{figure}

In Fig.\ \ref{fig:nbModels50} and Fig.\ \ref{fig:bModels50} the division of the parameters space is sketched for $N = 50$. Fig.\ \ref{fig:nbModels50} illustrates the parts of the $\{m_1,m_2,c_{12}\}$ space where the first five models in Tab.\ \ref{table:models} are the most likely according to $P({\cal M}_i|\hat{S})$ for a data length of $N = 50$. These five models are the ones in which the two spins are not directly connected. The region of the space covered by the union of these five models, shown in the last picture with different colors for each subregion, lies around the surface $c_{12} = m_1m_2$. In the limit $N \rightarrow \infty$, as one expects, the union of these five models will become identical to the surface $c_{12} = m_1m_2$. 

The last five models in Tab.\ \ref{table:models} describe models in which the spins do interact with each other. Fig.\ \ref{fig:bModels50} shows the regions in the $\{m_1,m_2,c_{12}\}$ space where these models are preferred over the first (disconnected spin) models. 

\begin{figure}[t]
\centering
\subfigure{\includegraphics[width = 0.49\columnwidth]{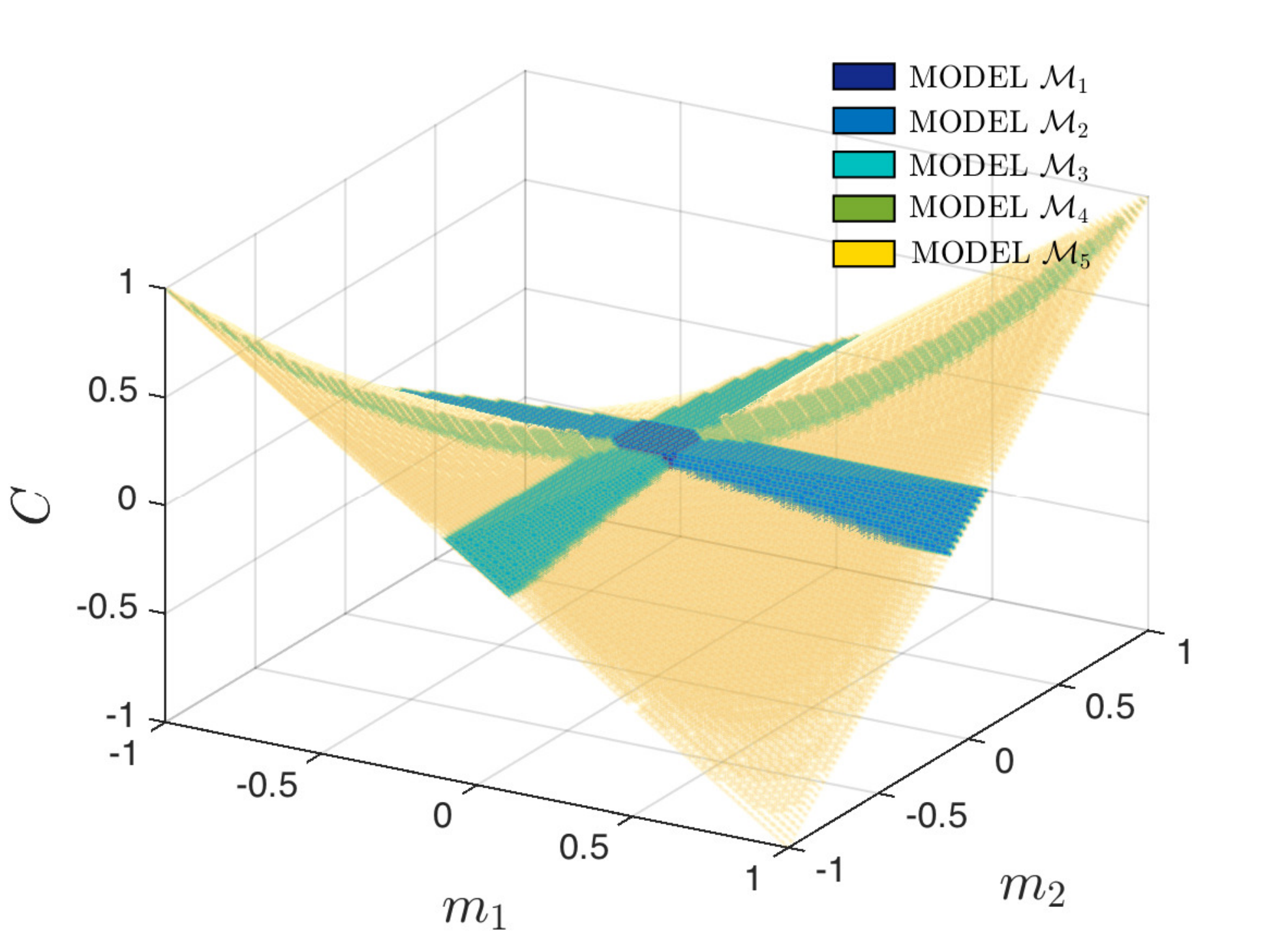}}
\subfigure{\includegraphics[width = 0.49\columnwidth]{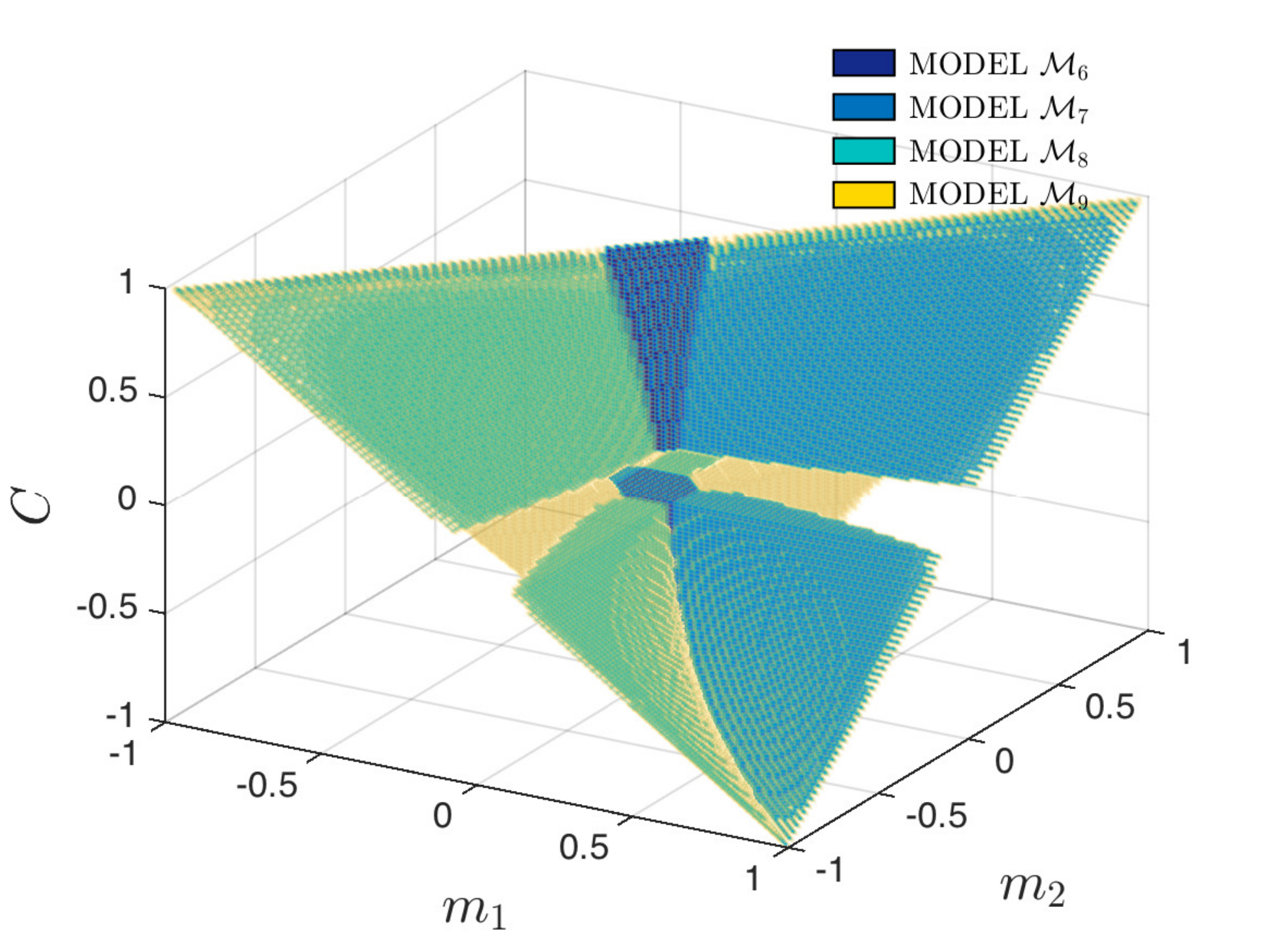}}
\caption{Similarly to the previous figures for $N=50$, these figures show the partitioning of the space of observations $\{m_1,m_2,c_{12}\}$ among models without a parameter for direct interaction between spins (left) and models with it (right) for $N=500$. }
\label{fig:Models500}
\end{figure}

In Fig.\ \ref{fig:Models500} we show the partitioning of the parameter space among the models for the case $N = 500$: left panel of Fig.\ \ref{fig:Models500} summarizes the case of models with no interactions between the spins and the right one represents models with interactions. One can see that for large $N$, models with less parameters extend over smaller regions. In this case, the regions belonging to the first five models shrink around the surface $c_{12} = m_1m_2$ and more complex models (${\cal M}_5$ among the first five models and ${\cal M}_{10}$ among the rest of them) occupy more volume. In fact, as remarked in the previous section, the penalty arising from the complexity of models is greater for smaller samples, and for very large values of $N$, the maximum likelihood estimates become predominant and models with more non-zero parameters are preferred.

\begin{figure}[thb]
\centering
\subfigure{\includegraphics[width = 0.49\columnwidth]{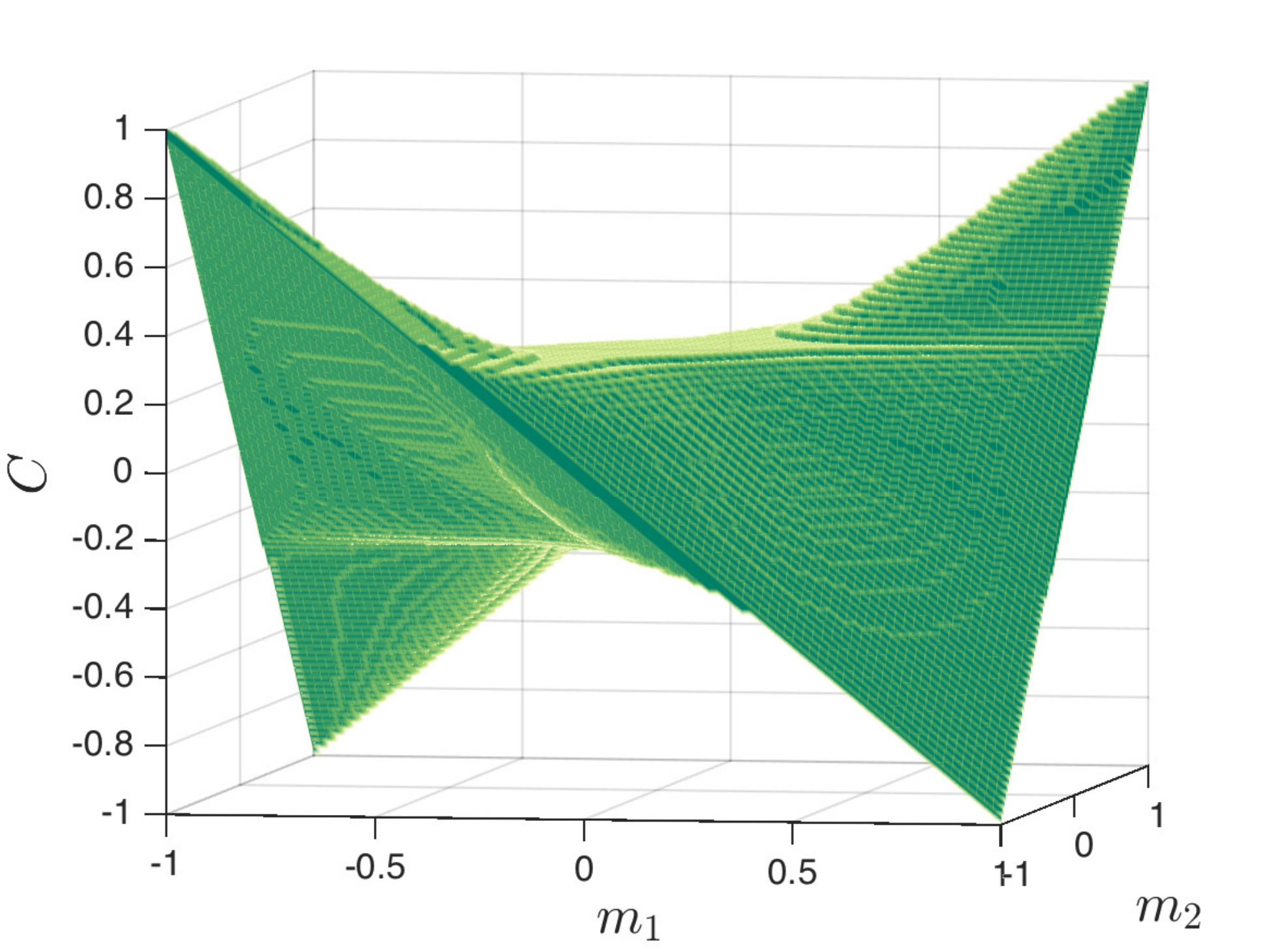}}
\subfigure{\includegraphics[width = 0.49\columnwidth]{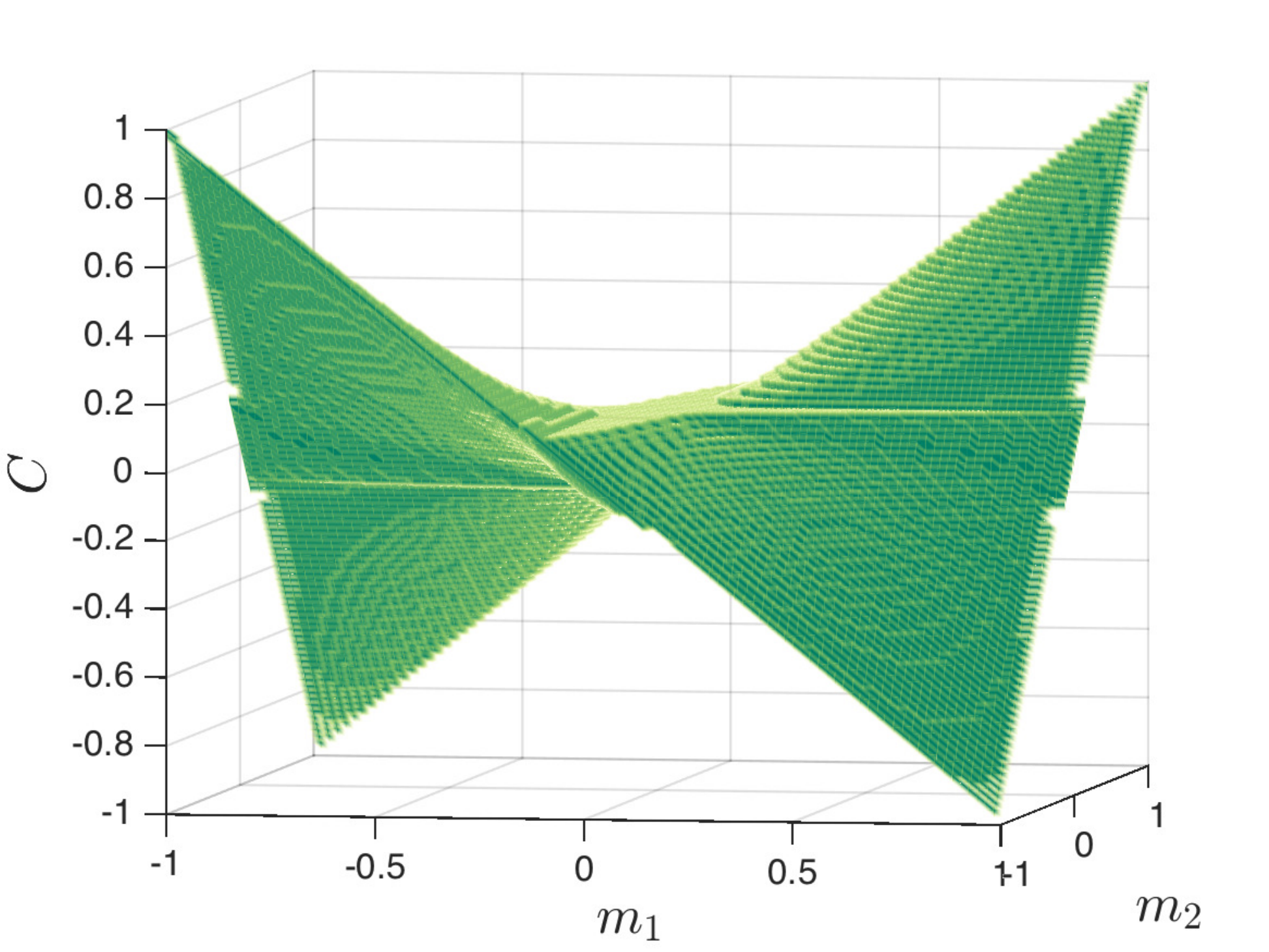}}
\caption{Portion of the space of observations in which the probability of not having a bond is greater than the one of having a bond for $N=50$ (left) and $N=500$ (right). }
\label{fig:BondNoBond}
\end{figure}

From what we have said so far, one can calculate the conditional probability $P(b |\hat{S})$ that the two spins are directly interacting. If we assume a priori all models have the same probability, it follows from Eq.\ \ref{eqn:Bayes} that $P(b |\hat{S})$ is proportional to the sum of $P(\hat{S}|{\cal M}_i)$ over all models ${\cal M}_i$ in which there is a bond between the pair of spins, that is the last 5 models in the Tab.\ \ref{table:models}
\begin{equation}
P(b |\hat{S}) = \Gamma^{-1}\sum_{i=6}^{10}
P(\hat{S}|{\cal M}_i).
\end{equation}
$\Gamma$ is given by 
 \begin{equation}
 \Gamma = \sum_{i=1}^{10}
P(\hat{S}|{\cal M}_i)
 \end{equation}
and ensures normalization $P(b|\hat{S}) + P(nb|\hat{S})  = 1$, with $P(nb|\hat{S}) = \Gamma^{-1}\sum_{i=1}^5
 P(\hat{S}|{\cal M}_i)$ denoting the probability of no bonds (nb). It follows that if the quantity $\eta = P(b |\hat{S}) - P(nb |\hat{S})$, which we will refer to as the {\itshape confidence} in the following, is greater than or equal to zero, we put a bond between the spins, otherwise we don't (Notice that {\itshape confidence} is defined as the difference between the probability of having a bond and the probability of not having it when a uniform prior on models is assumed). Accordingly the space $\{m_1,m_2,c_{12}\}$ becomes divided into two regions: one in which the probability of not having a bond is bigger and one in which the opposite is true. This subdivision is shown in Fig.\ \ref{fig:BondNoBond} for two data set of different length: $N=50$ and $N=500$. The no-bond region contains the surface $c_{12} = m_1m_2$ with a thickness that decreases as $N$ increases. 

This procedure for the identification of the bond will be performed for each pair of spins in a network independently in order to recover its topology.

\subsection{\label{sec: selfconsistent} A self consistent procedure for selecting the sparsity priors}

The assumption that the models are all a priori equally likely that we made in the last section can be easily relaxed in order to exploit additional knowledge or beliefs on the degree of sparsity of the network. For instance, we can assume that  $P_0({\cal M}_i) = P_0(b)$ for all models that employ a bond ($i=6,\ldots,10$) 
and $P_0({\cal M}_i) = P_0(nb) =1-P_0(b)$ for the rest ($i=1,\ldots, 5$).
The ratio $\epsilon = P_0(b)/P_0(nb)$ is the only parameter to choose. The value of $\epsilon$ reflects the a priori belief about the {\itshape sparsity} of the network and it is often compared in the text with the actual ratio between the number of 'bonds' and the number of 'no-bonds' in the graph, i.e. $r= n_b/n_{nb}$. The sparsity of the network is usually defined as the ratio between the number of bonds and the total number of possible connections which is equal to $r/(1+r)$. However, for sparse matrices $r\ll 1$, the \emph{`bond'-`no-bond' ratio} $r$ approximates very well the sparsity of the network $\sim r + O(r^2)$. With the introduction of the above priors, the quantity $P(b |\hat{S}) - P(nb |\hat{S})$, now called $\tilde{\eta}$, takes the following expression
\begin{equation}
P(b |\hat{S}) - P(nb |\hat{S}) = \frac{\Gamma}{2\bar{\Gamma}}\left( \eta - \frac{1-\epsilon}{1+\epsilon}\right)
\end{equation}
where $\eta$ is the confidence, $\Gamma$ has been defined in the previous section and 
\begin{equation}
{\bar{\Gamma}} = \frac{\epsilon}{1+\epsilon}
\sum_{i=6}^{10} 
P(\hat{S}|{\cal M}_i) + \frac{1}{1+\epsilon}
\sum_{i=1}^5 P(\hat{S}|{\cal M}_i).
\end{equation}
Therefore $P(b |\hat{S}) - P(nb |\hat{S}) \geq 0 $ if and only if $\eta \geq (1-\epsilon)/(1+\epsilon)$. This means that adding a prior of that kind basically implies a non-zero threshold for the quantity $\eta$ (for $\epsilon = 1$, the case of a flat prior is indeed retrieved).

An interesting procedure for deciding about the choice of $\epsilon$ is a self consistent one: one chooses $\epsilon$ such that after the graph recovery is performed,  the ratio of the present to absent bonds in the recovered graph also equals $\epsilon$, that is ($n_b/n_{nb} = P_{0}(b)/P_{0}(nb))$. A non-trivial self consistent choice for $\epsilon$ may not exist\footnote{$\epsilon=0$ or $\epsilon= +\infty$ are always trivially self-consistent.}, but as we have seen in our numerical simulations, it often does. When it exists, it can be reached as the fixed point of an iterative procedure where at step $t=0,1,\ldots$, given $\epsilon_t$, one computes $r_t$ and sets $\epsilon_{t+1}=r_t$ for the next iteration. In all cases we studied, either convergence to a fixed point $\epsilon=r$ was chosen, regardless of the first guess $\epsilon_0$, or a completely disconnected graph retrieved ($\epsilon=0$). In the latter case, when the point $r = \epsilon = 0$ becomes stable, a fixed prior could be used to check the result. 

\section{\label{sec: results} Numerical results for network recovery}
To assess the performance of the Model Selection (MS) approach described in the preceding sections in recovering a network topology, we applied it to synthetic data from an equilibrium Ising model with different connectivity patterns and analysed the goodness of reconstruction. In doing so, we also studied the effect of different parameters  on the performance of the MS approach, such as the strength of couplings, the number of samples, the prior belief $\epsilon$ on the sparsity and the size of the network. In addition, we compared the results with the ones obtained with PLM+$\ell_1$, which is among the best algorithms for recovering the connectivity. Unfortunately we found that the approach proposed in \cite{Decelle13} was not stable and convergent in the very low data limit that we were interested in, and therefore we do not discuss it below. 

The graphs we investigated are:
\begin{enumerate}
\item Gas of dimers in which each node is connected to one and only one other node, 
\item Star graph
\item Erd\"os R\'enyi  graphs with the average degree  $c=2$
\item Erd\"os R\'enyi  graphs with the average degree $c=3$ 
\item  two dimensional (2D) regular grid connectivity. 
\item diluted two dimensional regular grid
\end{enumerate}
Across these graphs the bond-to-no-bond ratio $r$ ranges from $0.01$ to $0.06$ (for $n=64$), we also studied network sizes ranging from $n=16$ to $100$ nodes and sample sizes from $N=50$ to $2000$. The couplings are drawn from a bimodal distribution $J=\pm \beta$, but we also investigated the ferromagnetic case ($J=\beta$), for different values of $\beta$. We did not analyse situations in which the couplings were drawn from a continuous distribution around zero because in order to distinguish a indefinitely small coupling from an absent one a very large sample would have needed and this is surely not the case of the regime we are investigating.

For the sake of brevity, we discuss the results highlighting few representative cases, while referring to the Appendix. We shall contrast the case of fully observed graphs, where all the variables on the nodes of the graph are observed, to that of partially observed ones, where the data only accounts for a fraction of the variables while the remaining ones act as hidden nodes or unknown unknowns.

\subsection{\label{subsec: results_full} Fully observed graphs}

\begin{figure}[t!]
\centering
\includegraphics[width = 0.95\columnwidth]{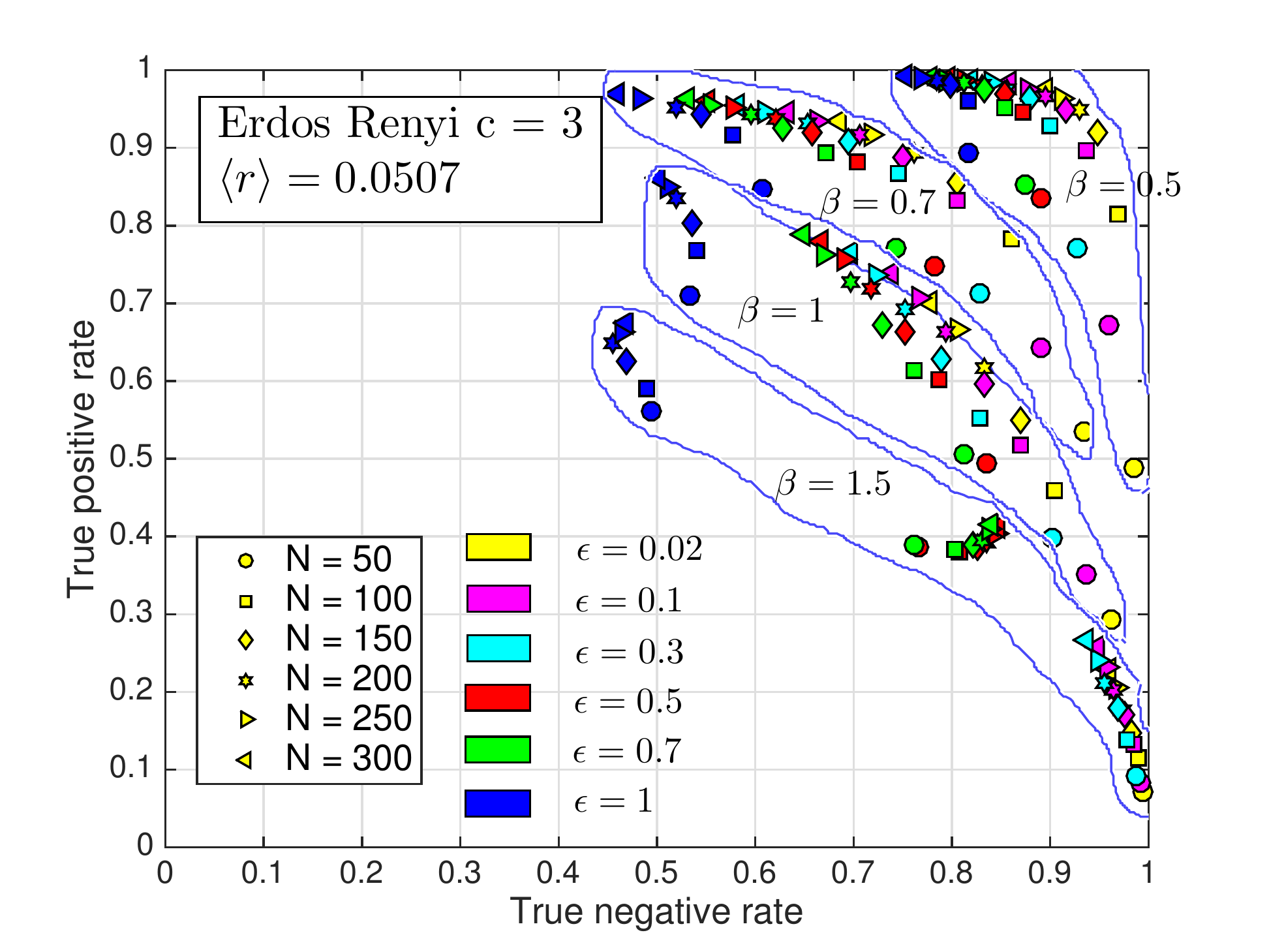}
\caption{Testing MS method on synthetic data for an Erd\"os R\'enyi  graphs with average degree  $c=3$. 
The coordinates of each point in the figure represents the average over one hundred realisations of the {\itshape true negative rate} ($TNR$) and the {\itshape true positive rate} ($TPR$) for a given value of $N$ and $\epsilon$. Contours have been drawn to cluster the points corresponding to the same value of $\beta$. The average sparsity, i.e. the ratio between the number of connected pairs of spins and the number of not connected ones, is $\langle r \rangle =0.0507$. 
Similar plots for other topologies can be found in the Appendix.}
\label{fig:scatterMs}
\end{figure}

Fig.\ \ref{fig:scatterMs} summarizes the results for the representative case of the Erd\"os R\'enyi graph with $c=3$, for different samples sizes $N$, strength of the couplings $\beta$ and different choices of the prior sparsity $\epsilon$. The coordinates of each point in the figure represent the average values over one hundred independent realisations of {\itshape true negative rate} ($TNR$) and the {\itshape true positive rate} ($TPR$) for a given value of $N$, $\epsilon$ and $\beta$, where TPR (TNR) stands for the fraction of bonds (no-bonds) correctly recovered.  

A perfect inference of the graph corresponds to a point on the top right corner of the graph. For graphs with very few bonds and no loops (e.g. a gas of dimers) MS achieves almost perfect recovery (see Appendix). Fig.\ \ref{fig:scatterMs} shows instead a typical case where MS is expected to provide only an approximate reconstruction. MS network recovery works best for weak interactions ($\beta=0.5$), because when interaction gets strong (e.g. $\beta=1.5$) the effects of indirect interactions, that are neglected by MS, become important. Interestingly, in the case of a gas of dimers we see the opposite, i.e. reconstruction improves when the interaction gets stronger (see Appendix). Secondly, the dependence of TPR and TNR with $N$  conforms to what is expected from the discussion in the previous sections: for small samples, MS favours simpler models, i.e. those with few bonds. Hence one expects the TPR (TNR) to increase (decrease) with $N$, as in Fig.\ \ref{fig:scatterMs}.

\begin{figure}[t!]
\centering
\subfigure{\includegraphics[width = 0.45\columnwidth]{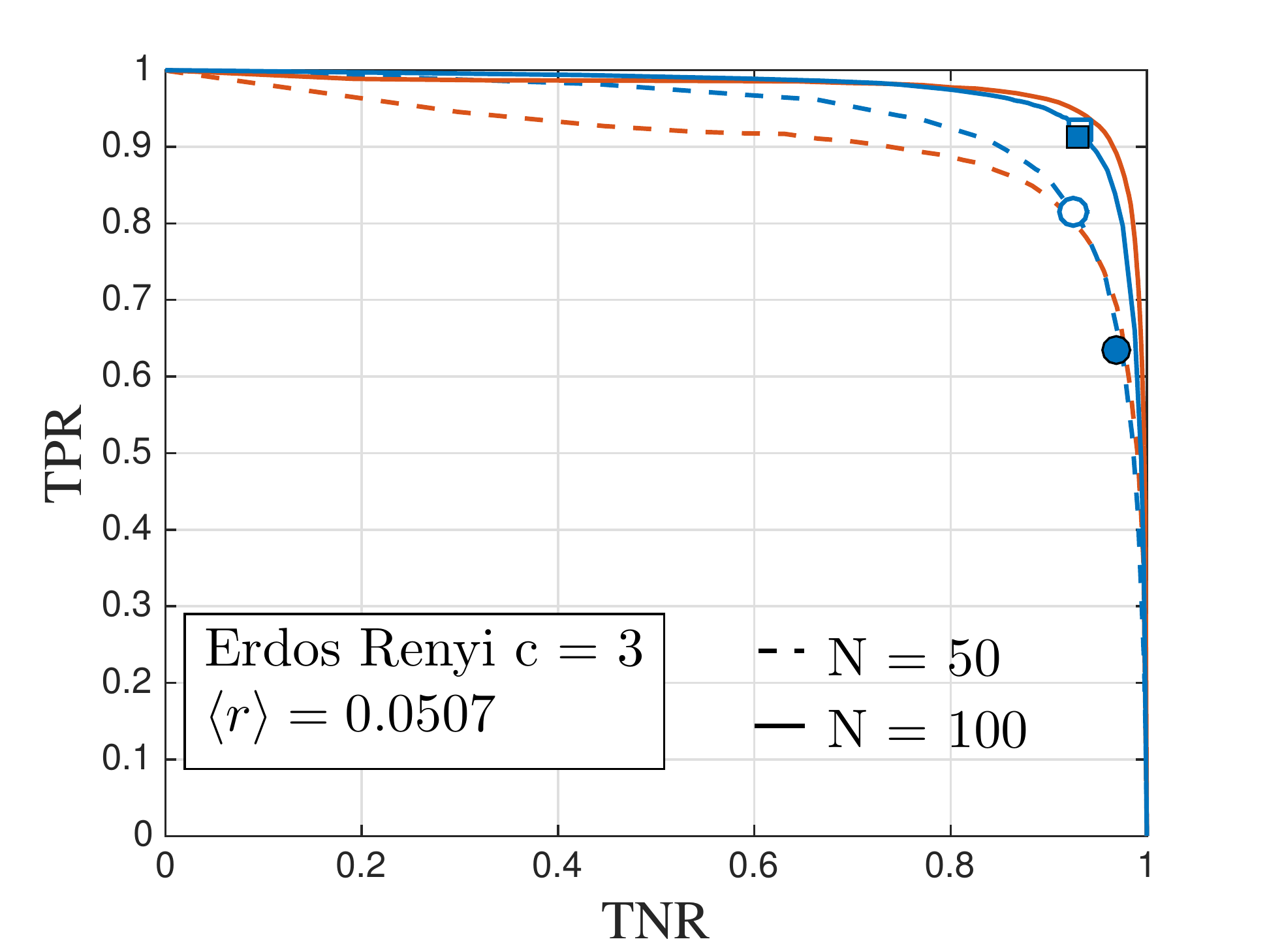}}
\subfigure{\includegraphics[width = 0.45\columnwidth]{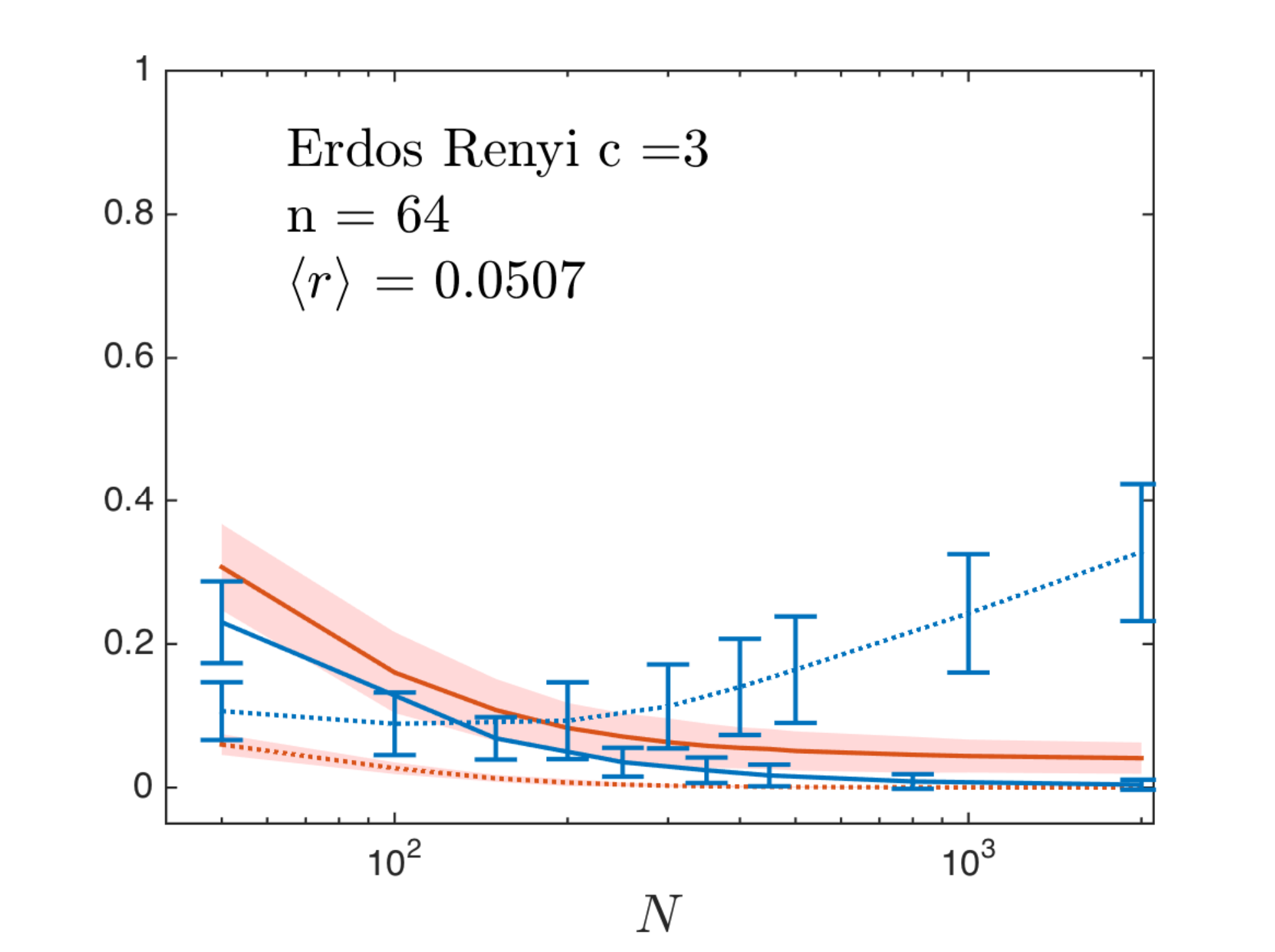}}
\caption{(Left) Comparison between MS (blue lines) and PLM+$\ell_1$ (red lines) for Erd\"os R\'enyi graphs with average degree $c=3$, of $n =64$ nodes. Each point of the curves represents the average of TNR and TPR over one hundred different realisations. The comparison is drawn for two different values of $N$ and in the weak couplings regime ($\beta = 0.5$). The curves for PLM+$\ell_1$ are obtained by varying the regularizer; the ones for MS by varying $\epsilon$. Two different points are highlighted for each curve showing the effects of employing the self-consistent (coloured markers) procedure for selecting the models' prior coefficient, and an {\em ad hoc} $N$-dependent (white markers) procedure $\epsilon (N) = r_g + (1-r_g) \exp{(-N/50)}$, where $r_g$ represents our belief of the sparsity of the network ($r_g = 0.01$ in the figures). \\
(Right) $FPR$ ($= 1- TNR$) with dotted lines and $FNR$ ($= 1-TPR$), solid lines, versus the size of the sample $N$. The reconstruction using MS (blue lines plus error bars) is compared to that using PLM+$\ell_1$ (red lines plus shaded error bars). The thresholding procedure employed for MS is the N-dependent one; the regularizer used for PLM+$\ell_1$ is half of the maximal regularizer for which the inferred network becomes empty. We found that the latter fixed choice of the regularizer optimise PLM+$\ell_1$ results across the investigated topologies and sample sizes.}
\label{fig:scatterMsVsL1}
\end{figure}

Finally, we observe that the choice of the prior is very important, and it gauges the trade off between true positives and true negatives: with larger value of $\epsilon$ the reconstructed network is denser, hence the TPR is larger. Besides this, we observe that the reconstruction becomes more accurate for values of $\epsilon$ that are closer to the true sparsity $r$. This shows that the self-consistent procedure for the choice of the prior is an important ingredient of the MS method. Fig.\ \ref{fig:scatterMsVsL1} (left) shows that indeed the self-consistent procedure selects points close to the top-right corner of the plot. For small sample sizes ($N=50$) MS selects too few bonds, which results in a low TPR. This can be improved by an {\em ad hoc} correction for small sample sizes (see Figure caption). 

Fig.\ \ref{fig:scatterMsVsL1} (left) also reports the comparison of the MS reconstruction with the one obtained with the PLM+$\ell_1$ method for the case of Erd\"os R\'enyi graphs with average degree $c=3$ (see the Appendix for similar results on other topologies and different network sizes). Surprisingly, even for a graph with loops as the one in the Fig.\ \ref{fig:scatterMsVsL1}, the MS algorithm exhibits performance comparable to those of PLM+$\ell_1$, when $N$ is small. In fact, the curves of the two methods in the ROC plot are almost always superimposed and this fact implies an equivalent discriminative power. We found the same results even in the presence of small external fields. As expected, as $N$ becomes large (see Fig.\ \ref{fig:scatterMsVsL1} right), MS produces many false positives even when using the self-consistent procedure for selecting the prior. Yet the FNR stays below that of PLM+$\ell_1$ even for large $N$, and it tends to zero with very small error bars. This fact makes MS interesting as a pruning algorithm for large sparse network and as a pre-treatment procedure for pseudo-likelihood based techniques in order to save computational time and enlarge their domain of application to larger networks. In summary, in the investigated cases of sparse graphs with few loops, MS demonstrated its ability in classifying relevant feature in the deep under-sampling regime as well as one of the best existing algorithm and, moreover, its quality in spotting irrelevant couplings, especially for large $N$. 

Before moving to partially observed graphs, we discuss a simple way to improve the quality of the MS reconstruction, that originates from the excessive number of bonds recovered.

\subsection{\label{subsec: results_condition} Removing indirect interactions by conditioning}

\begin{figure}[t!]
\centering
\includegraphics[width = 0.45\textwidth]{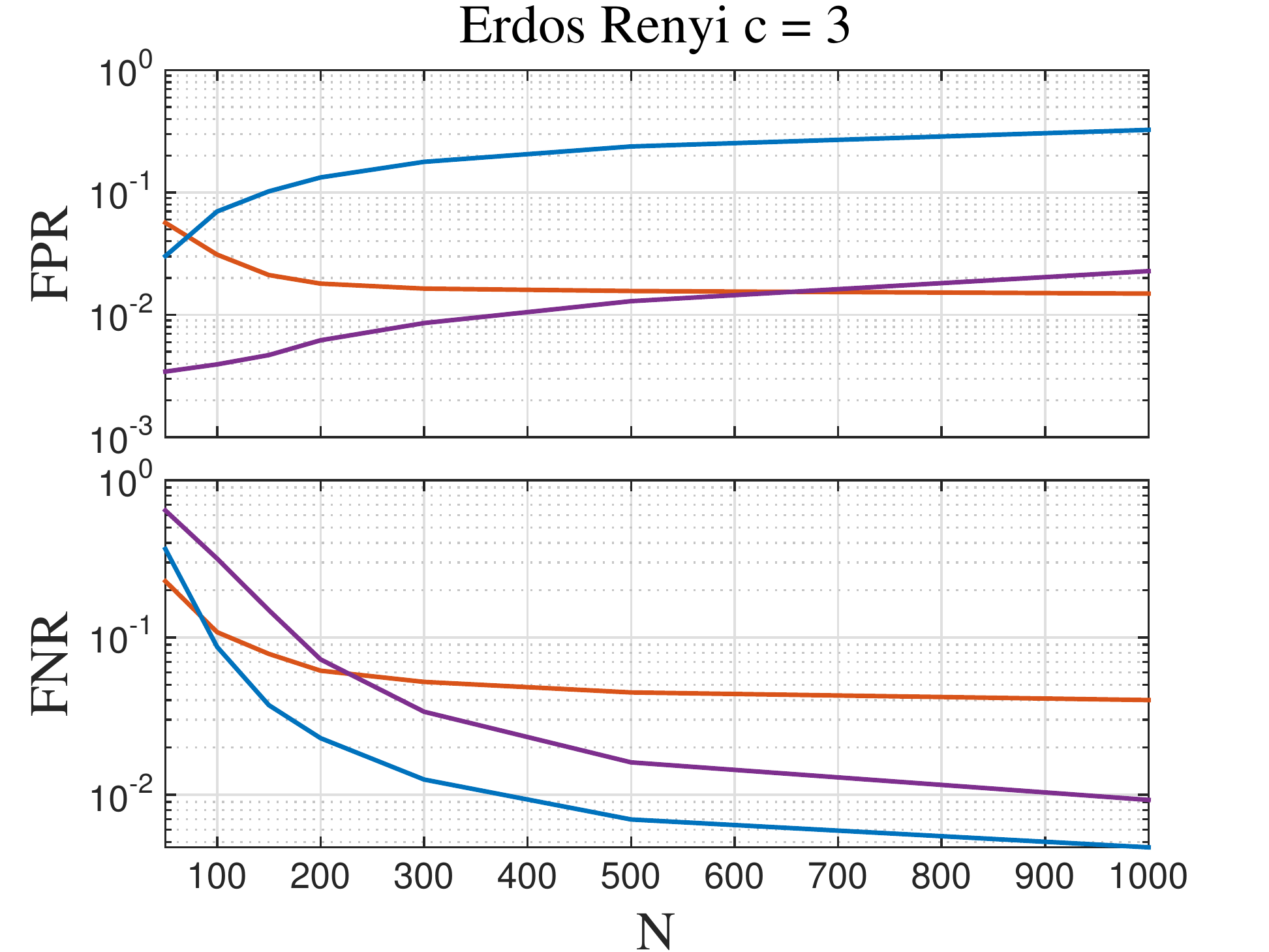}
\includegraphics[width = 0.45\textwidth]{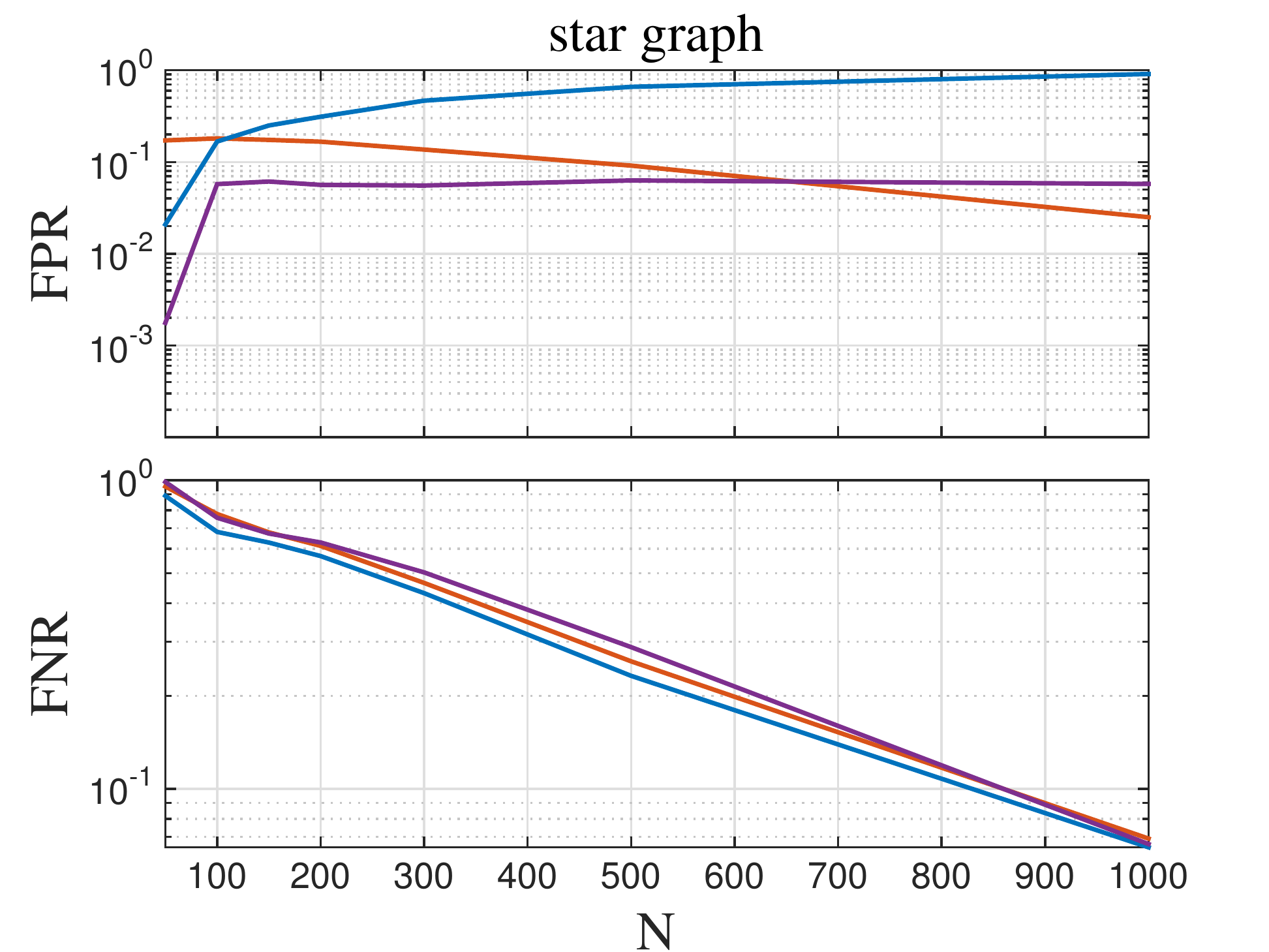}
\caption{FNR and FPR for network recovery from data generated from simulations of an Ising model with $n=64$ spins and $\beta=0.5$ on Erd\"os R\'enyi c = 3 (left) and star (right) topologies. The performances of MS (blue) and PLM+$\ell_1$ (red) are compared with the corrected MS algorithm described in section \ref{subsec: results_condition} (violet, see text and Appendix).}
\label{fig:ho_simulations}
\end{figure}

The excess of false positives in the interaction graph recovered by MS can be cured, at least partially, by considering model selection on larger graphs. An even simpler recipe, that requires a minimal additional complexity, is that of re-running the algorithm conditioning on the value of a spin or of a subset of them. Let us take the simple example of three spins, where all the bonds between spins $1,2$ and $3$ have been recovered by MS. In order to ascertain whether the interaction between $1$ and $2$ is genuine, one can condition on the value of $S_3$ and perform the inference, separately, on the sample where $S_3=+1$ and $S_3=-1$. This produces two predictions $\tilde{\eta}_{1,2}^+$ and $\tilde{\eta}_{1,2}^-$, for the difference between the probabilities of having and not having a bond between spins $1$ and $2$. If the quantity $\tilde{\eta}_{1,2 | 3} = \tilde{\eta}_{1,2}^{+}\ \nu(S_3 = 1) + \tilde{\eta}_{1,2}^{-}\ \nu(S_3 = -1)$, where $\nu(S_3 = \pm1)$ represents the fraction of times $S_3 = \pm1$ in the dataset, is greater than zero then the interaction between $S_1$ and $S_2$ is genuine and is not induced by $S_3$. Furthermore, one can extend this argument to all the other spins $S_k$ that interact with both $S_1$ and $S_2$ in the recovered graph, and check whether the interaction between $S_1$ and $S_2$ is fictitiously induced by some other spin. There are different ways in which the effect of different spins can be combined to arrive at a prediction on whether the bond between $S_1$ and $S_2$ exists. We refer to the Appendix for a detailed discussion. Here we just report the result for the conservative case where we take the minimal value of $ \tilde{\eta}_{1,2 | k}$ over all the common neighbours $S_k$ of $S_1$ and $S_2$ on the recovered graph.
Fig.\ \ref{fig:ho_simulations} shows the performance of the corrected reconstruction for the case of an Erd\"os R\'enyi graph and the hard case of the star graph, i.e. a graph built on one node to which all the others are connected, which is a particularly hard situation for MS: each pair of not directly connected nodes are conditionally independent given the node in the centre of the star. Therefore typically MS returns a fully connected matrix with many false positives. It is interesting to notice that also PLM+$\ell_1$ is quite inaccurate when dealing with stars and only for large sample sizes $N$ one recovers satisfactory results. In this regime and in general for all the investigated topologies, we observed that the corrections explained above allow for lowering the false positive rate to at least one order of magnitude while keeping the false negative rate almost always below the PLM+$\ell_1$ curve. Even for hard cases this recipe is able to substantially improve MS performance in the large $N$ regime, as is shown in Fig.\ \ref{fig:ho_simulations} (right panel) for the case of a star graph (see Appendix for similar results on other topologies). 
In summary this simple correction allow for an extension of the domain of applicability of our method to hard classes of sparse topologies and to larger samples' length, where usually ordinary MS returns many false positive, reaching results comparable to those of PLM+$\ell_1$.

Finally, it is worth to stress that this correction requires almost the same computational effort as the original algorithm: in the first case given the length of the sample $N$ one needs to evaluate only one classifier (i.e. the borders of the partitions of Fig.\ \ref{fig:BondNoBond}); when adding this correction, the number of classifiers required grows with the number of loops of 3 nodes found in the recovered graph and with the polarisation of the nodes involved in those structure. In fact if these nodes are all $1$ or $-1$ roughly half of the times then again approximately only one classifier is needed.  

\subsection{\label{subsec: results_partial} Partially observed graphs}

\begin{figure}[t!]
\centering
\includegraphics[width = 0.45\textwidth]{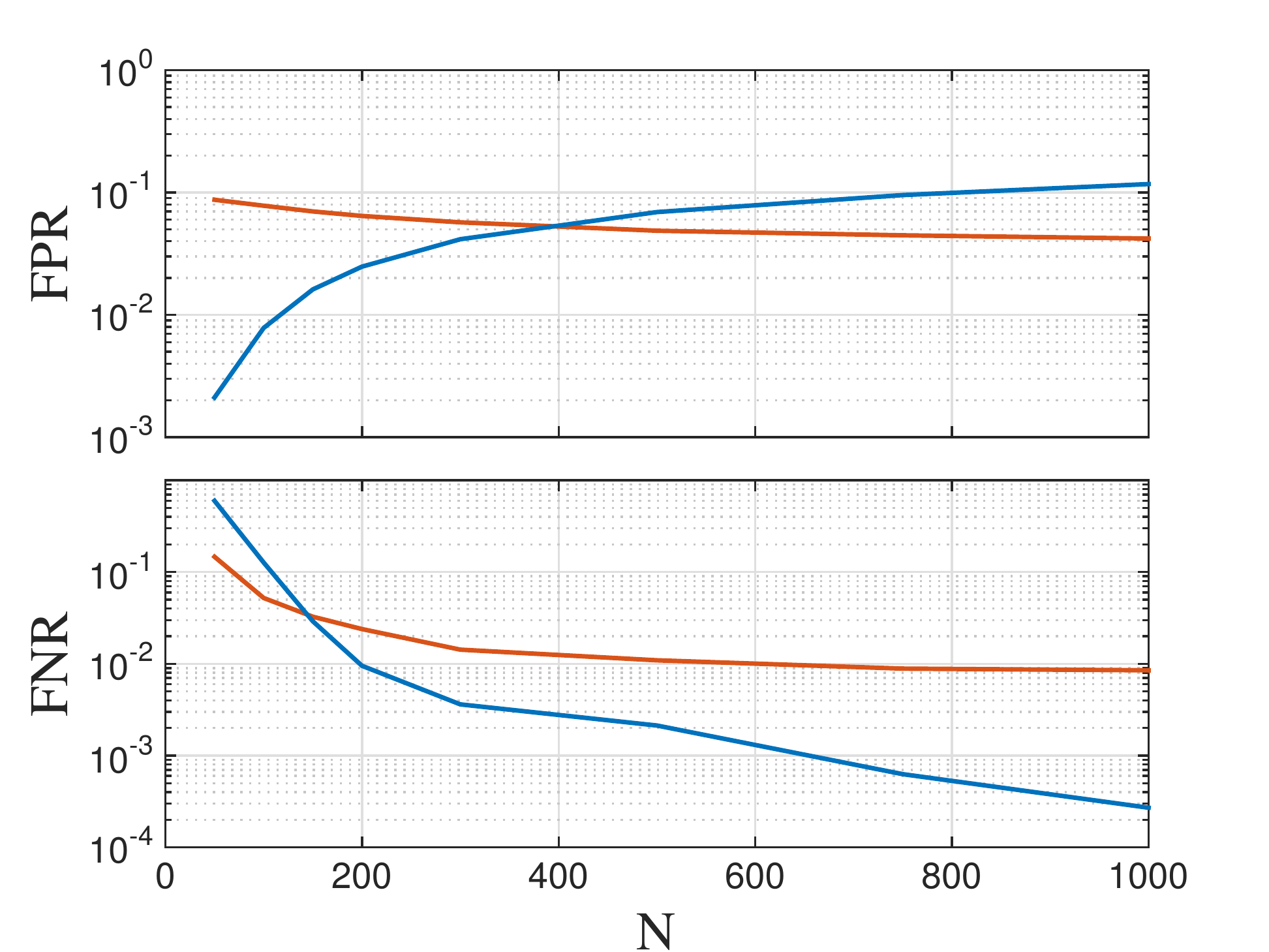}
\includegraphics[width = 0.45\textwidth]{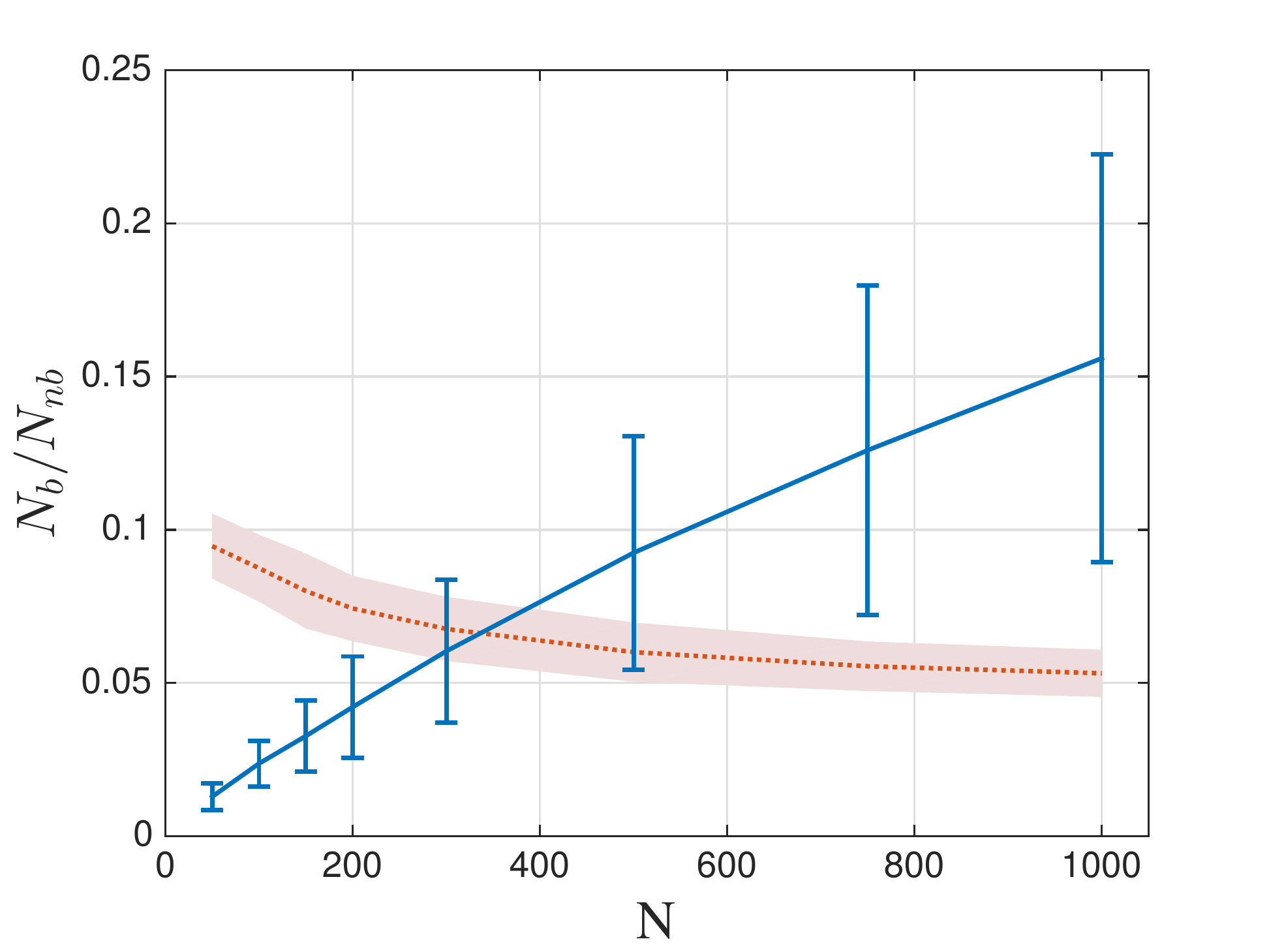}
\caption{FNR and FPR for network recovery from data generated from simulations of an Ising model with $\beta=0.5$ on a Erd\"os R\'enyi graph with $c=3$ and $250$ nodes. The data contains observations of the spins on $n=64$ randomly chosen nodes. Each point on the curves in the figures represents the average over one hundred different realisations of the same topology and the standard deviations are depicted with (shaded) error bars. The performances of MS (blue) are compared to  the PLM+$\ell_1$ (red) algorithm: (left) false positive rate (FPR), false negative rate (FNR) and (right) sparsity of the reconstructed graphs versus the length of the data sample. The value of the regularizer used is the same as the case of a fully observed graph.}
\label{fig:part_obs}
\end{figure}

Let us now discuss the case where the data contain only a partial observation of the nodes of the graph. For example, Fig. \ref{fig:part_obs} shows the results of the inference on a Erd\"os R\'enyi random graph of $250$ nodes where only $n=64$ of them are observed (see caption for details). From the figure it is clear that PLM+$\ell_1$ looses accuracy with respect to MS in the case of a partial observed network. In fact, PLM+$\ell_1$ returns more false positives than MS for small $N$, at odds with the case of fully observed graphs. Even for large $N$, the number of false positives remains high with respect to the fully observed case. On the other hand, the number of false negatives is, as usual, always bigger than the one resulting from MS algorithm, except for very small values of $N$ where, as previously discussed, the self consistent approach turns out to be extremely selective. However, this value can be reduced by employing an {\em ad hoc} correction for small sample sizes (see caption in Fig.\ \ref{fig:scatterMsVsL1}). As a consequence, for small $N$, the graph recovered by PLM+$\ell_1$ is always denser than the correspondent one with MS whereas in the fully observed case the opposite is true. This tendency of overestimating the number of bonds of the graph 
is due to the fact that PLM+$\ell_1$ tries to explain accurately all observed statistical dependences in the data which are noisy and often mediated by hidden variables. As a consequence, it tends to assign a bond, and therefore give the same distance on the graph, to each of the identified statistical dependencies, failing in discerning between direct and indirect interactions. This is particularly interesting for real data applications, where often the observations of the system under investigation are few and concern only a small portion of it. In fact, as we will see in the following sections when dealing with real data in this regime, networks recovered with PLM+$\ell_1$ will always contain more bonds.

\section{\label{sec: real_data} Application to real data}

In this section we apply our method (MS) to real data and we compare its predictions with those from PLM+$\ell_1$. We discuss a dataset of financial returns of the stocks in the Dow Jones index and a dataset on the neural activity of cells in the medial entorhinal cortex of a moving rat. In both cases, the statistics of the data is likely non-stationary, so that inference on the whole dataset may lead to confounding effects. The virtue of the MS method is that it is ideally suited to very small samples, and hence it allowed us to study the dynamics in these datasets on smaller windows of time for financial data, and space for the neural data, thereby revealing genuine statistical dependencies. 

\subsection{\label{sec: fin_data} US stock market data}

In this section we apply our method (MS) to real data and as usual compare predictions with those from PLM+$\ell_1$. The dataset is the same as the one studied in \cite{ponsot}, which was taken from Yahoo Finance and consists of the returns (i.e. the logarithm of the ratio between the closing price and the opening one) for 41 stocks in the Dow Jones index from 1st Jan 1980 to 25 June 2007 for each trading day. In order to subtract the overall market trend and focus on the correlations between the stocks, the average values of the returns for each day has been subtracted from the values of the returns. 
Finally we took the sign of the resulting returns in order to get configurations of 41 spins, i.e. $ S_i(t) \;\mbox{with}\; i = 1,..,41$ for each day $t$. This is expected to remove, at least partly, long-term auto-correlation effects in the (absolute) size of returns -- the so-called volatility  \cite{bouchaud,ponsot}.
The list of stocks taken into account along with the industrial sector are listed in Tab.\ \ref{table:stocks} in Sec.\ \ref{sec: SM2} of the Appendix.

\begin{figure}[th]
\centering
\includegraphics[width=1.1\columnwidth] {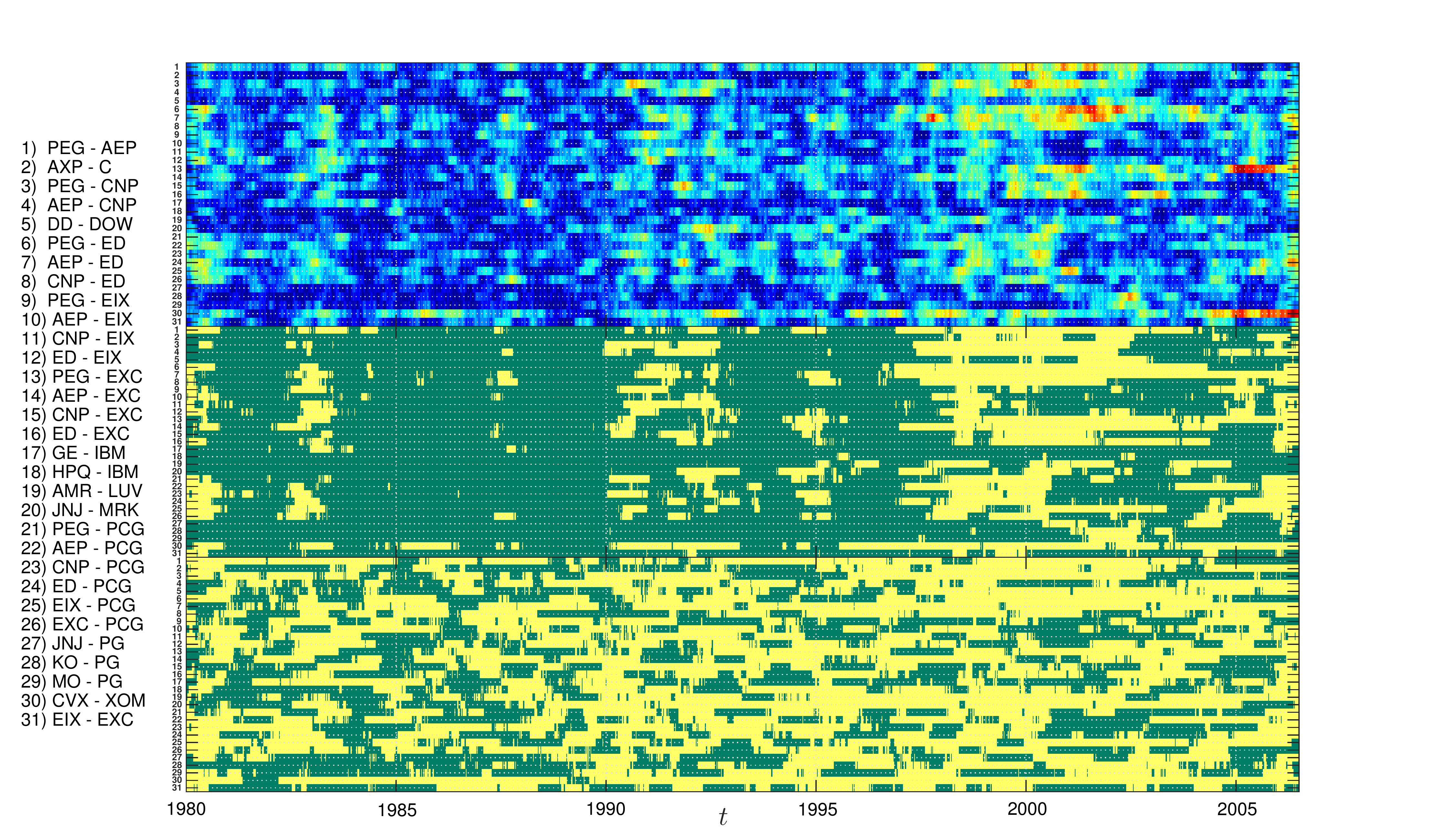}
\caption{Connections between the stocks belonging to the investigated clique from  the 1st of January 1980 to the 25th of June 2007 obtained with MS (middle) and with PLM+$\ell_1$ (bottom panel). For comparison, the absolute values of the correlations $|C^N_{ij}(t)|$ between the same stocks, during the same period, is also shown (top panel). Colour maps: (top) very small correlations are in blue and very high ones in red; (middle and bottom) green corresponds to no-bond and yellow means a bond. MS results have been obtained with a self-consistent threshold whereas PLM+$\ell_1$ assumes a regularizer that is half of the maximal regularizer for which the inferred network becomes empty. For such a small samples and across the considered graph, the latter recipe for the regularizer was found to work better on synthetic data than the one given in \cite{ravikuumar10}.}.
\label{fig:reconstruction_clique}
\end{figure}

As a preliminary step, we check that auto-correlation 
\[
c^N_{ij}(t,\tau) = \frac{1}{N} \sum_{t' = t}^{t+N-1} S_i(t')S_j(t'+\tau)
\]
are statistically insignificant for $\tau\neq 0$ (see Appendix). This is consistent with the Efficient Market Hypothesis, that states that 
current prices fully reflect all available information, and hence future returns cannot be predicted \cite{bouchaud}. Yet, the time series of returns is far from stationary, as evidenced by the plot in Fig.\ \ref{fig:reconstruction_clique} of the equal time connected correlation 
\[
C^N_{ij}(t) = \frac{1}{N} \sum_{t' = t}^{t+N-1} S_i(t')S_j(t') - m^N_i(t)m^N_j(t)
\]
with $m^N_{i}(t) = \frac{1}{N} \sum_{t' = t}^{t+N-1} S_i(t')$, between a selected set of pairs of stocks (see later). This shows periods of high correlations interspersed with stretches where stocks dependencies are weaker. In situations like this one, any effort of representing the evolution of the statistical dependence by an inferred interaction network, faces the unavoidable limit that the sample size $N$ cannot be arbitrarily large. Inference on the whole dataset would mix different market phases. On the contrary, it is reasonable to assume the market to be ``locally'' in equilibrium at a given time $t$, if $N$ is taken to be small enough. From our analysis (see Appendix) $N=200$ is large enough to afford statistically stable results, that are consistent with results obtained with smaller values of $N$, but that capture the non-stationary evolution of market correlations. 

We performed inference of the underlying interaction network using MS with a fixed and self-consistent threshold and with PLM+$\ell_1$, on rolling windows of $N$ days. 
Here we present the main results for $N=200$, that correspond roughly to 10 months, and refer to the Appendix for a detailed derivation. 

\begin{figure}[t!]
\centering
\includegraphics[width=0.45\columnwidth] {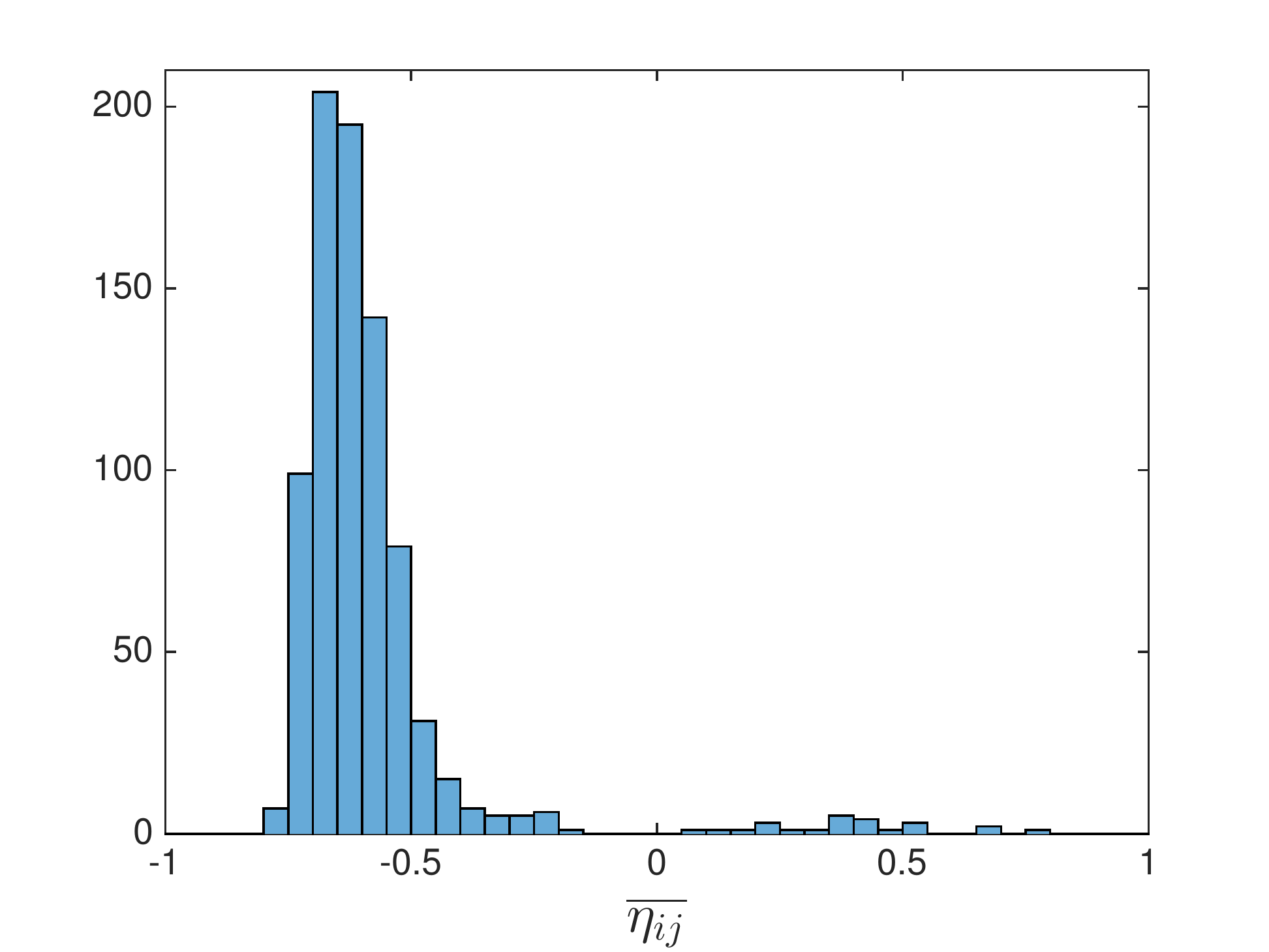}
\includegraphics[width=0.45\columnwidth] {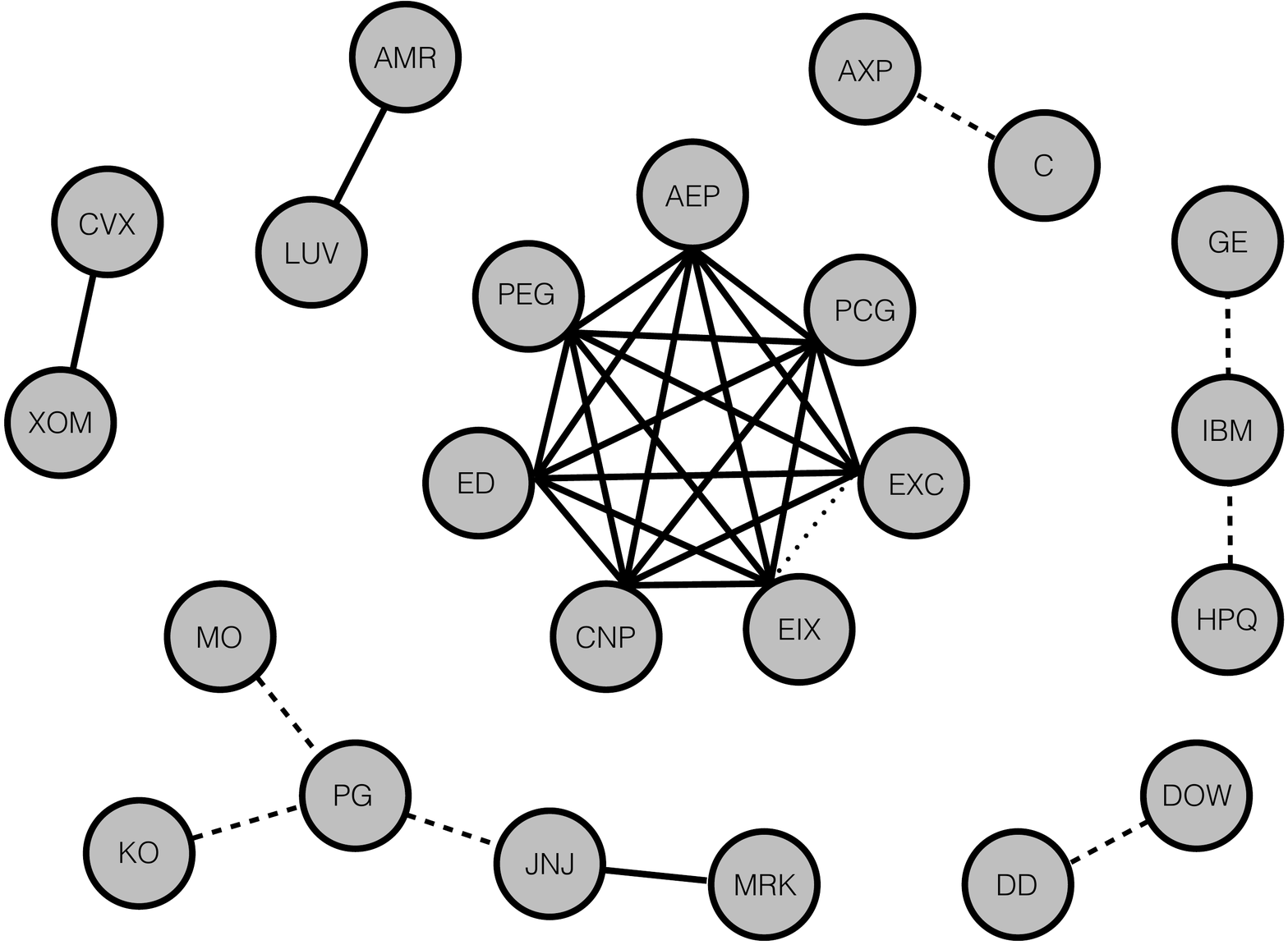}
\caption{Histogram of the averaged confidence $\overline{\eta_{ij}}$ (left) and resulting equilibrium graph (right, thick links). Dashed links refer to additional links inferred by the PLM+$\ell_1$ method, whereas the dotted one between EIX and EXC indicates the only connection not recovered by PLM+$\ell_1$. The names of the stocks present in this graph are listed in Table \ref{table:stocks} and highlighted with different colours: one for each connected component recovered by MS.}.
\label{fig:supportMS}
\end{figure}

For each pair $i<j$, we compute the confidences $\eta_{ij}(t)$ of the MS inference on windows of $N=200$ days and their average $\overline{\eta_{ij}}$ over $t$. Fig.\ \ref{fig:supportMS} (left) shows that the links separate nicely into a group with $\overline{\eta_{ij}}>0$ that correspond to those pairs that are often connected by a bond in the reconstructed network, and those with $\overline{\eta_{ij}}<0$ that are hardly or never connected. This implies a persistent underlying network structure, which is shown in Fig.\ \ref{fig:supportMS} (right, thick bonds). 
Inspection of Table \ref{table:stocks} reveals that all the components highlighted correspond to stocks in the same economic sector: a large clique groups stocks in the utilities sector (PEG, AEP, CNP, ED, EIN and PCG), whereas other isolated bonds connect stocks in the Airline industry (AMR, LUV), Oil (CVX, XOM) and Healthcare (MRK, JNJ). 

Likewise, we infer the couplings $J_{ij}(t)$ on the same data, with PLM+$\ell_1$ and compute the average $\overline{J_{ij}}$ over $t$. The resulting histogram exhibits a long tail for positive values of $\overline{J_{ij}}$ (see Appendix) that can be distinguished from a noisy bulk by setting a threshold. The links corresponding to values $\overline{J_{ij}}>0.02$ are shown in Fig.\ \ref{fig:supportMS} (solid and dashed lines). In addition to the bonds retrieved with MS, this PLM+$\ell_1$ procedure also retrieves other bonds that connect the Healthcare cluster to stocks in the Consumer Goods sector (KO, MO, PG) and it identifies three additional components, in the Chemicals (DD, DOW) and Financial (AXP, C) sectors, and one that mixes General Electrics (GE) with Technology firms (IBM, HPQ).

The full power of our approach, however, lies in its ability to fully account for the non-stationary nature of the correlations. This is evidenced by Fig.\ \ref{fig:reconstruction_clique}, that reports, besides the value of the connected correlations (top panel), the results of network inference with MS (middle panel) and PLM+$\ell_1$ (bottom panel). Each row corresponds to one of the $31$ pairs of links identified in Fig.\ \ref{fig:supportMS} and presence or absence of a bond at time $t$ is indicated by a bright to dark spot. Visual inspection reveals that MS accounts much better for the non-stationary nature of correlations than PLM+$\ell_1$. 

\begin{figure}[t!]
\centering
\includegraphics[width = 0.49\columnwidth]{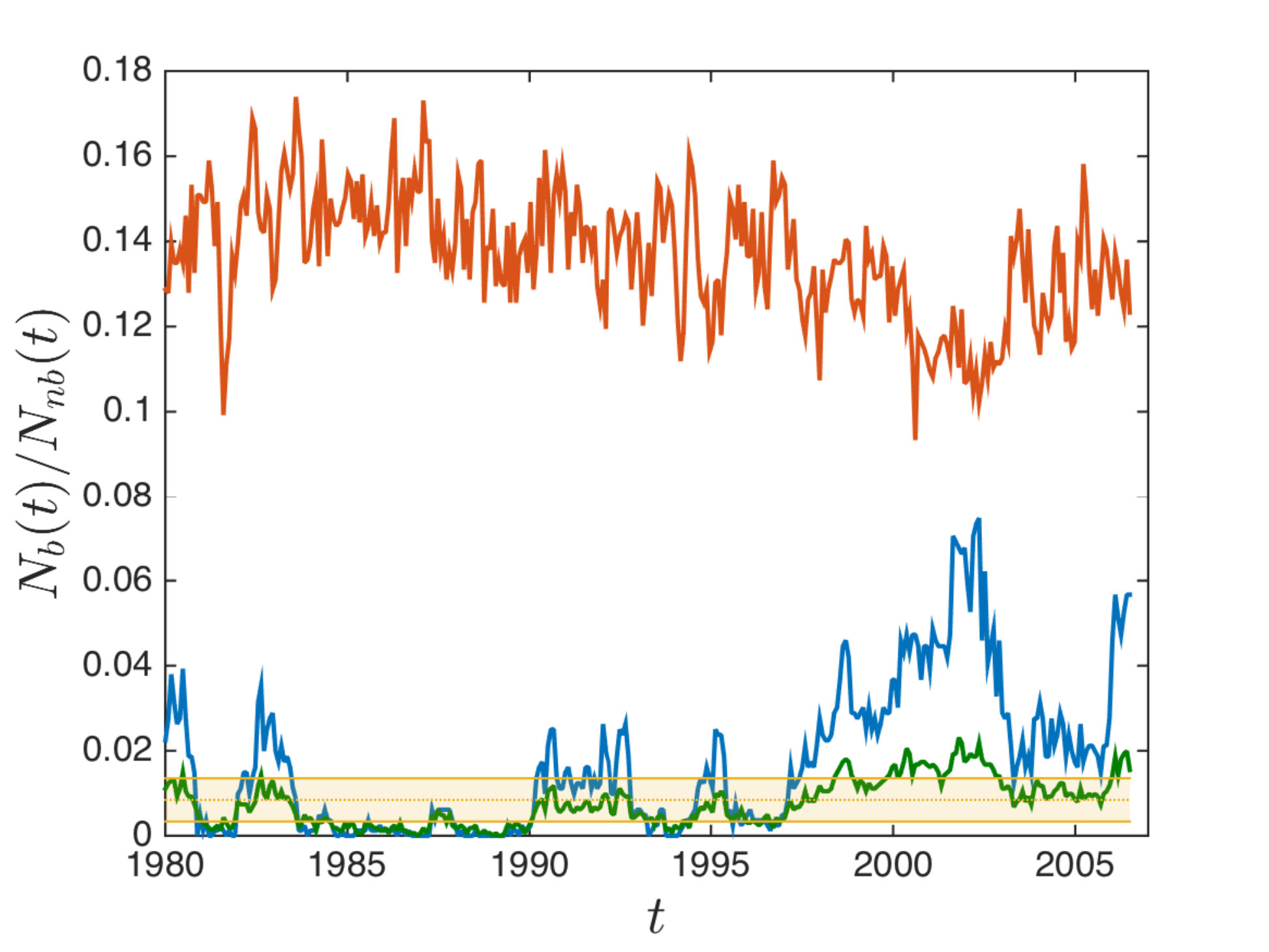}
\includegraphics[width = 0.49\columnwidth]{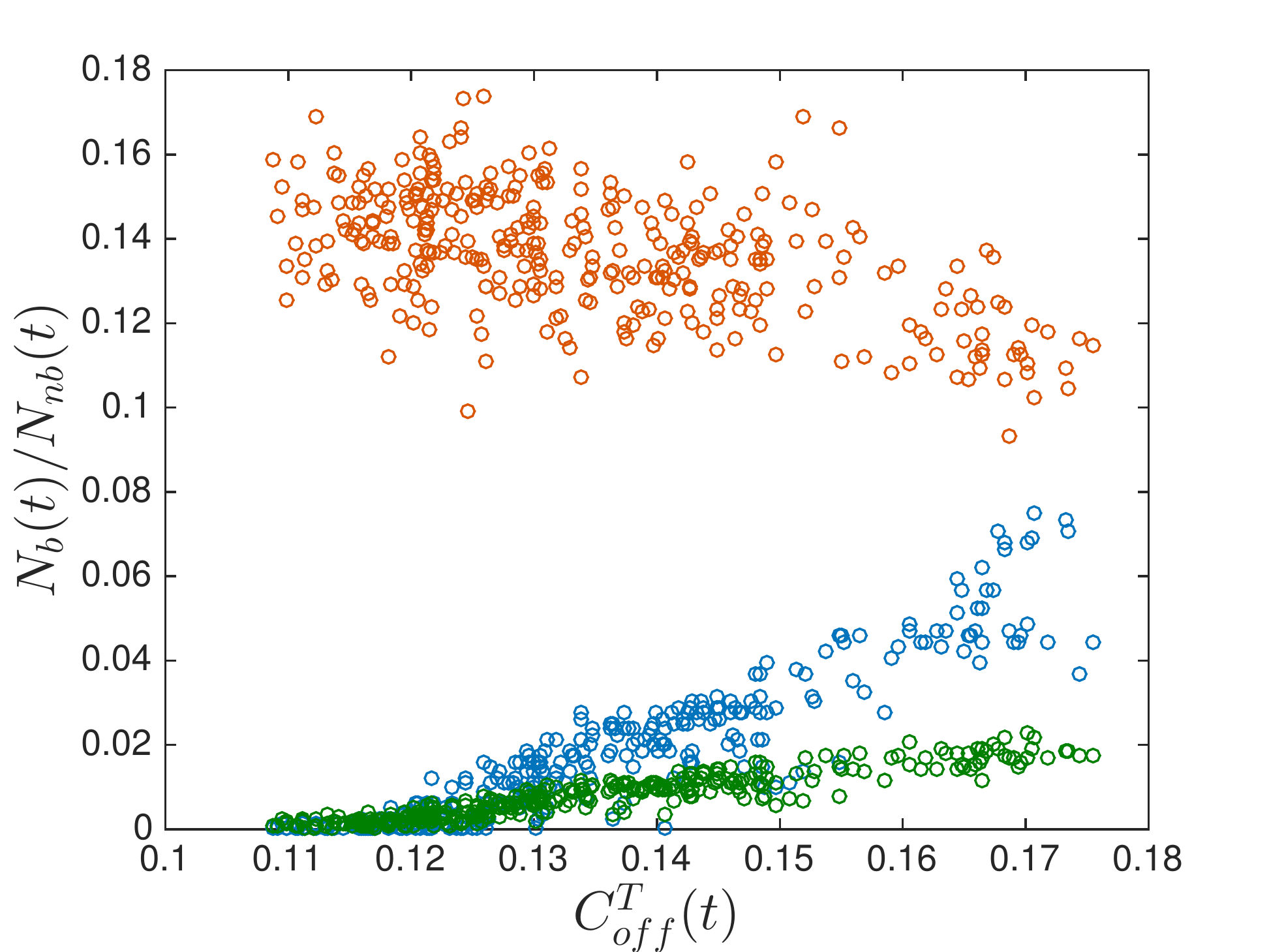}
\caption{Left: Comparison between the recovered `bond'-`no-bond' ratio with MS assuming a self consistent approach (blue), a fixed threshold (green) and with PLM+$\ell_1$ (red). 
The yellow shaded band corresponds to results obtained by MS with a self consistent approach on a reshuffled dataset (mean $\pm$ standard deviation).
Right: Scatter plot of the sparsity for MS with self-consistent (blue) and fixed (green) priors, and for PLM+$\ell_1$ (red) versus the off diagonal r.m.s. values of the connected correlations ($C_{\rm off}(t)$)}. 
\label{fig:sparsityMS}
\end{figure}

More precisely, Fig.\ \ref{fig:sparsityMS} (left) shows the time evolution of the sparsity of the network inferred with MS and with PLM+$\ell_1$. While with the former, the rises and falls of the network are rather sharp, for PLM+$\ell_1$ the density of the network appears to be higher and
much smoother. We also run MS on a reshuffled dataset, where each time $N$ configurations $\vec S(t)$ are chosen at random. Fig.\ \ref{fig:sparsityMS} (left) also reports the average and typical fluctuations that one would expect from MS on these system sizes. Clearly, density fluctuations in the true dataset are much more pronounced than what would be expected in the shuffled one. This also shows that the average density of bonds in MS is smaller than what PLM+$\ell_1$ would predict.

The right panel of Fig.\ \ref{fig:sparsityMS} relates the behaviour of the network to the dynamics of the correlations. As a proxy of the latter we take the root mean square value $C_{\rm off}(t)$ of the connected correlations $C^N_{ij}(t)$ ($i<j$) and compare it to the sparsity of the inferred networks at the same time. While the network inferred by MS follows closely the dynamics of correlations, that inferred by PLM+$\ell_1$ does not. Fig.\ \ref{fig:sparsityMS} also shows the effect of the self-consistent prior in making the MS algorithm more sensible to non-stationary effects, with respect to MS with a fixed prior.

It is important to stress that, in spite of these differences, MS produces results that are highly consistent with those of PLM+$\ell_1$. Indeed 94\% of the bonds identified by MS are also found by PLM+$\ell_1$. This confirms, as could be anticipated by its vary nature, that MS delivers rather conservative predictions on network inference.

Still, a sparser and more dynamic network of interaction does not {\em per se} makes of MS a method which is superior to PLM+$\ell_1$. A more stringent test is possible by analysing the ability of the reconstructed network to describe data out-of-sample. In practice, we estimate the network of interactions and the values of parameters in a window of $N$ data points (training sample) and evaluate the likelihood of the data in the subsequent window of $N$ data points (test sample). For completeness, we also compare the results with a sample of $N$ data points randomly generated (random sample). 
In order to avoid problems of non stationarity, we focus on windows of $N=50$ and $N=100$ points. In this way, the train and the test samples together fills a time window smaller than the assumed stationary time scale.

\begin{figure}[t]
\centering
\includegraphics[width = 0.48\columnwidth]{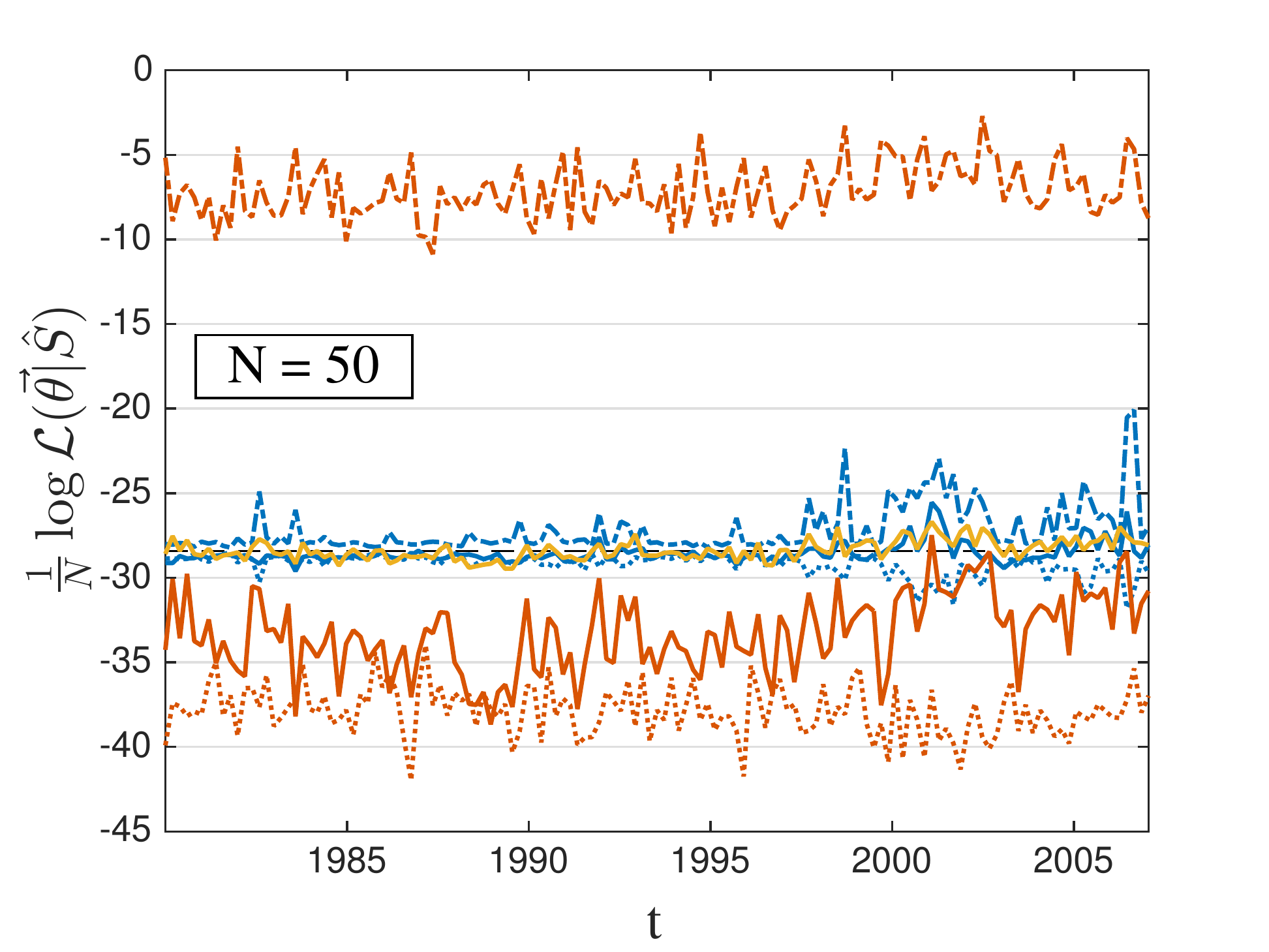}
\includegraphics[width = 0.48\columnwidth]{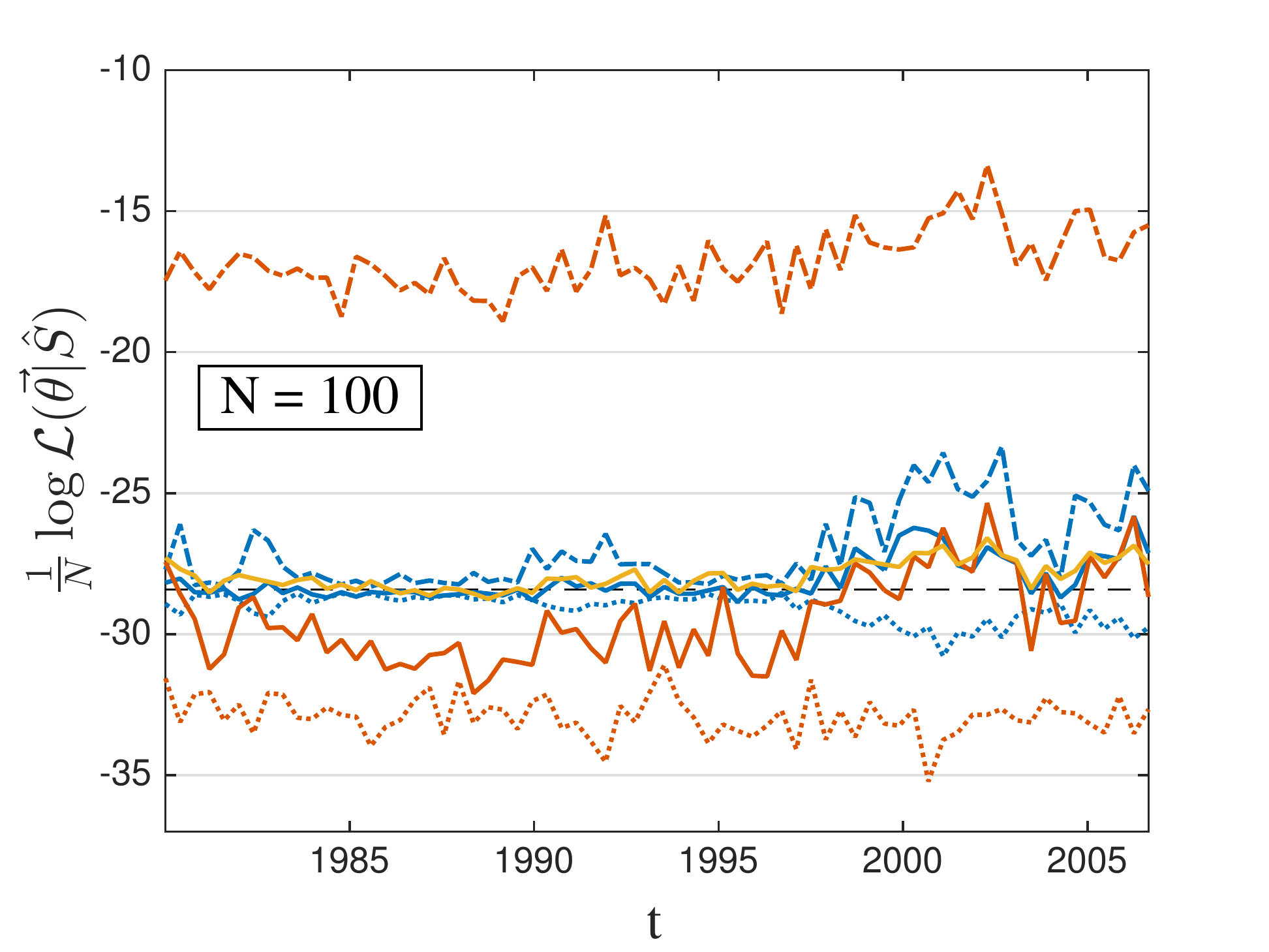}
\caption{Normalised log-likelihood, $\frac{1}{N}\log{\cal L}(\vec{\theta} | \hat{S})$, on the training (dash-dot lines), test (solid lines) and random (dotted lines) samples as a function of time $t$, for $N=50$ (left) and $N=100$ (right). Blue lines refer to networks inferred on the training sample with the MS method with the self consistent prior, whereas the red lines refer to PLM+$\ell_1$ inference. As a reference, the normalised log-likelihood of a random spin model with no parameters is also reported as a black dashed line. Finally yellow solid lines indicate the likelihoods on test samples evaluated using directly the couplings inferred with PLM+$\ell_1$.}
\label{fig:likelihood_test}
\end{figure}

For each $t$ which is a multiple of $N$, we recover the MS (blue lines) and the PLM+$\ell_1$ (red lines) networks on the training sample, estimate the non zero parameters using an {\itshape expectation consistent} inference method following \cite{Opper05} and evaluate the likelihood of the training (dash-dot lines), test (solid lines) and random (dotted lines) samples, for both networks. The resulting values of the likelihood are shown in Fig.\ \ref{fig:likelihood_test} over the whole timespan of the dataset. 

The maximum likelihood values on the training samples (dash-dot lines) are always above the others, as expected, since they corresponds to the maximum likelihood values on the training samples. In addition, since PLM+$\ell_1$ always recovers a more complex model (denser graph), it fits better the in-sample data and consequently the correspondent likelihood of the training set is larger than the MS one. The situation is reversed, however, when analysing the values of the likelihood in the test samples (solid lines). There the MS method achieves a larger likelihood with respect to the PLM+$\ell_1$ method, both for $N=50$ and for $N=100$. The difference in likelihood is smaller in periods of strong correlations (e.g. after 2000) where the MS network is denser.  
In addition, the difference between the likelihoods of the training and the test samples are much smaller for MS than for PLM+$\ell_1$. 

The likelihoods of the random sample (dotted lines) are smaller than those of the test set, but while the difference is negligible for MS it is considerable for PLM+$\ell_1$. This is related to the fact that the sample is also reasonably well described by a model of random spins, where $P(\vec S)=2^{-n}$ does not depend on any parameter. The corresponding normalised log-likelihood ($-n\log 2$) is shown as a black dashed line in Fig.\ \ref{fig:likelihood_test} and it lies above the likelihood of the test samples of MS and PLM+$\ell_1$ for most of the time. 

When using directly the couplings inferred with PLM+$\ell_1$ for estimating the likelihood on test data  (yellow solid lines), the outcomes become comparable to those obtained with MS (blue solid lines).  However, this is accompanied by a decrease in the training likelihoods that now assume values close to the corresponding ones from MS. This result is a consequence of the fact that the couplings inferred using PLM+$\ell_1$ (i.e. the symmetrised version of them: $[J_{ij}+J_{ji}]/2$) are penalised by the regularizer and are, therefore, smaller than those obtained via optimising the likelihood over graph recovered using the training sample. In particular, the couplings corresponding to no-bonds in MS graphs are significantly smaller and this makes the results close to those of a null model and to MS. Moreover, for such a small samples, the asymmetry in the couplings may assume non-negligible values such that a symmetrization actually mixes two noticeably different estimates. Given the issues caused by the regularizer and the symmetrization procedure, we feel that a fair likelihood based comparison between two algorithm for graph selection should invoke the same method for the couplings' estimation step, which is what we discussed in the previous paragraphs, namely using a maximum likelihood estimate for non-zero couplings for both algorithms.

In summary, the very sparse topologies recovered by MS represent better the US stock market interactions with respect to the ones of PLM+$\ell_1$. The MS topologies are very similar to the ones obtained with an independent spin or zero-parameters model and they have very similar out-of-sample likelihoods. This suggests that the effects of interactions need to be invoked only sporadically and especially in the last years (around 2000-2007) which are characterised by a correlated dynamics. This shows that MS is a quite sensitive inference method in situations where sample sizes are small (e.g. for non-stationary effects) and the data is quite noisy.

\subsection{\label{sec: neuro_data} Neural data}

The dataset is the same as the one studied in \cite{Maria15}. It consists of the recording of the activity of 65 neurons from the entorhinal cortex of a rat moving in a $1.5 \times 1.5$ meters box. The position of the head of the rat is also recorded (see \cite{Stensola} for experimental details). A detailed analysis \cite{Stensola} reveals that 27 of the recorded cells are grid cells, 5 of them are interneurons and the remaining ones are of unknown type. 

Here we use the Model Selection (MS) method to address the issue of inferring the interactions between these cells. It has long been observed that inference of interactions between neurons is a particularly difficult problem (see e.g. \cite{Capone15} for a discussion). In fact, the recorded neurons are sampled from a large pool of different kinds of neurons and other cells interacting with characteristic time scales through directed connections. Given the experimental limitations and the scarce amount of data, resolving directed synaptic connections remains an hard task. Therefore, our approach, as well as all Inverse Ising inference methods, aims at detecting {\itshape effective connections}, i.e. strong statistical dependencies between neurons which can also arise as the result of indirect connections mediated by unobserved components of the network. 
    
    One relevant issue, for this dataset, is that neural activity depends on the position of the rat. If inference is carried out using data recorded when the rat is in different positions, the spatial dependence of the neural activity induces correlations that result in effective couplings between neurons that do not reflect genuine statistical dependence. If inference can be performed controlling for the position of the rat, these effects can be avoided and genuine (hopefully) direct statistical dependencies can be revealed. For this reason, we divide the dataset in many parts, each corresponding to a different spatial position. As in Ref. \cite{Maria15}, we divided the box in a grid of $20 \times 20$  cells, divided the database correspondingly and performed the analysis on the recording in each cell separately. This again brings us to an inference problem where the number of samples is very limited and for which we expect a sparse interaction network. We expect the network to be sparse also because the dataset documents the activity of only a fraction of the cells that are actually involved in spatial cognition and navigation tasks. In other words, this is a problem where inference has to be performed in the presence of many missing degrees of freedom. The effect of these hidden variables is to make our dataset very noisy. In this situation, a model selection approach is necessary in order to correctly evince how much structure can be inferred from the data. Hence the MS approach is a valuable alternative to other approaches \cite{Ben13,Claudia15} to inference in the presence of hidden variables.
    
\begin{figure}[t]
\centering
\includegraphics[width = \columnwidth]{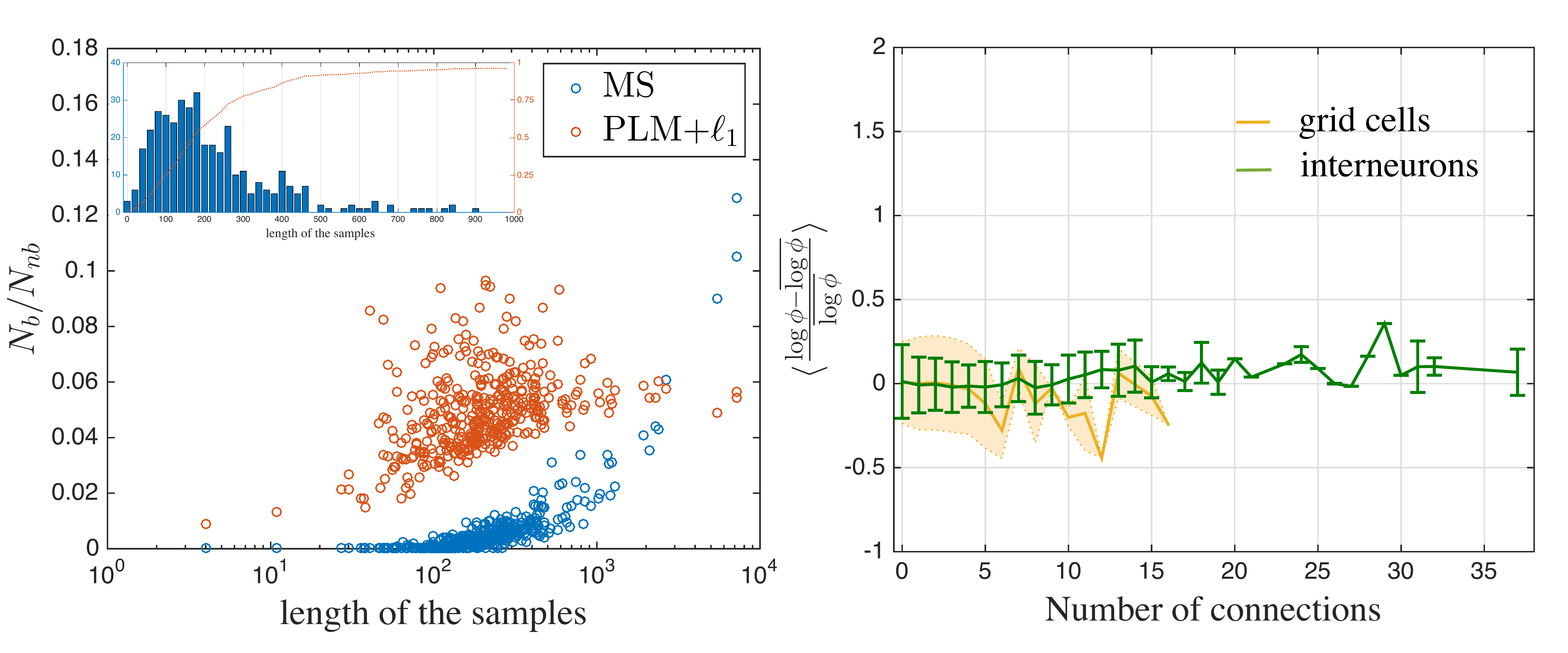}
\caption{(Left) Bond-to-no-bond ratio for inferred graphs with MS (blue circles) and PLM+$\ell_1$ (red circles) versus the length of the samples (left panel). The time bin employed for the figure is 10 ms but the same results are found also for different choices. The inset in the figure shows the distribution of the length of samples and the relative cumulative function (right y-axis). \\
(Right) We report the logarithm of the firing rate, $\log\phi$, for a given neuron normalised with its mean over cells, $\overline{\log\phi}$, and averaged over cells and neurons belonging to the same class (i.e. interneurons and grid cells) conditioned upon the number of connections (solid lines in the figure on the right panel). The standard deviations are depicted with error bars for interneurons and shaded error bars for grid cells.}
\label{fig:gridcell1}
\end{figure}

    We limit our discussion here to the typical properties of our algorithm, some illustrative example and the comparison with PLM+$\ell_1$, leaving a full analysis of the neural activity for a future publication. We considered the neural activity in time bins of 5, 10 and 20 ms and partition the time series into subsets where the position of the rat is in one of the 400 cells. For each cell, we defined spin variables to attain a value $S_i(t)=-1$ if neuron $i$ is inactive in time bin $t$ and $S_i(t)=+1$ if it is active (here $i=1,\ldots,65$ runs over neurons whereas $t$ runs over all the time bins where the rat is in the  given cell). On each of the sub-samples obtained in this way for the different cells, we run both PLM+$\ell_1$ and the MS method in each of the 400 cells. A further clue that MS detects genuine interactions arises from the expectation that interaction should be more likely between neighbouring cells. A proxy of the spatial location of cells is the tetrode index, i.e. cells recorded by the same tetrode are expected to be closer than those recorded by different tetrodes. At 10ms we found that out of the 9820 interactions detected, 1606 correspond to cells recorded by the same tetrode. Since there are 10 tetrodes, the probability that two cells are recorded by the same tetrode is 10\%. We then tested the null hypothesis that the recovered bonds connect two randomly selected neurons regardless of their location (belonging to the same tetrode). In this case the z-score, which measures the distance between the sample mean and the expected value of the number of intra-tetrode connections in units of standard deviations, is $20,99$ ($\mbox{P-Value} = 2.93\cdot 10^{-82}$) and since the z-score is positive we can conclude that they mostly connect neurons belonging to the same tetrode. We found the same results also for 5ms and 20ms. By contrast, the same analysis for PLM+$\ell_1$ (77883 contacts of which 8204 inter-tetrode) yields a much lower z-score of $4,97$ ($\mbox{P-Value} = 9.18\cdot 10^{-5}$).
    
    One key aspect, is that different positions are visited with different frequency by the rat. Hence different cells correspond to samples of widely varying size and about 75\% of them have length less than 300 data-points. This is documented in Figure \ref{fig:gridcell1} (see inset). The left panel of the figure shows that PLM+$\ell_1$ estimates way more links than MS. The number of interactions detected with MS increases with the size of the sample, which means that the inferred network is denser in cells that are visited more often. This is expected in the MS approach. The results of MS are highly consistent with those of PLM+$\ell_1$ in that 99\% of the interactions detected by MS are also detected by PLM+$\ell_1$. Conversely, PLM+$\ell_1$ detects many more interactions, many of which are attached a very low confidence by MS. Indeed, we found PLM+$\ell_1$ to suffer from instability problems in cases of very small samples. The top interactions inferred by MS are found to be very consistent across the grid in the sense that their confidence is positive in several different cells. Interestingly, the network of the most frequent interactions connect the five interneurons. Moreover the number of connections found for grid cells and interneurons do not correlate with their activity making the detected dependency stable against high fluctuations in the neural activity (right panel). 
    
\begin{figure}[t]
\centering
\includegraphics[width = \columnwidth]{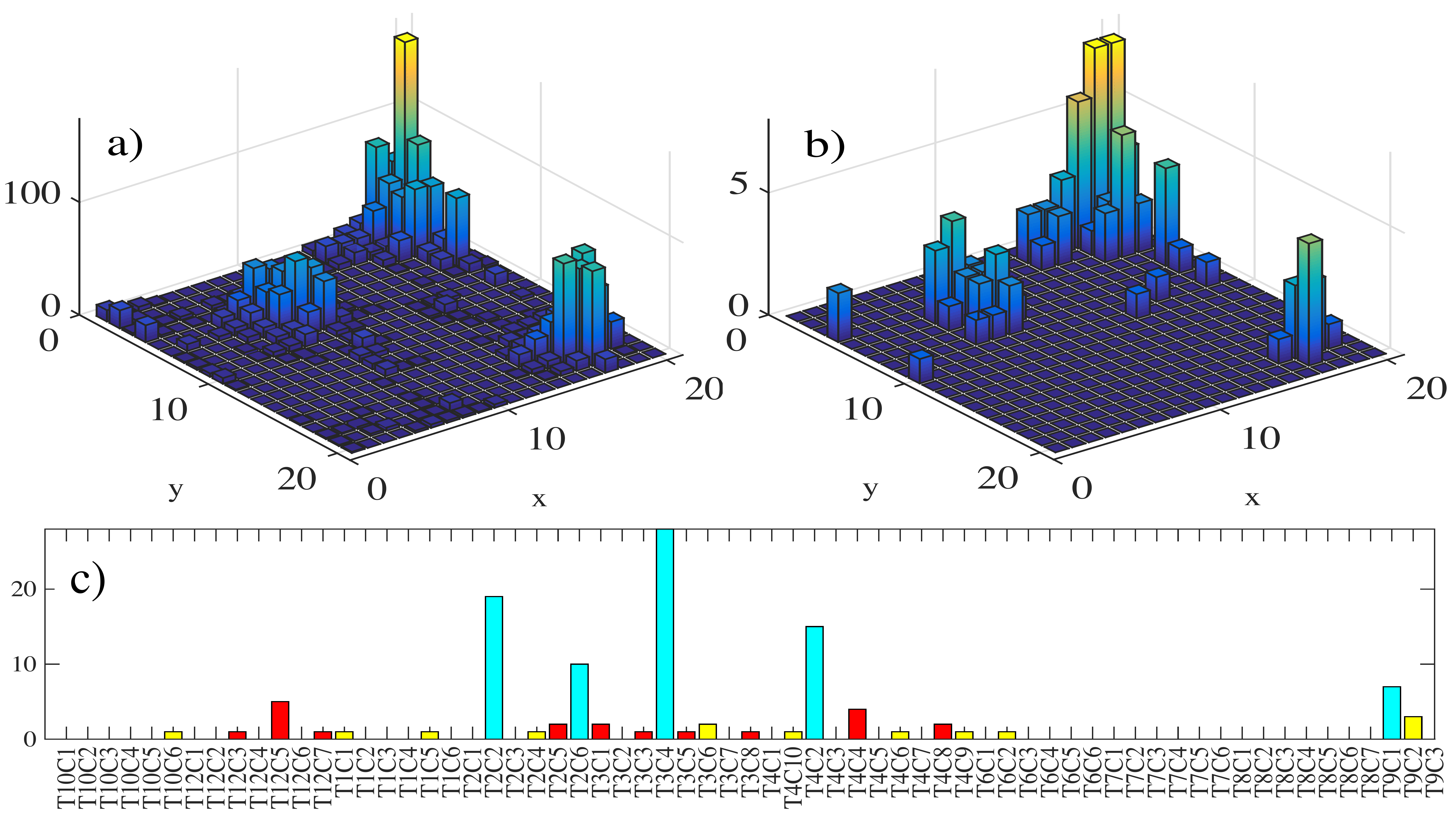}
\caption{The figure shows the spatial spiking pattern (a) and the spatial connectivity pattern (b) of the grid cell T7C1, namely the number of spikes and the number of connections inferred by MS with a time bin of 10 ms for the above mentioned grid cell versus the spatial position of the rat in the box. Interestingly for this particular grid cell, the connectivity pattern follows the spiking one, a feature not common to all the grid cell in the considered ensamble. This neuron makes connection mostly with the five interneurons (70\%) but also with grid cells (20\%) and other neurons (10\%). This is illustrated in detail in panel (c) with the following color code: cyan for interneurons, red for grid cells and yellow for all other neurons. In particular, panel (c) reports the number of connections inferred between the grid cell T7C1 and all the other neurons which are listed on the x axis.}
\label{fig:gridcell2}
\end{figure}     
    
    The firing rates of grid cells exhibits a remarkable variation across space that reveals the characteristic hexagonal patterns \cite{Stensola}, whereas interneurons have a more uniform firing activity in space. Figure \ref{fig:gridcell2} shows a representative example for a particular grid cell. The pattern of interactions between grid cells and between grid cells and interneurons follows the spatial dependence of their activity, but it exhibits a more heterogeneous behaviour. This is expected, because in regions where the firing activity of the grid cell is very low, simpler models with fewer parameters (i.e. without a bond) are preferred because the corresponding spin is almost frozen. Yet we found no statistically significant dependence on the spatial position of the interactions that were detected, i.e. statistically the interactions found for a given neuron in one cell do not depend on the position of the cell in the space, suggesting that these correspond to genuine interactions. 
The grid cell shown in Figure \ref{fig:gridcell2} has significant interaction with the interneurons that is statistically independent from the spatial position. This was found to be true of several other grid cells, but we also found examples of grid cells preferentially interacting with grid cells. This shows that neighborhoods of interaction networks are very heterogeneous, which is to be expected if one thinks of this network as being a very sparse sub-sampling of a much larger network. 

        In summary, the application of the MS approach to neural data shows its potential for statistical dependencies between the neural activity of different cells that is not affected by the confounding effect of the spatial position variable. Furthermore, at odds with other methods, our approach naturally takes into account the fact that only a tiny region of a larger network of interactions is observed and recognizes statistical dependencies on the basis of the quality of the observations, being more selective when only few observations are available.

\section{\label{sec: conclusions}Conclusions}
In this paper we presented a model selection approach for topology reconstruction of a network of interacting binary nodes. Our approach explores the idea of calculating, in a fully Bayesian fashion, the posterior probabilities of all possible graphical models and it relies on the observation that in the small sample limit, complex models are heavily penalised with respect to simpler one. It makes sense, in this limit, to consider only minimal clusters composed of just two spins. Therefore, we derive a full Bayesian approach to model selection for a system of two spins, potentially interacting with each other and with the rest of the network through effective fields. The use of such minimal clusters reduces the problem to a simple two body system with a relative small number of possible graphs allowing a fully analytic treatment and a direct geometric visualization. This choice is justified in the noisy data regime when usually simple schemes may capture better the underlying structure than more complex techniques, better suited for high quality data. In fact, we have proven on very noisy synthetic data sets that such a simple method is at least equivalent, if not superior, to the well known $\ell_1$ regularized pseudo-likelihood method for sparse networks and weakly coupled spins. Moreover a list of advantages accompanies the use of our method: it is a fully automatic procedure and does not need selecting additional {\itshape ad hoc} parameters as the choice of the regularizer and the thresholding value in PLM+$\ell_1$; it is computationally convenient since it can be implemented as a direct online tool: the boundaries of the partitions in the parameters space are computed once for a given value of $N$, and they can then be used for all pairs. 
This sounds particularly appealing for large networks applications where pseudo-likelihood methods typically becomes slow. 
Furthermore in Sec.\ \ref{sec: real_data} we have shown that MS is a method that is particularly suited to study non-stationary data, as shown by the analysis of the US stock market data,  or to remove confounding effect of external variables, as in the case of neural data, because it is ideally suited to work on small samples.
Finally, when the network of interactions is only partially observed, it represents a valuable alternative to existing algorithm for dealing with hidden variables. In fact, our method is able to detect significative statistical dependences among the observed spins given a limited amount of data and return stable predictions, since they are not influenced by how much of the network we are able to probe. This is very interesting for biological applications, as we discussed in in Sec.\ \ref{sec: real_data}, where existing experimental techniques might not be able to record from the whole network.
Our methods looses accuracy when less sparse networks are taken into account and when more loops arise from the connectivity pattern of spins. In these cases it would be straightforward and very interesting to extend our approach to 3-body or 4-body clusters of spins in order to count for small loops in the network. There are several other directions where the MS approach could be generalised in a straightforward manner as, e.g. to non-equilibrium spin models \cite{Yasser11}, to Potts spin models such as those used in contact prediction for proteins \cite{Aurell12} etc.

Even for high quality data, for sparse topologies and weak couplings, the above presented Model Selection approach exhibits the interesting property of recognizing all direct connections in the network at the cost of a reasonable small number of false positive. Therefore our method candidates as an efficient pruning algorithm for sparse networks of any size and as a pre-treatment tool for reducing the number of features and saving computational time in subsequent applications of more involved methods, as the pseudo-likelihood ones. Furthermore the excess of false positive can be cured at least partially in a very simple way improving significantly the accuracy of our method especially in the large $N$ regime.

\begin{acknowledgments}
The authors would like to thank Aurélien Decelle for interesting discussions and sharing his code and Ole Winther for sharing the code used for the expectation consistent algorithm. This work was supported by the Centre of Excellence scheme of the Research Council of Norway (Centre for Neural Computation, grant number 223262) and the Marie Curie Training Network NETADIS (FP7, grant 290038) as well as computational resources provided by NOTUR.
\end{acknowledgments}

\medskip
 
\bibliographystyle{unsrt}
\bibliography{mybib}

\newpage


\appendix

\section{\label{sec: SM1}Synthetic data: methods and supplementary analysis}
In the appendices we present two sections with details about the main results of the paper. In the first section we report additional simulations for the different topologies discussed in the paper and give further insights on the methods; in the second section we will focus on US stock market interactions showing that, under our assumptions, configurations of stocks at time $t$ are uncorrelated with those at time $t + \tau$ for $\tau \neq 0 $ and for different values of $N$; moreover we also further contrast MS and PLM+$\ell_1$ and compare their outcomes with respect to the absolute values of connected correlations. Finally we provide a Matlab implementation of our method.

\begin{figure}[t!]
\centering
\subfigure{\includegraphics[width = 0.45\columnwidth]{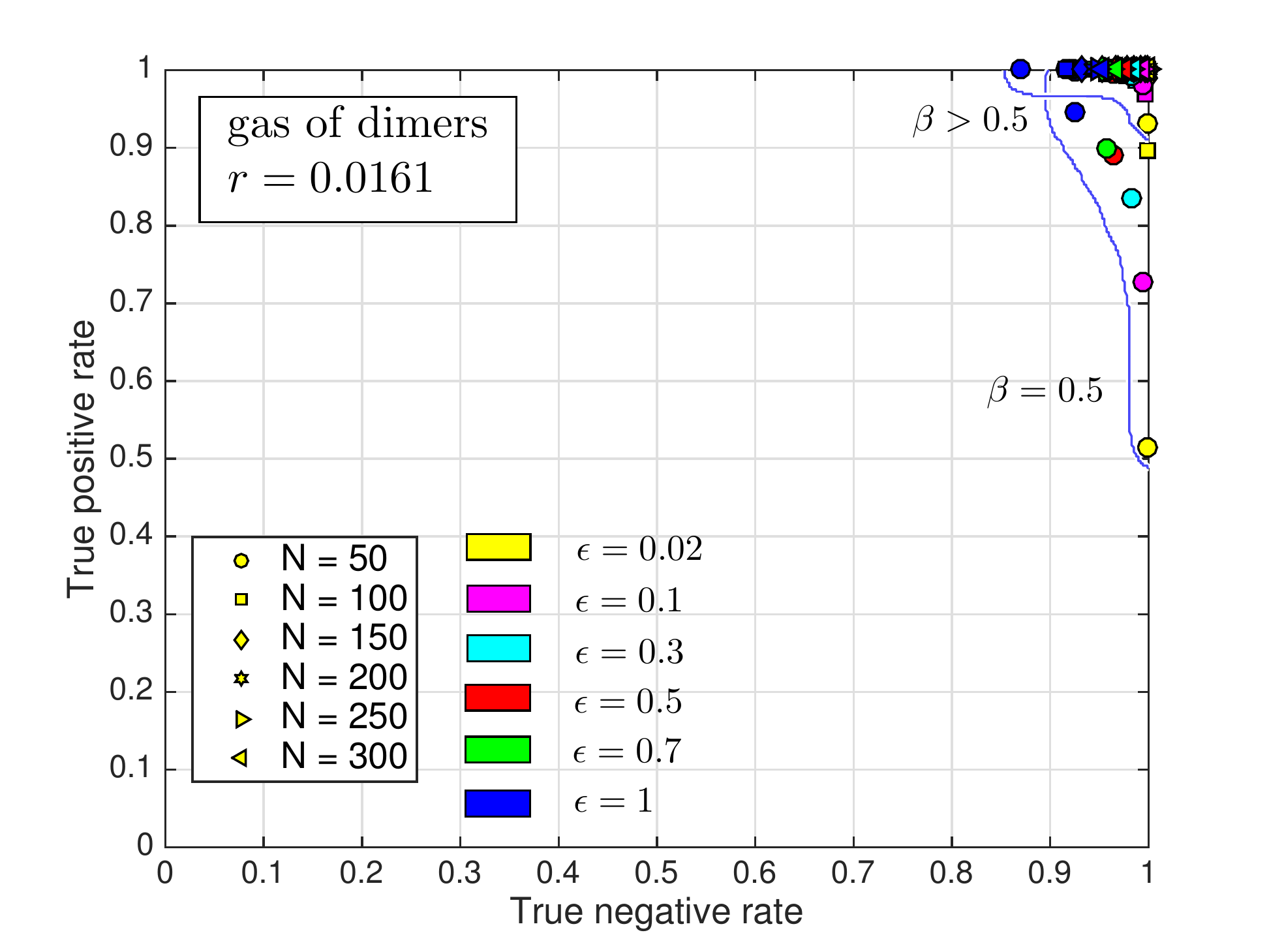}}
\subfigure{\includegraphics[width = 0.45\columnwidth]{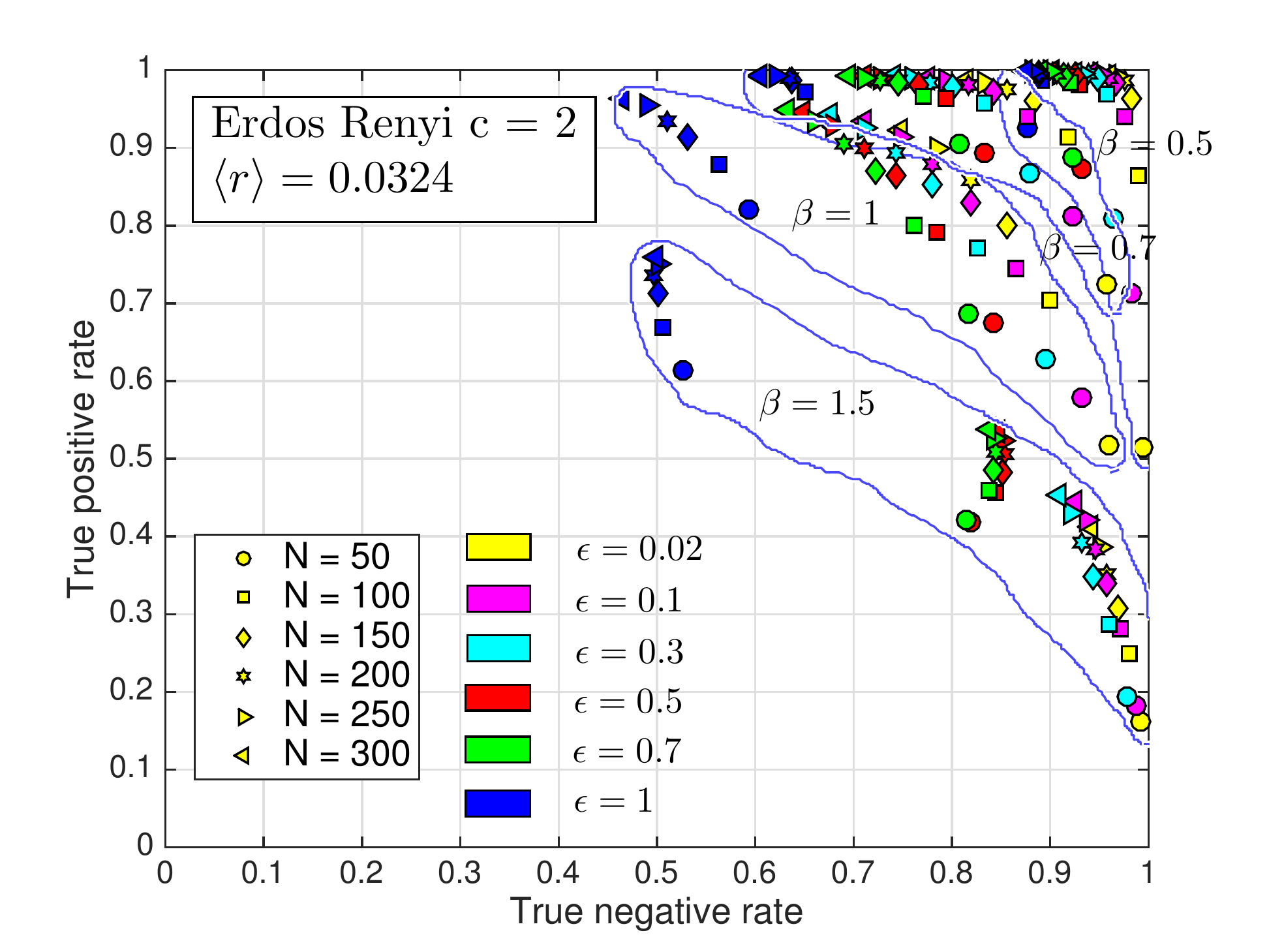}}
\subfigure{\includegraphics[width = 0.45\columnwidth]{Erdos3ALL_B.pdf}}
\subfigure{\includegraphics[width = 0.45\columnwidth]{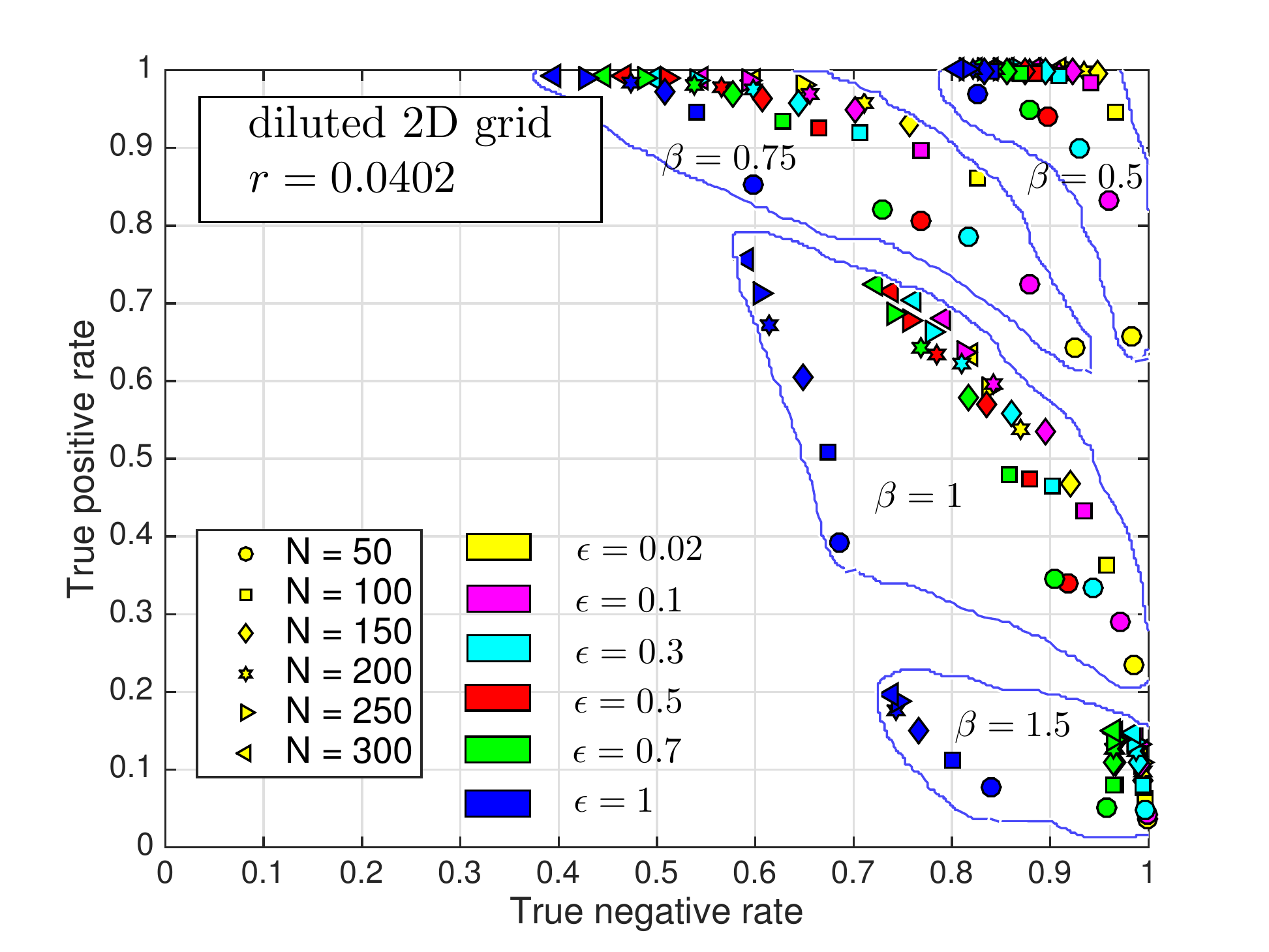}}
\subfigure{\includegraphics[width = 0.45\columnwidth]{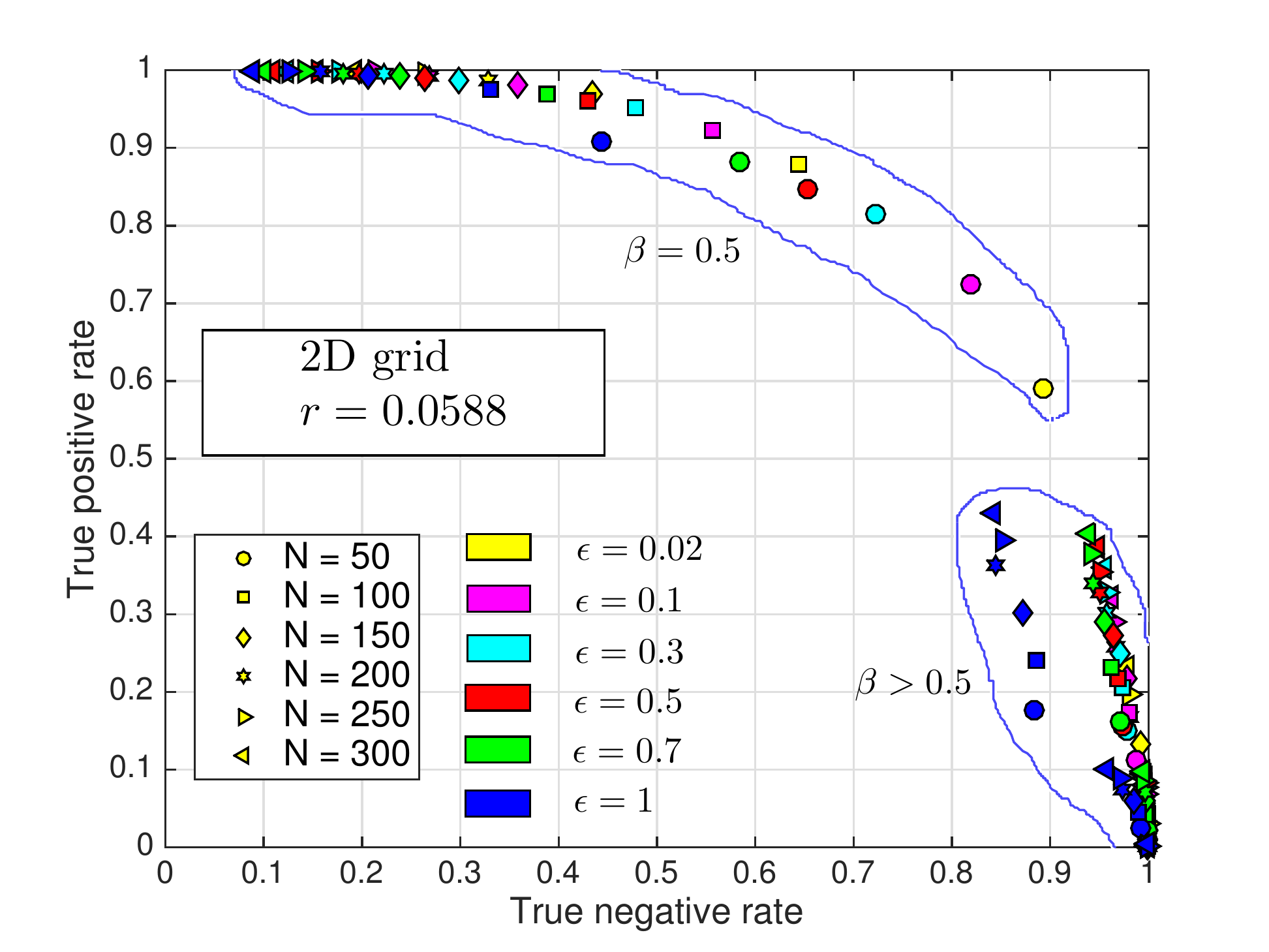}}
\caption{Testing MS method on synthetic data generated assuming five sparse topologies: gas of dimers, Erdos Renyi graphs with $c = 2$ and $c = 3$, a full and diluted bidimensional grid.}
\label{fig:scatterMsAppendix}
\end{figure}

In Fig.\ \ref{fig:scatterMsAppendix} we summarize the performances of our method at varying the length of the sample $N$, the strength of the couplings $\beta$ and the value of $\epsilon$ for some of the sparse topologies investigated: gas of dimers, Erdos Renyi graphs with $c = 2$ and $c = 3$, a full and diluted (30\% of dilution) bidimensional grid. In particular for each topology and each choice of $\beta$, we generated one hundred different instances of the same topology and for each of them we generated samples with different N through Monte Carlo sampling. For all cases of Fig.\ \ref{fig:scatterMsAppendix}, the size of the network has been fixed to $n = 64$ and the value of the couplings $\beta$ have been drawn from a unimodal distribution for the bidimensional grids and from a bimodal distribution in the other cases. After having evaluated the matrix of the confidence for each single realisation and each N, we calculated the TPR and TNR for different choices of $\epsilon$ and averaged their values over all realisations with the same topology, $\beta$, N and $\epsilon$. The ideal case is the one of the gas of dimers for which the method provides an almost perfect recovery, especially for high values of $\beta$. In all other cases the best performances are achieved in the weak coupling regime ($\beta = 0.5$) and for topologies with few bonds and loops. In this respect, the case of the full bidimensional grid is representative since it shows the failure of our method in recovering a graph with many loops even in the weakly interacting regime.

\begin{figure}[t!]
\centering
\subfigure{\includegraphics[width = 0.4\columnwidth]{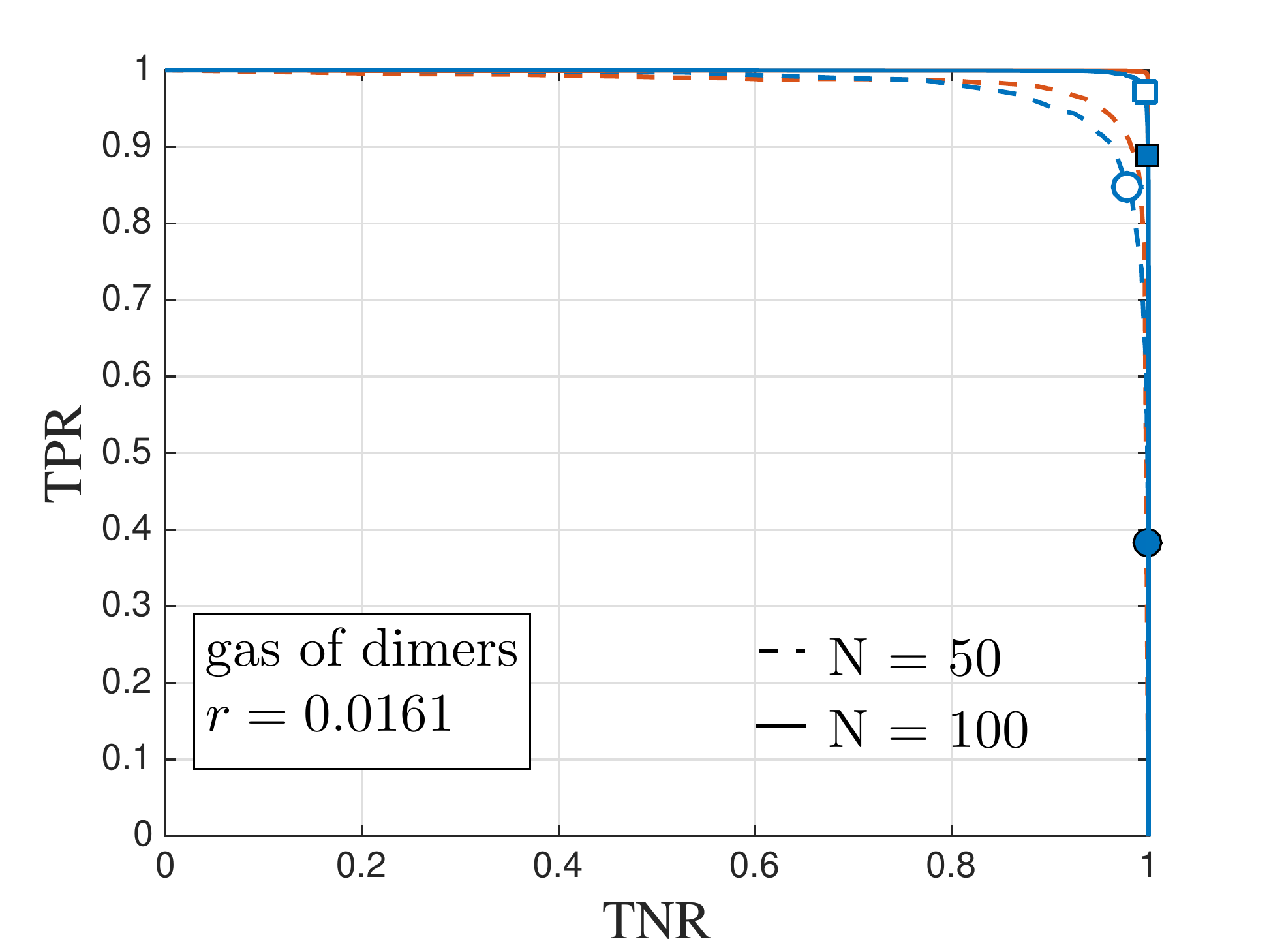}}
\subfigure{\includegraphics[width = 0.4\columnwidth]{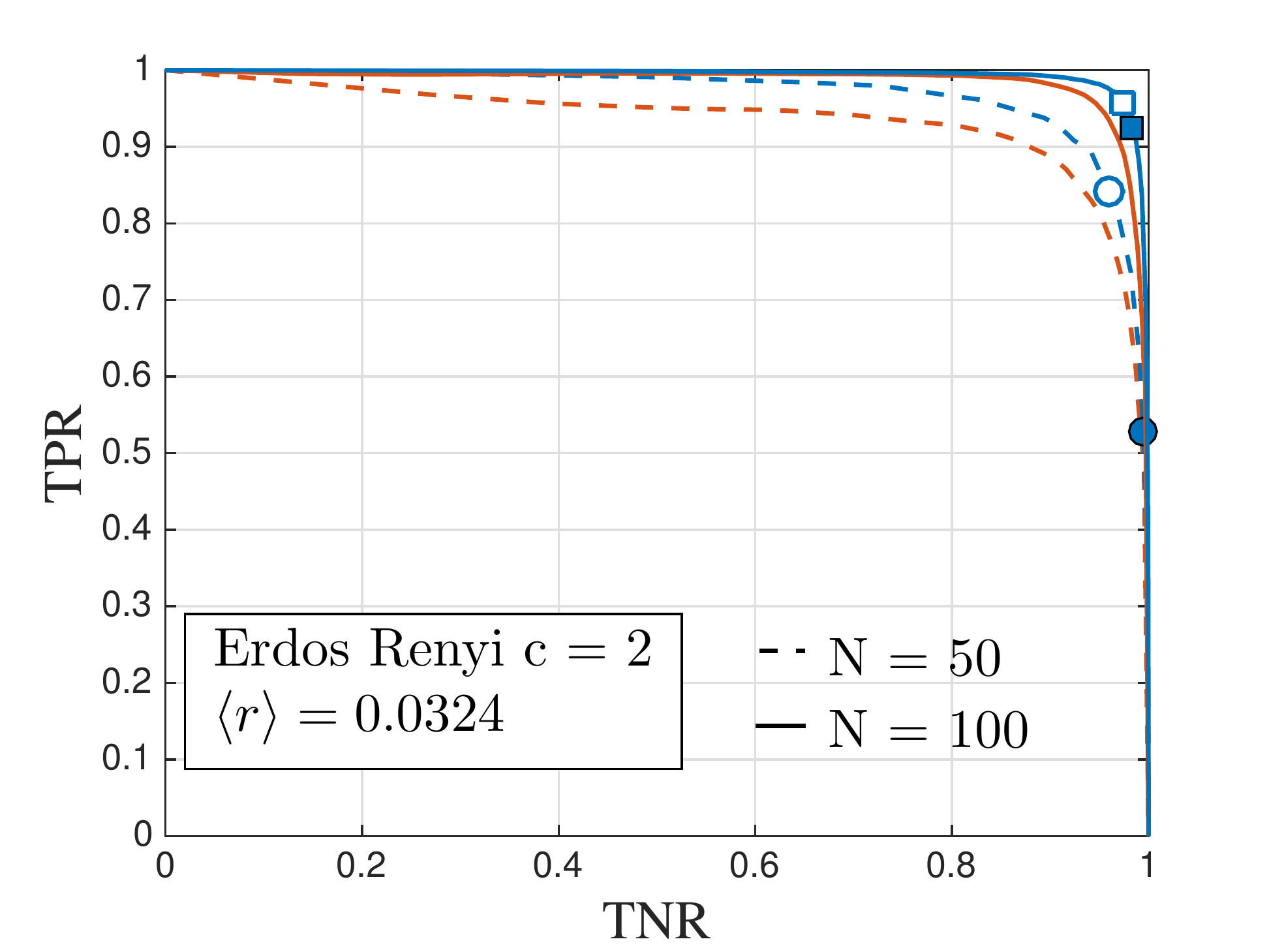}} \\
\subfigure{\includegraphics[width = 0.4\columnwidth]{scatterER3.pdf}}
\subfigure{\includegraphics[width = 0.4\columnwidth]{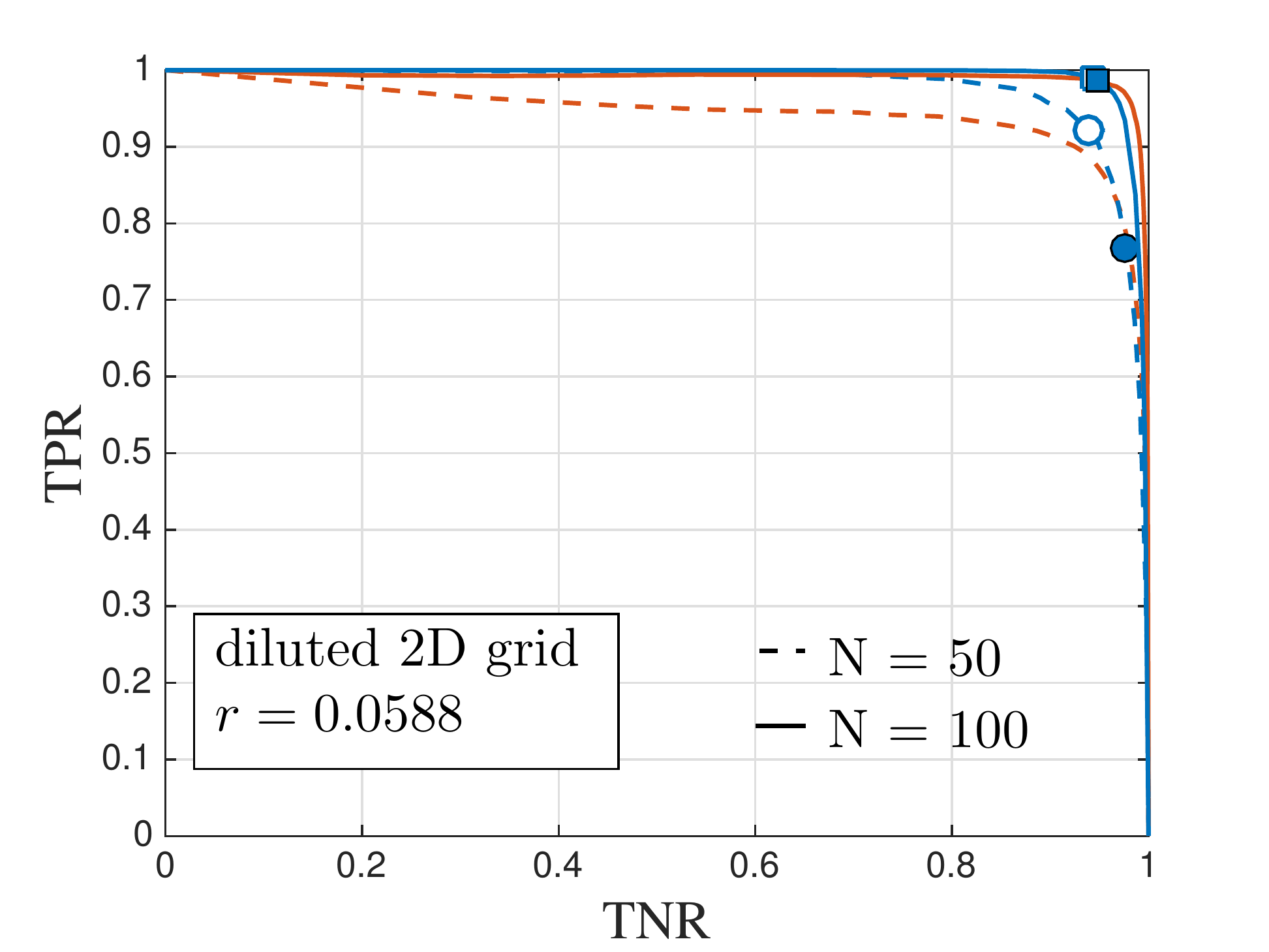}} \\
\caption{Comparison between MS (blue lines) and PLM+$\ell_1$ (red lines) for network of $n =64$ nodes arranged in four different topologies. Each point of the curves represents the average of TNR and TPR over one hundred different realisations of the same topologies with a mean `bond'-`no-bond' ratio $\langle r \rangle$ ($r$ when its value is constant across realisations). The comparison is drawn for two different values of $N$ and in the weak couplings regime ($\beta = 0.5$). Two different points are highlighted for each curve showing the effects of employing a self consistent (coloured markers) and a N-varying procedure (white markers) for selecting the models' prior coefficient.}
\label{fig:scatterMsVsL1Appendix}
\end{figure}

\begin{figure}[t!]
\centering
\subfigure{\includegraphics[width = 0.3\columnwidth]{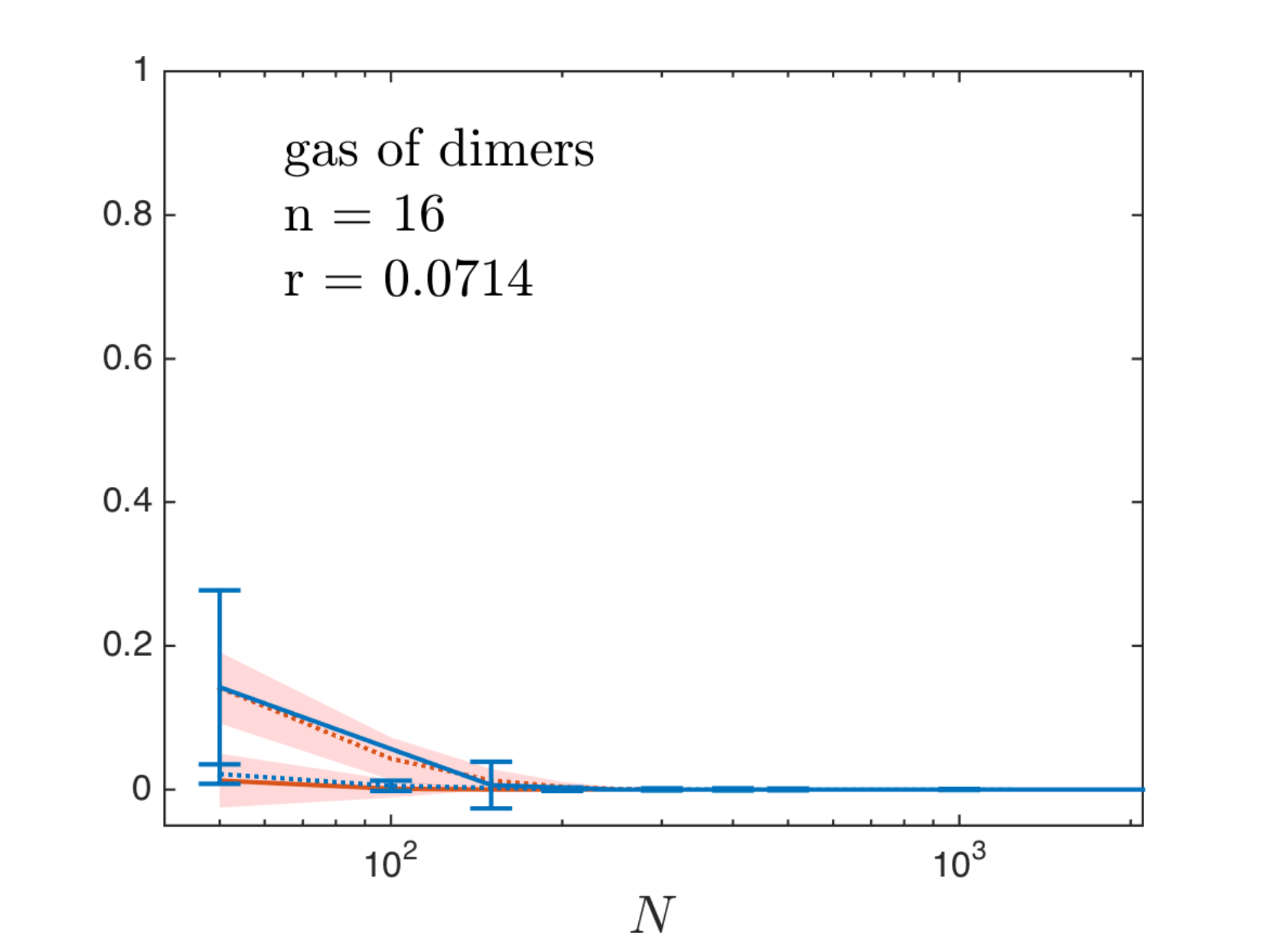}}
\subfigure{\includegraphics[width = 0.3\columnwidth]{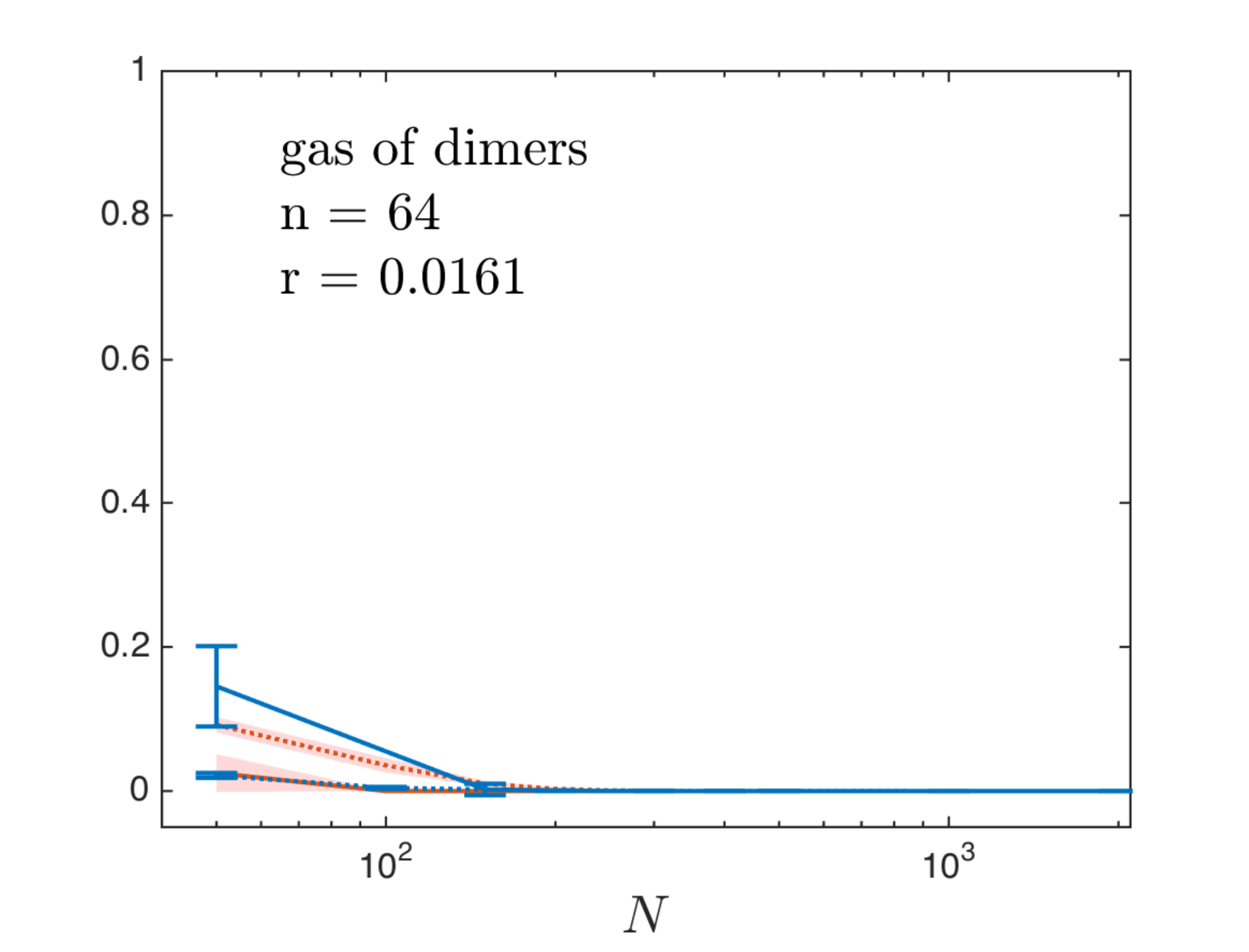}}
\subfigure{\includegraphics[width = 0.3\columnwidth]{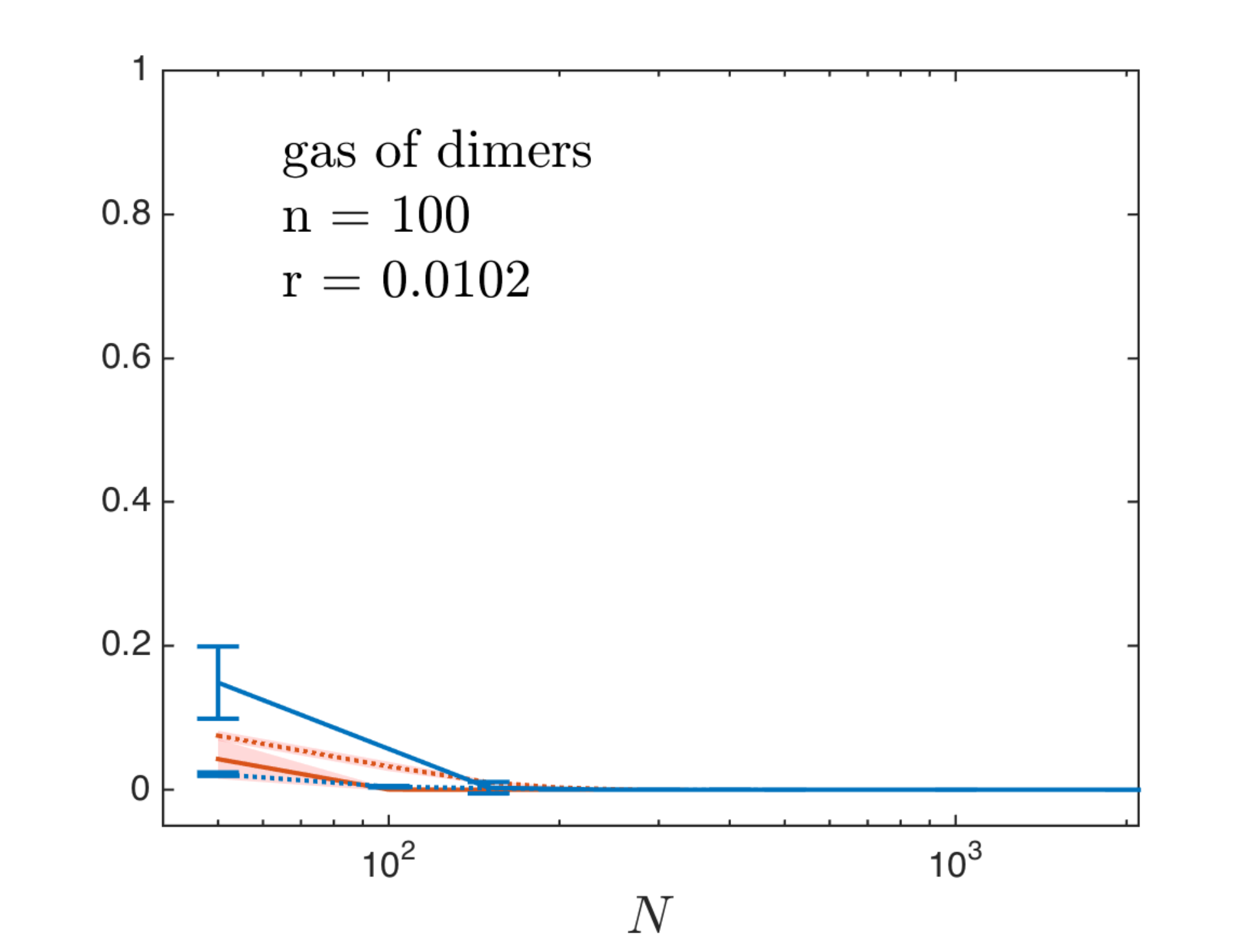}} \\
\subfigure{\includegraphics[width = 0.3\columnwidth]{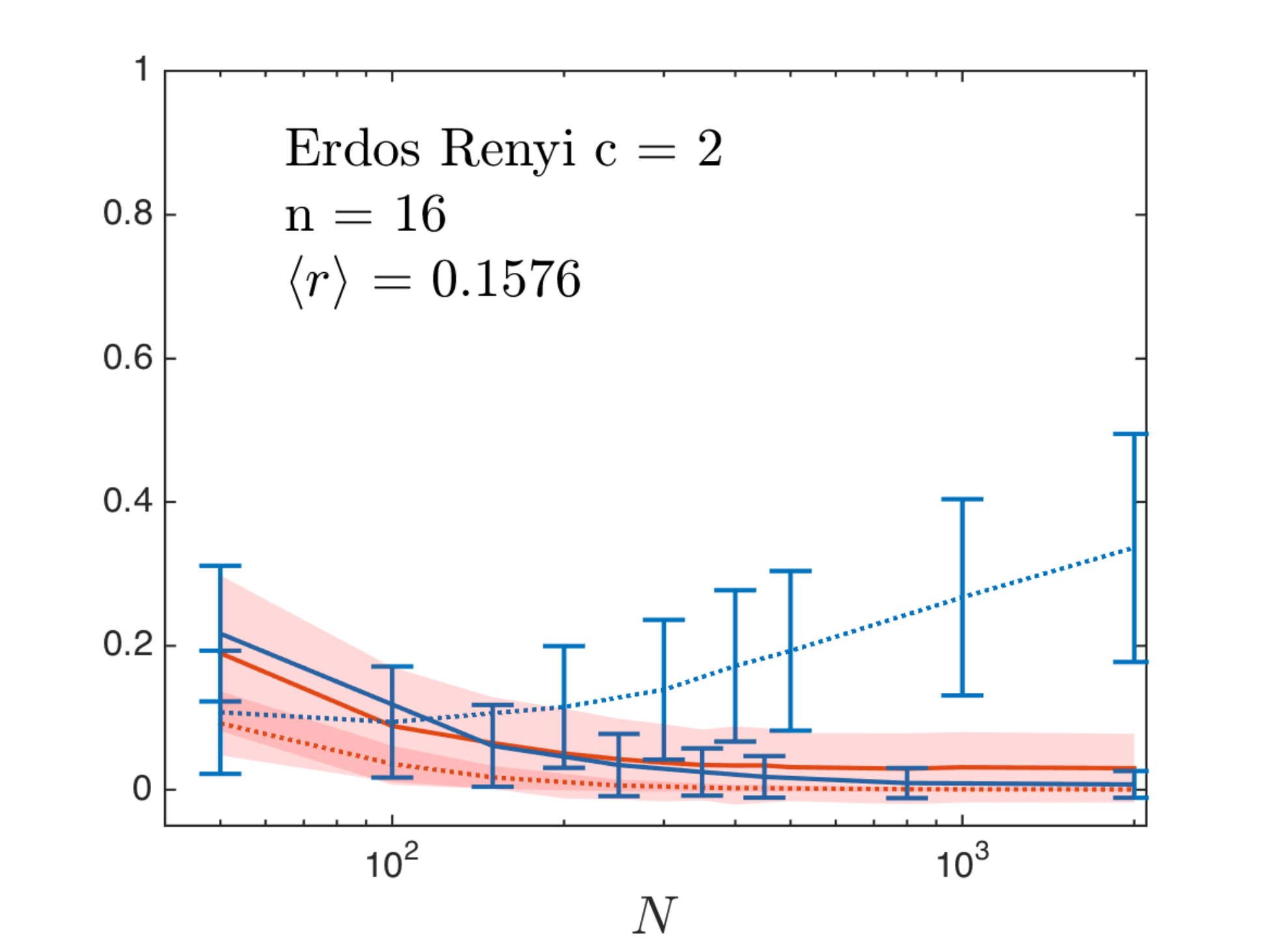}}
\subfigure{\includegraphics[width = 0.3\columnwidth]{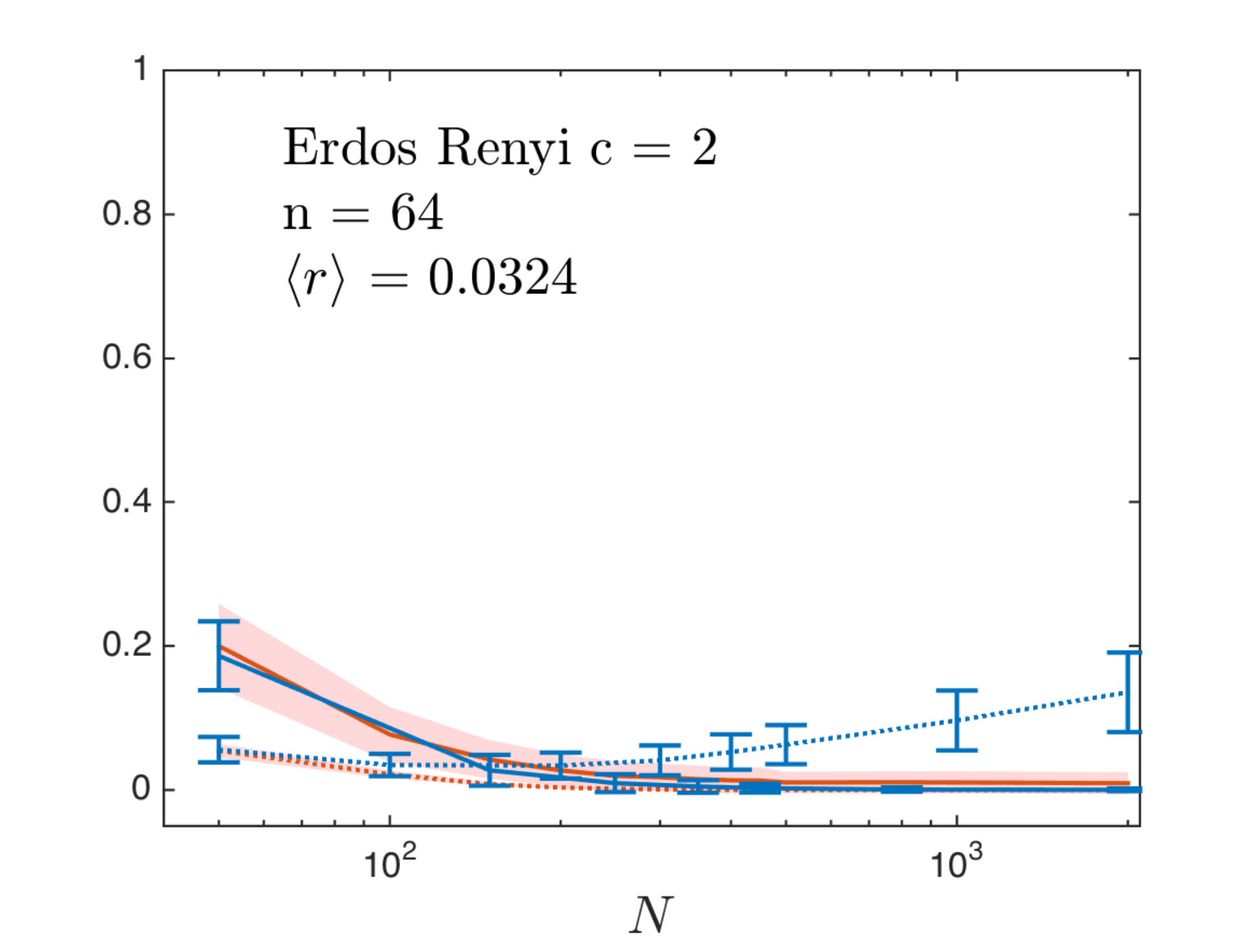}}
\subfigure{\includegraphics[width = 0.3\columnwidth]{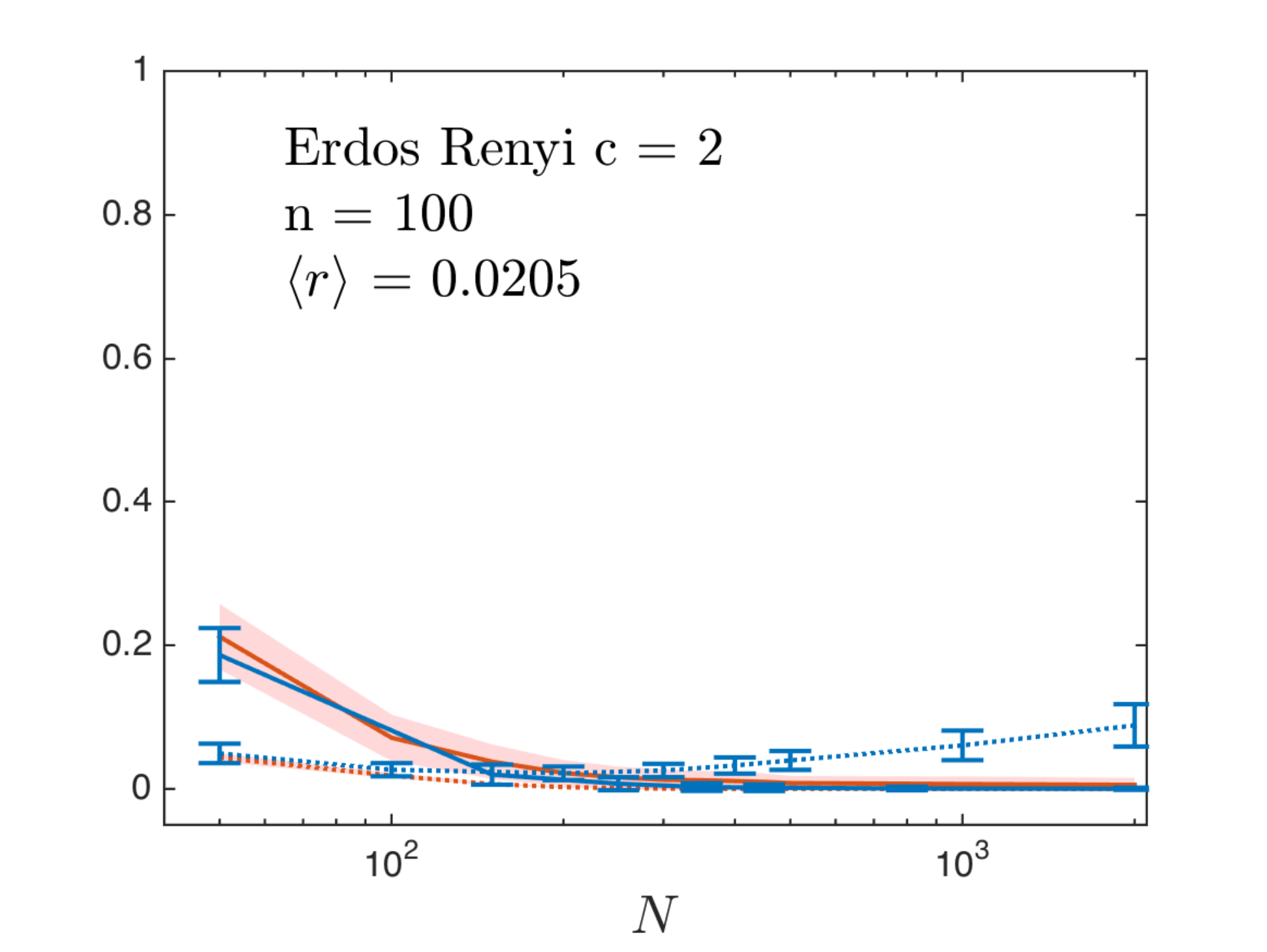}} \\
\subfigure{\includegraphics[width = 0.3\columnwidth]{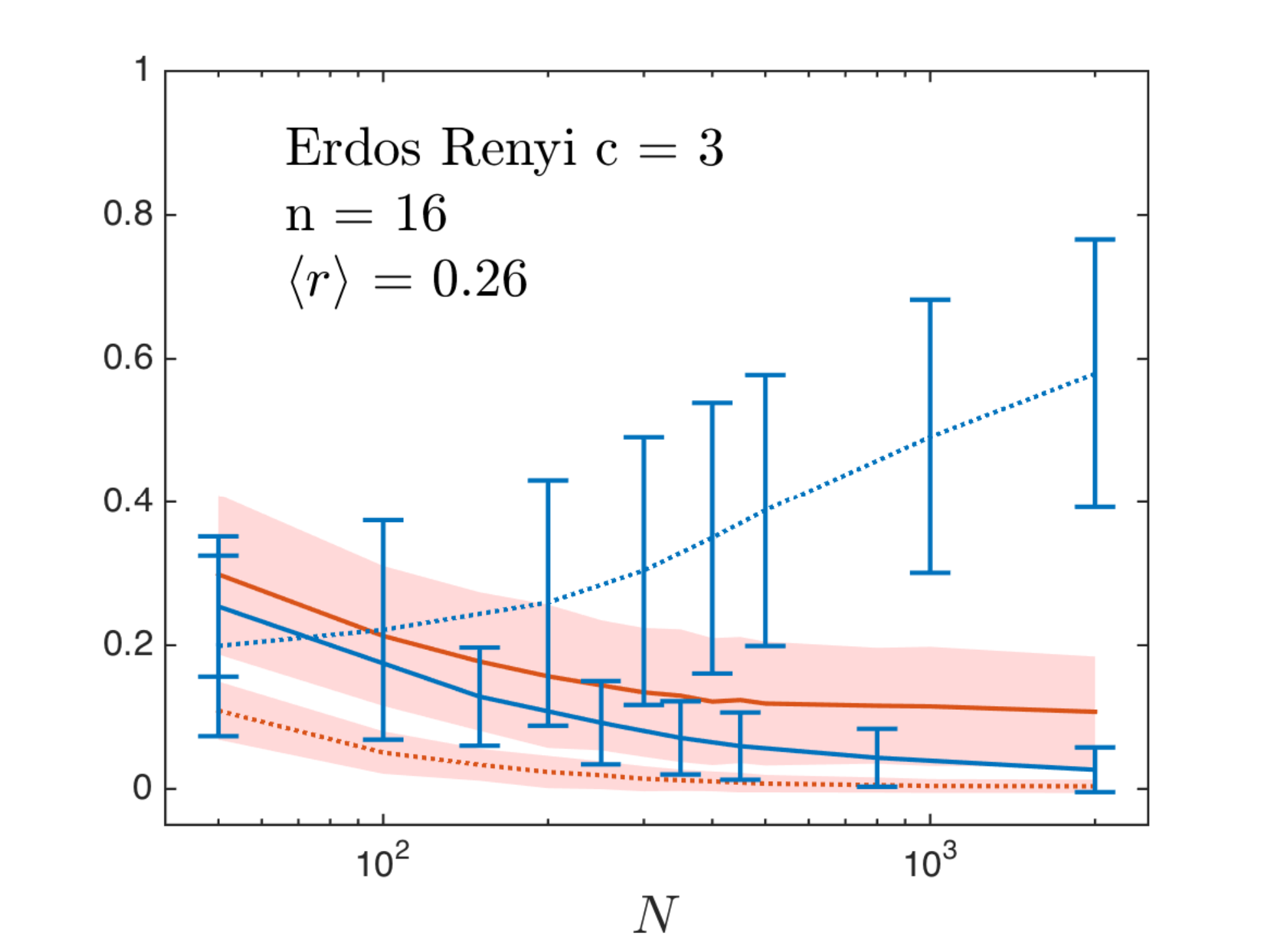}}
\subfigure{\includegraphics[width = 0.3\columnwidth]{Erdos3Long64.pdf}}
\subfigure{\includegraphics[width = 0.3\columnwidth]{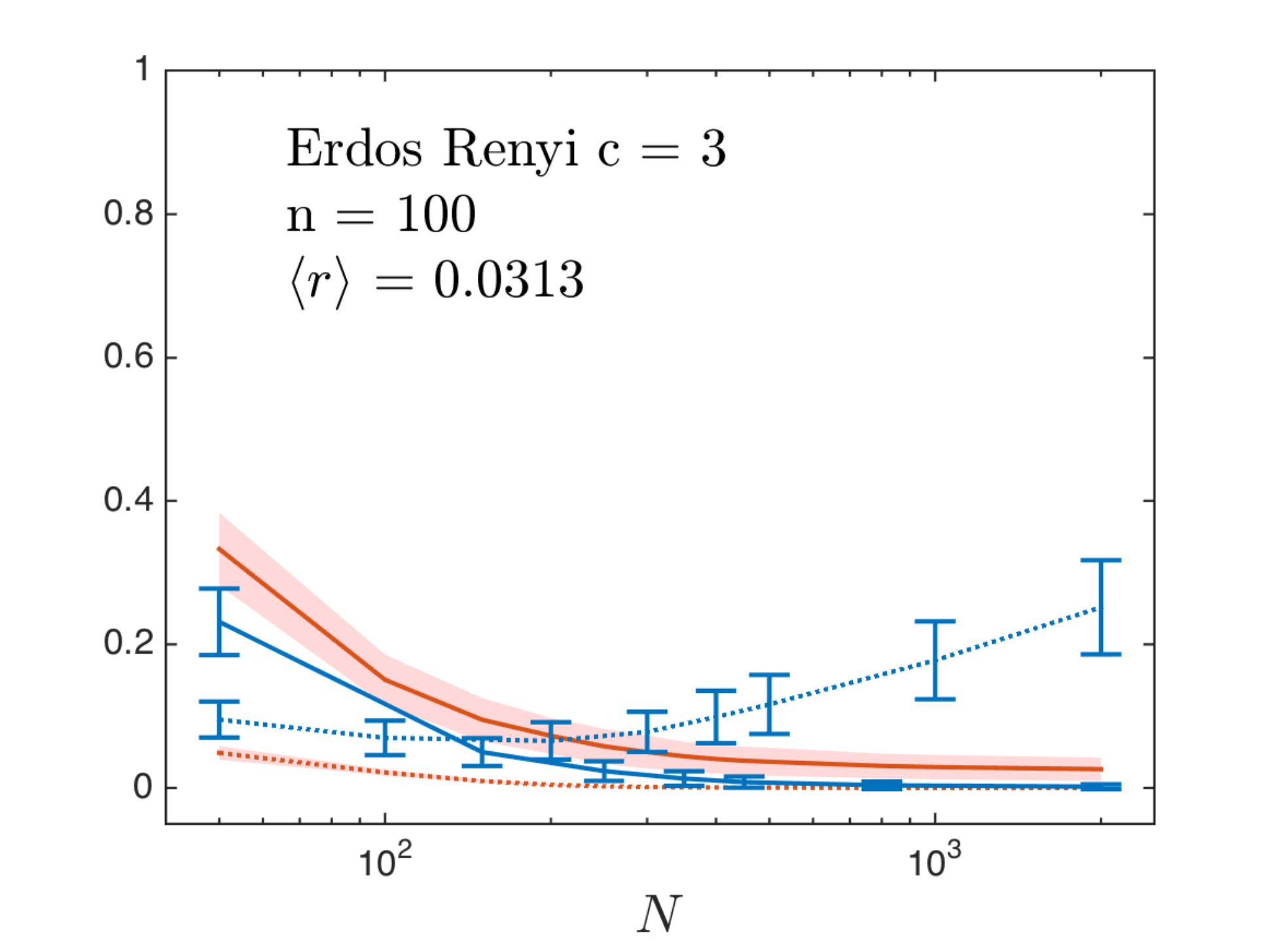}} \\
\subfigure{\includegraphics[width = 0.3\columnwidth]{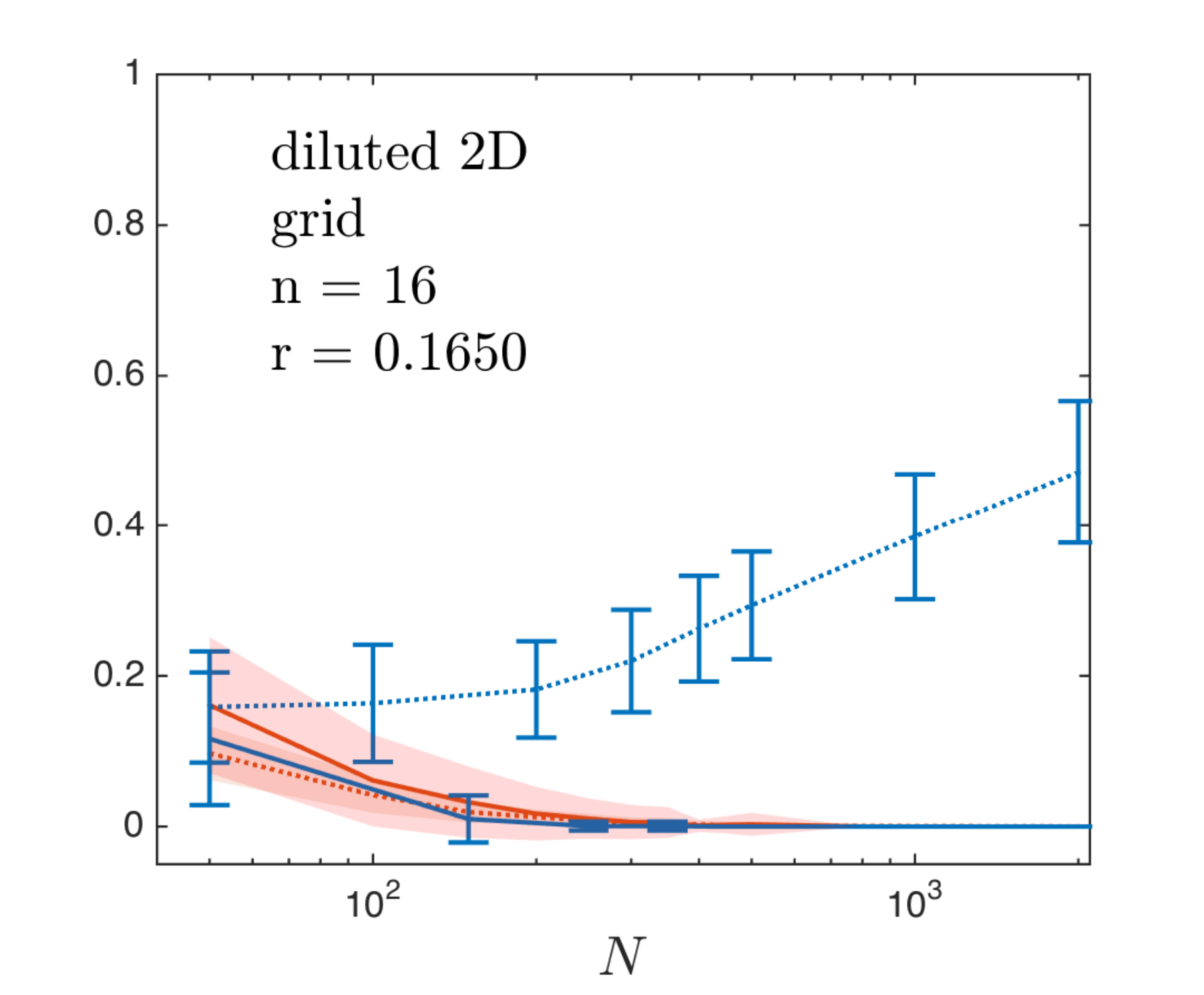}}
\subfigure{\includegraphics[width = 0.3\columnwidth]{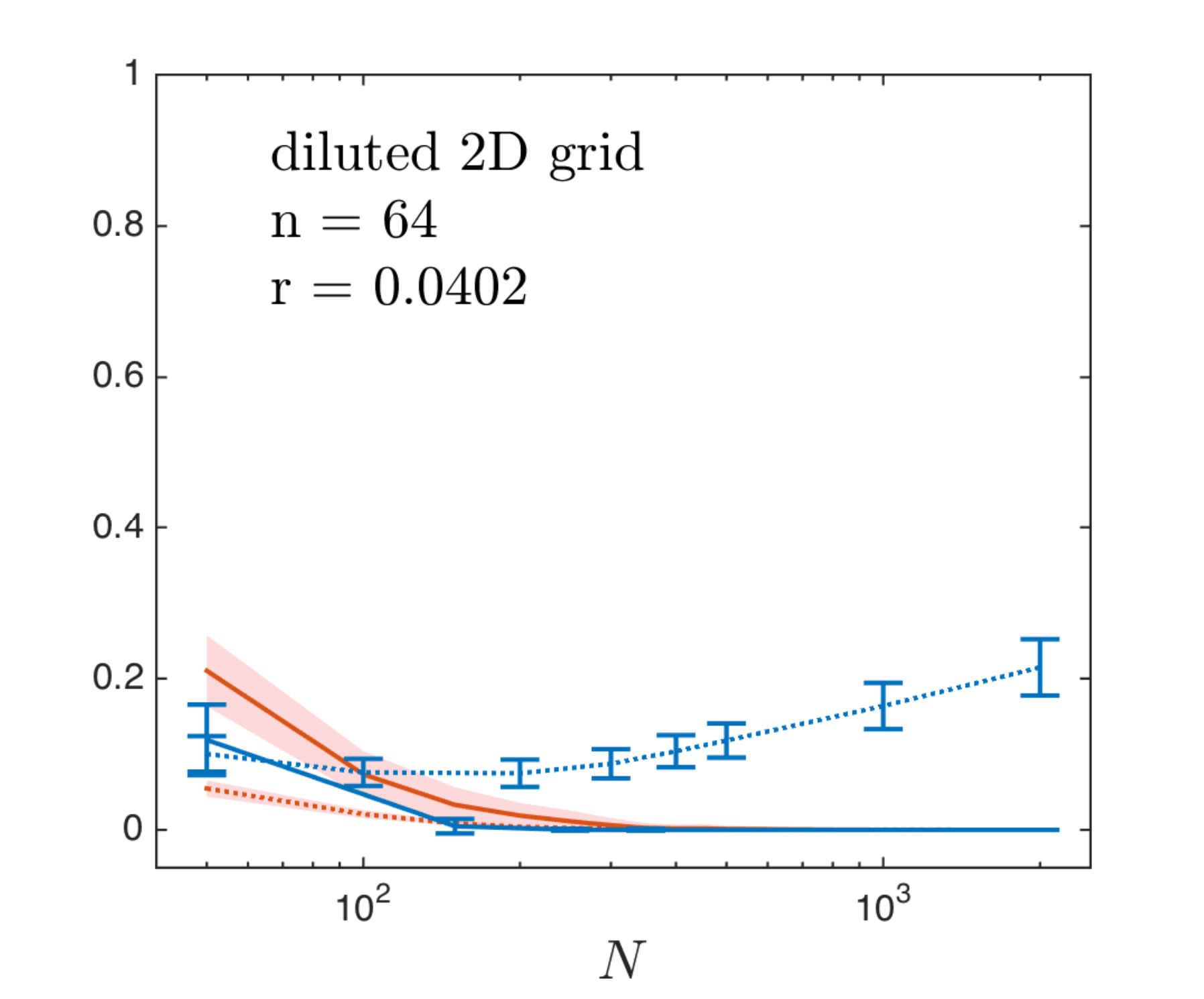}}
\subfigure{\includegraphics[width = 0.3\columnwidth]{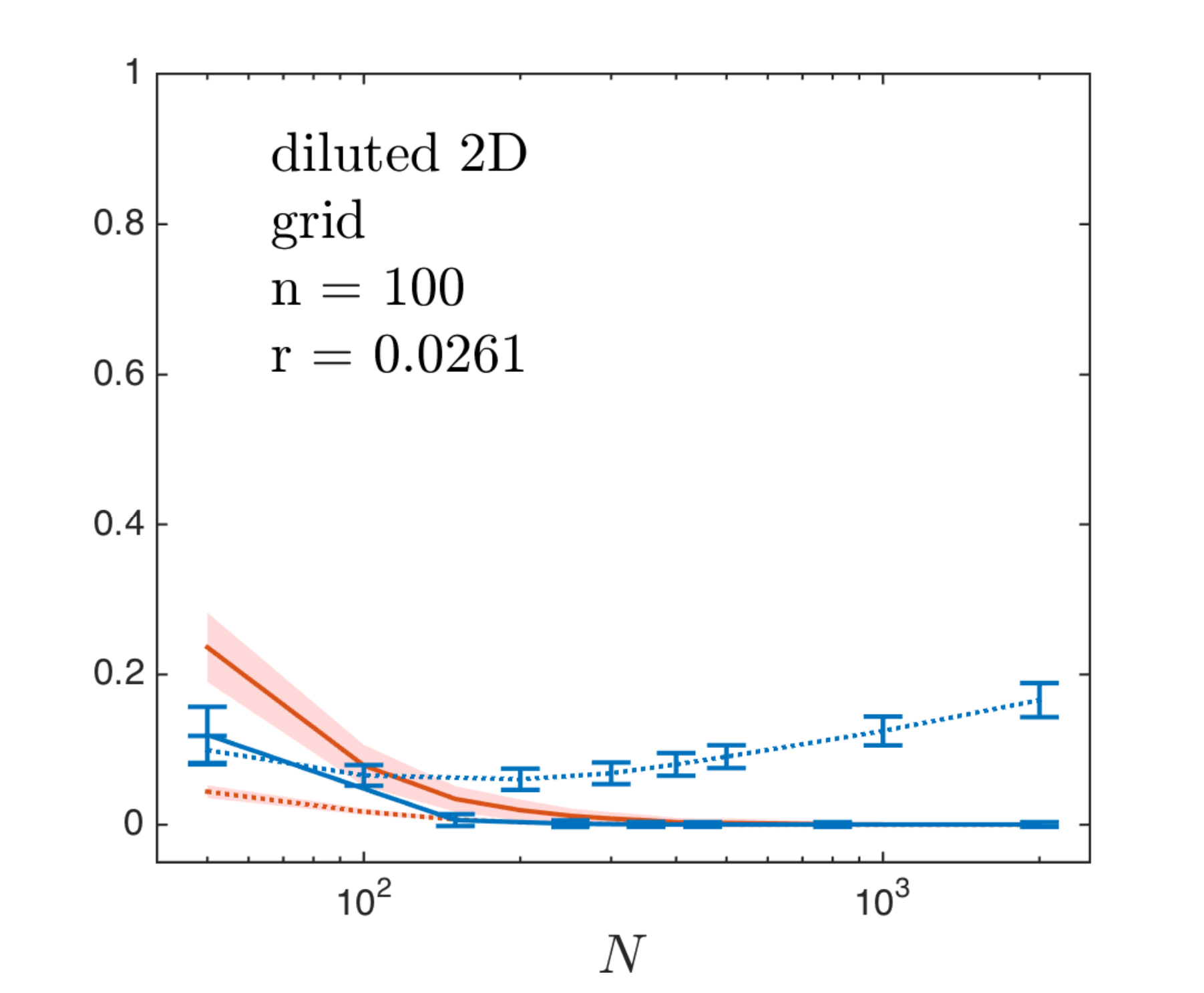}} \\
\caption{Comparison between MS (blue lines plus error bars) and PLM+$\ell_1$ (red lines plus shaded error bars) for network of different sizes $n =16,64 ,100$ with nodes arranged in four different topologies. FPR (dotted lines) and FNR (solid lines) are shown versus the size of the sample $N$.}
\label{fig:scatterMsVsL1longAppendix}
\end{figure}

The comparison with PLM+$\ell_1$ in the weak coupling regime ($\beta = 0.5$) is in Fig.\ \ref{fig:scatterMsVsL1Appendix} for all the previous topologies except for the grid.
The curves are drawn at varying $\epsilon$ for MS and the $\ell _1$ regularizer $\lambda$ for PLM+$\ell_1$ ($\lambda$ ranging from 0 to $1.5 \cdot\lambda _{max}$ where $\lambda _{max}$ represents the threshold above which the inferred graph is completely disconnected). The comparison between MS and PLM+$\ell_1$ for these topologies exhibits the same features discussed for the representative case of an Erdos Renyi graph $c = 3$. It is interesting to note how this inference tecnique behaves almost perfectly (the curve pass through the right upper corner), even with only $N = 100$ data points, in the ideal case of a gas of dimers. Important points outlined in the paper regarding the performance of MS, expressed now in terms of FPR and FNR, at varying $N$ and $n$: the algorithm improves at enlarging the size of the network $n$ and the FNRs resulting from the application of MS go to zero faster and with smaller error bars than those of PLM+$\ell_1$. This is shown more in detail in Fig.\ \ref{fig:scatterMsVsL1longAppendix} for all the topologies investigated along with the comparison with PLM+$\ell_1$ outcomes for a fixed value of the regularizer $\lambda = 0.5\lambda_{max}$ which optimizes PLM+$\ell_1$ results.

\begin{figure}[t!]
\centering
\includegraphics[width = 0.45\textwidth]{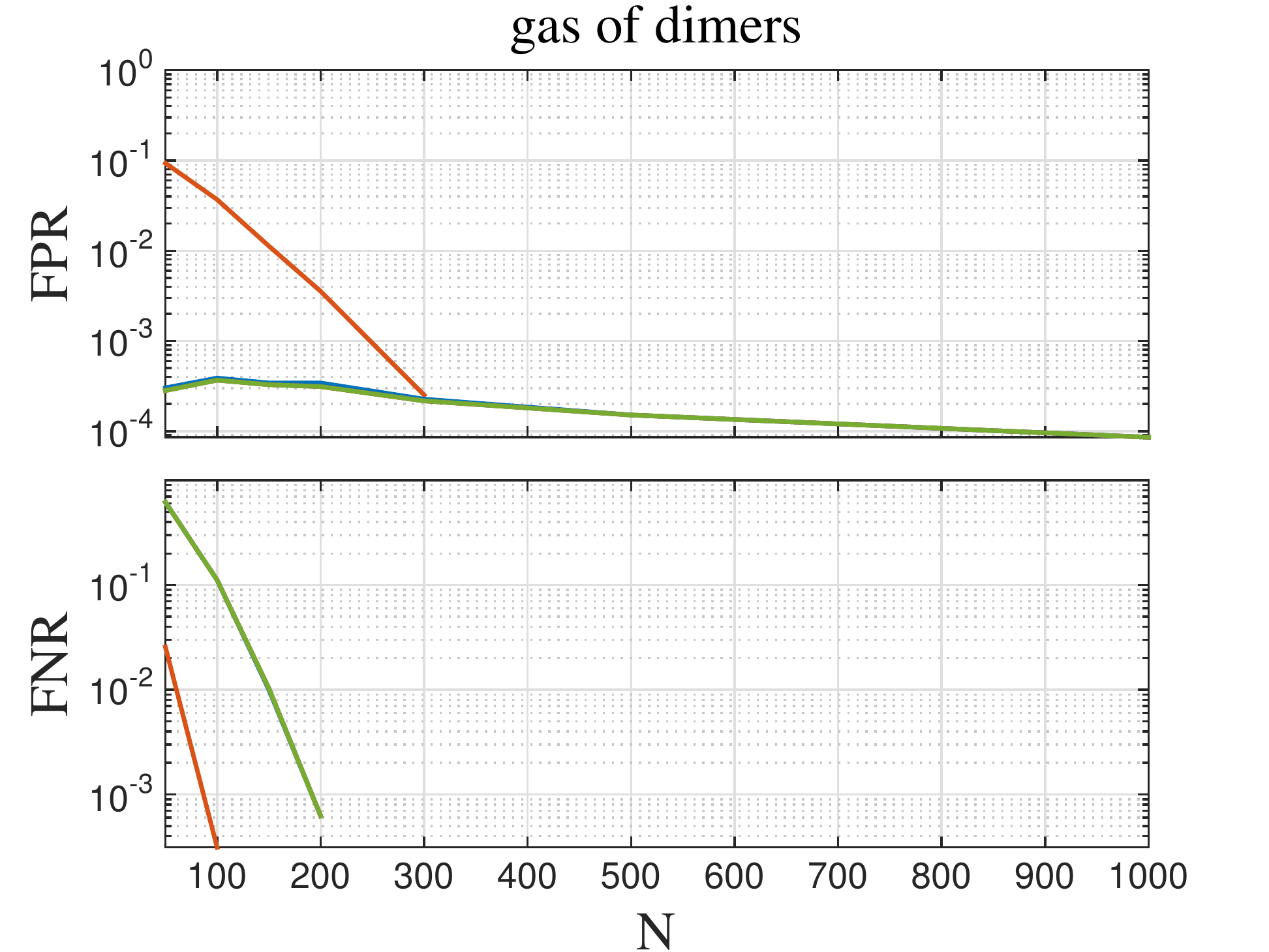}
\includegraphics[width = 0.45\textwidth]{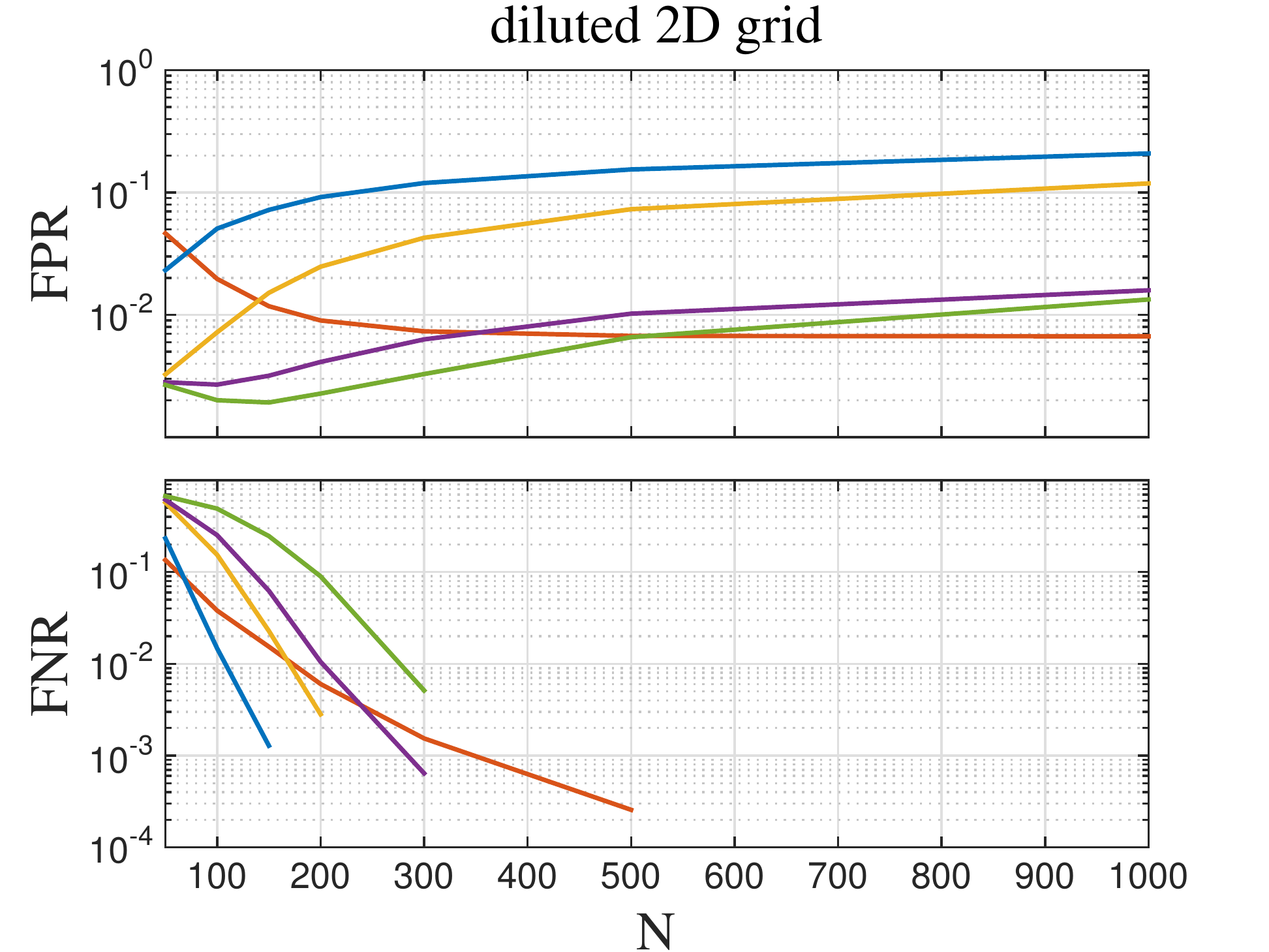}\vspace{2mm}\\
\includegraphics[width = 0.45\textwidth]{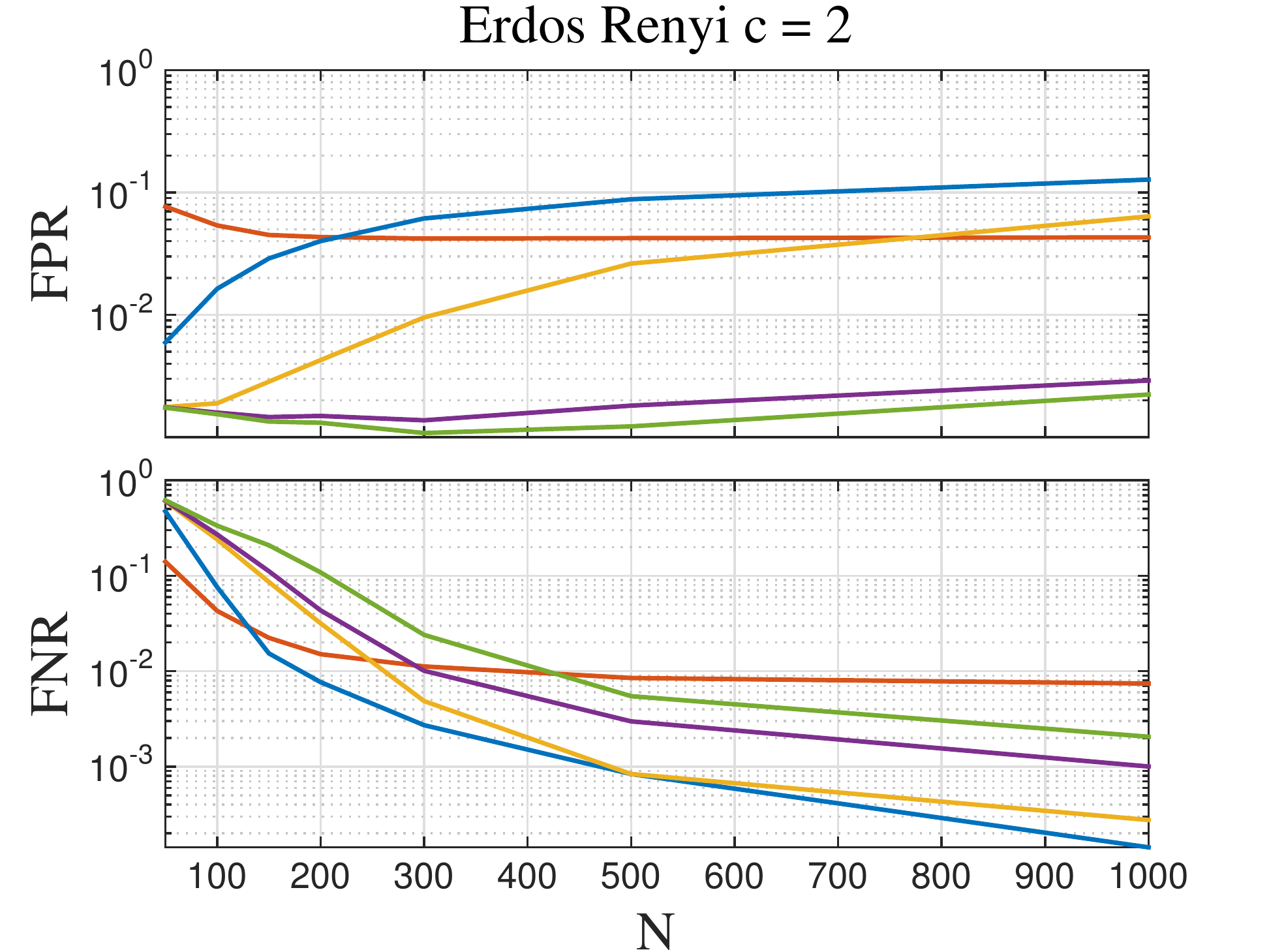}
\includegraphics[width = 0.45\textwidth]{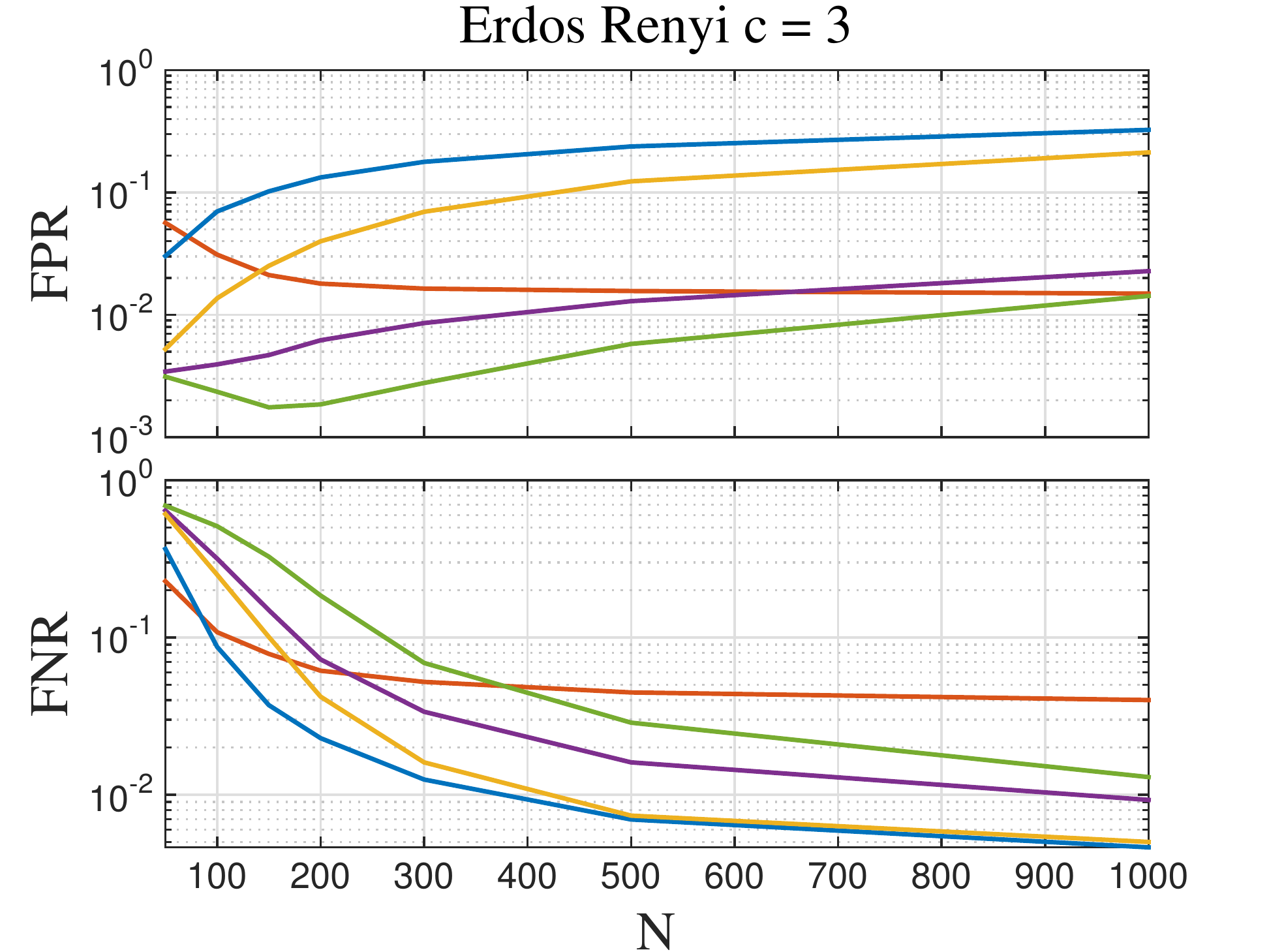}\vspace{2mm}\\
\includegraphics[width = 0.45\textwidth]{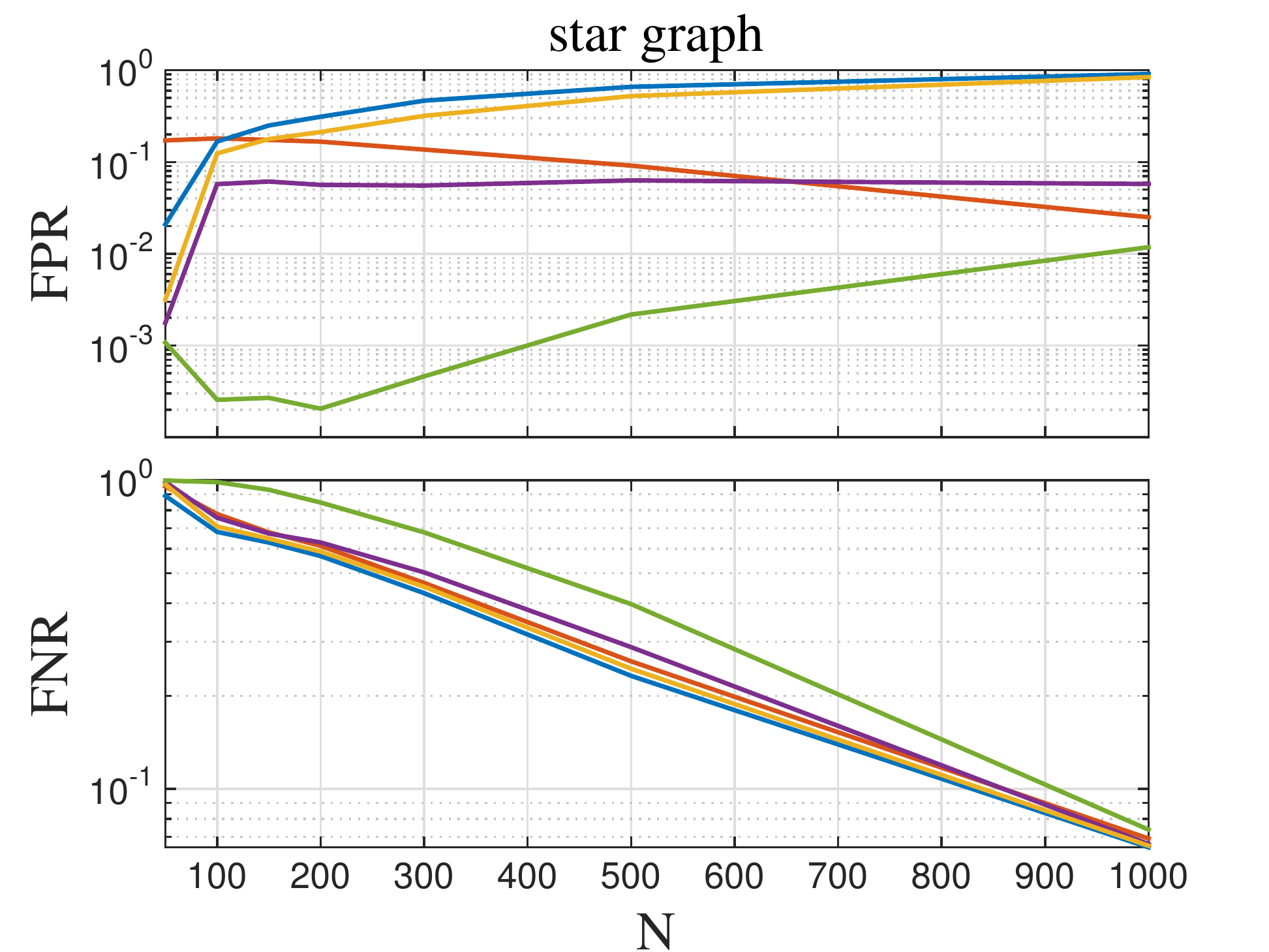}
\includegraphics[width = 0.45\textwidth]{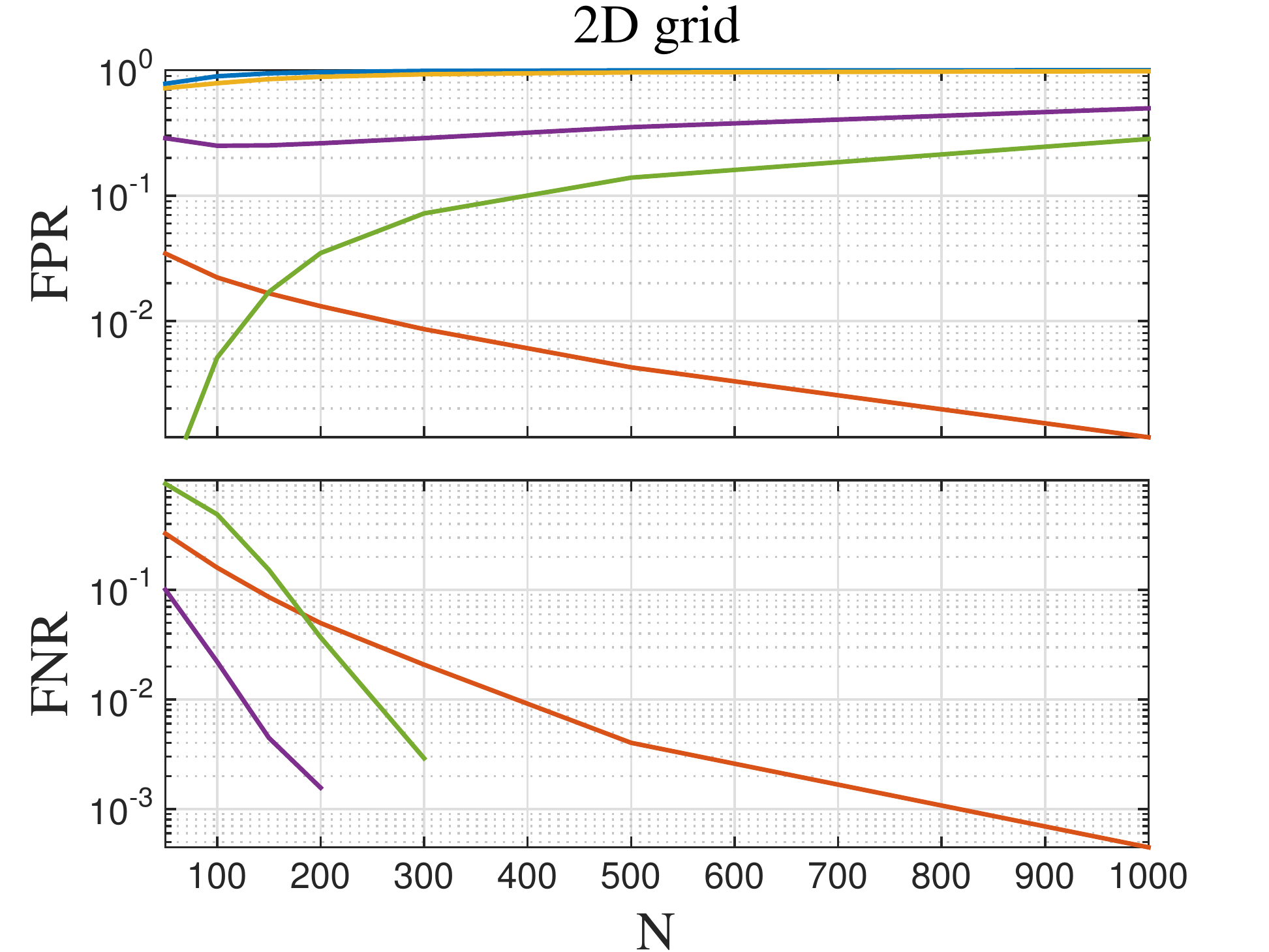}\vspace{2mm}\\
\caption{Simulations comparing the performances of MS (blue) and PLM+$\ell_1$ (red) with the ones of MS employing corrections for high quality data for all topologies investigated. The corrections are depicted in the graph with the following color code: MS avg (yellow), MS min (violet) and MS prod (green).}
\label{fig:ho_simulationsAppendix}
\end{figure}

As already pointed out in the paper and shown in detail in Fig.\ \ref{fig:scatterMsVsL1longAppendix}, when N becomes large enough the number of false positive becomes important and more evident in less sparse networks. Even though our approach is thought to be used in the highly under-sampled regime where only poor predictions are possible and one expects to find a very sparse matrix, one simple recipe is actually available to overcome this problem under the framework we have developed in the paper with a small additional computational cost. The main idea is already described in the paper in section $ III.B$, therefore here we just summarise it including a few further details and simulations. The recipe can be schematized as follows:
\begin{enumerate}
\item use MS with a self consistent thresholding procedure for recovering the graph. Typically for topologies containing many loops (e.g. a regular grid) or many conditional independencies (e.g. a star graph), the self consistent approach fails and retrieves a densely or fully connected graph, namely a self consistent point $r = \epsilon >1$. In the latter case, if no additional knowledge about the expected sparsity of the network is available, the choice of a uniform prior, $\epsilon =1$, represents the only possibility;
\item check for each couple of connected nodes if their neighbourhood sets share some vertices (i.e. find the number of loops of length three based on the considered connected couple);
\item in the latter case, given the two values of the confidence for the couple $(i,j)$ in the two sub-sets made up by conditioning on the value of a common neighbour $k$, i.e. $\eta_{ij|S_k = S}$, evaluate the difference between the probability of having a bond and the one of not having a bond in both the sub-sets as follows
\begin{equation}
\tilde{\eta}_{ij|S_k = S} = P(b|\hat{S},S_k = S) - P(nb|\hat{S},S_k = S) = \frac{\Gamma_k}{2\bar{\Gamma}_k}\left(\eta_{ij|S_k=S} - \frac{1-\epsilon}{1+\epsilon}\right).
\end{equation}
An unique estimate for the latter probability is given by the weighted average $\tilde{\eta}_{ij|k}= \sum_{S=\pm 1}\tilde{\eta}_{ij|S_k = S} \nu(S_k = S)$ where $\nu(S_k = S)$ represents the cardinality of the sub-sets.
\item After having calculated $\tilde{\eta}_{i,j|k}$ for each node $k$ belonging to the intersection set, ${\cal I}$, of the neighbourhoods of node $i$ and $j$, there are different ways to obtain a new estimate, $\tilde{\eta}^1_{ij}$, for the difference between the probability of having a bond and the one of not having a bond. Given the fact that couples are considered as being independent each other, the posterior probability of finding a node in the intersection set $\cal{I}$ of the neighbourhoods of $i$ and $j$, denoted as $P(\Delta_{ij}^k)$, is given by 

\begin{equation}
\displaystyle P(\Delta_{ij}^k) = \displaystyle\frac{P(b_{ik}|\hat{S})P(b_{jk}|\hat{S})}{\sum_{k\in {\cal I}} P(b_{ik}|\hat{S})P(b_{jk}|\hat{S})} \quad\mbox{with}\quad\displaystyle P(b|\hat{S}) = \displaystyle\frac{\Gamma}{2\bar{\Gamma}}\frac{\epsilon}{1+\epsilon}(1+\eta),
\end{equation}
where $P(b|\hat{S})$ corresponds to the posterior probability of having a bond. We tested the following three approaches:
\begin{description}
\item[MS avg]  $\displaystyle \tilde{\eta}^1_{ij} = \sum_{k \in {\cal I}} \tilde{\eta}_{i,j|k}\,P(\Delta_{ij}^k) \vspace{2mm}$ 
\item[MS min]  $\displaystyle \tilde{\eta}^1_{ij} = \min_{k \in {\cal I}} \left(\tilde{\eta}_{i,j|k}\right) \vspace{2mm} $
\item[MS prod] $\displaystyle \tilde{\eta}^1_{ij} = 2\prod_{k \in {\cal I}} \left(\frac{1+\tilde{\eta}_{ij|k}}{2}\right) - 1 \vspace{1mm}$
\end{description}
\end{enumerate}
According to the above prescriptions, $\tilde{\eta}_{ij|k} = 0$ represents the threshold above which the bond between spins $i$ and $j$ is considered genuine and not induced by the fact that the nodes share common neighbours. The results are shown in Fig.\ \ref{fig:ho_simulationsAppendix} for all investigated topologies. It is clear that this simple recipe allows for lowering considerably the false positive rate while keeping the false negative rate almost always below the PLM+ $\ell_1$ curve in the high $N$ regime. The three approaches exhibit different degree of selectivity being MS prod the most selective and MS avg the least one. Even in hard cases, e.g. star graph and 2D grid in Fig.\ \ref{fig:ho_simulationsAppendix}, these corrections proves their effectiveness in improving ordinary MS performance in the high $N$ regime and in doing sometimes even better than PLM+ $\ell_1$.

\section{\label{sec: SM2} Additional results on US stock market interactions}
As a first step in this section we are going to investigate the behaviour of the correlation functions $c^N_{ij}(t,\tau)$ at varying $\tau$. In particular we focused on the r.m.s of the diagonal and off-diagonal elements of the correlation matrices 
\begin{equation}
\displaystyle c^N_{diag}(t,\tau) = \sqrt{\frac{1}{N}\sum_{i = 1}^N {c^N_{ii}}^2(t,\tau)} \qquad \displaystyle c^N_{off}(t,\tau) = \sqrt{\frac{2}{N(N-1)}\sum_{i < j} {c^N_{ij}}^2(t,\tau)}
\end{equation}
as a measure of the mean magnitude of the diagonal and off-diagonal terms of the correlation matrices. 
\begin{figure}[t!]
\centering
\includegraphics[width = 0.49\columnwidth]{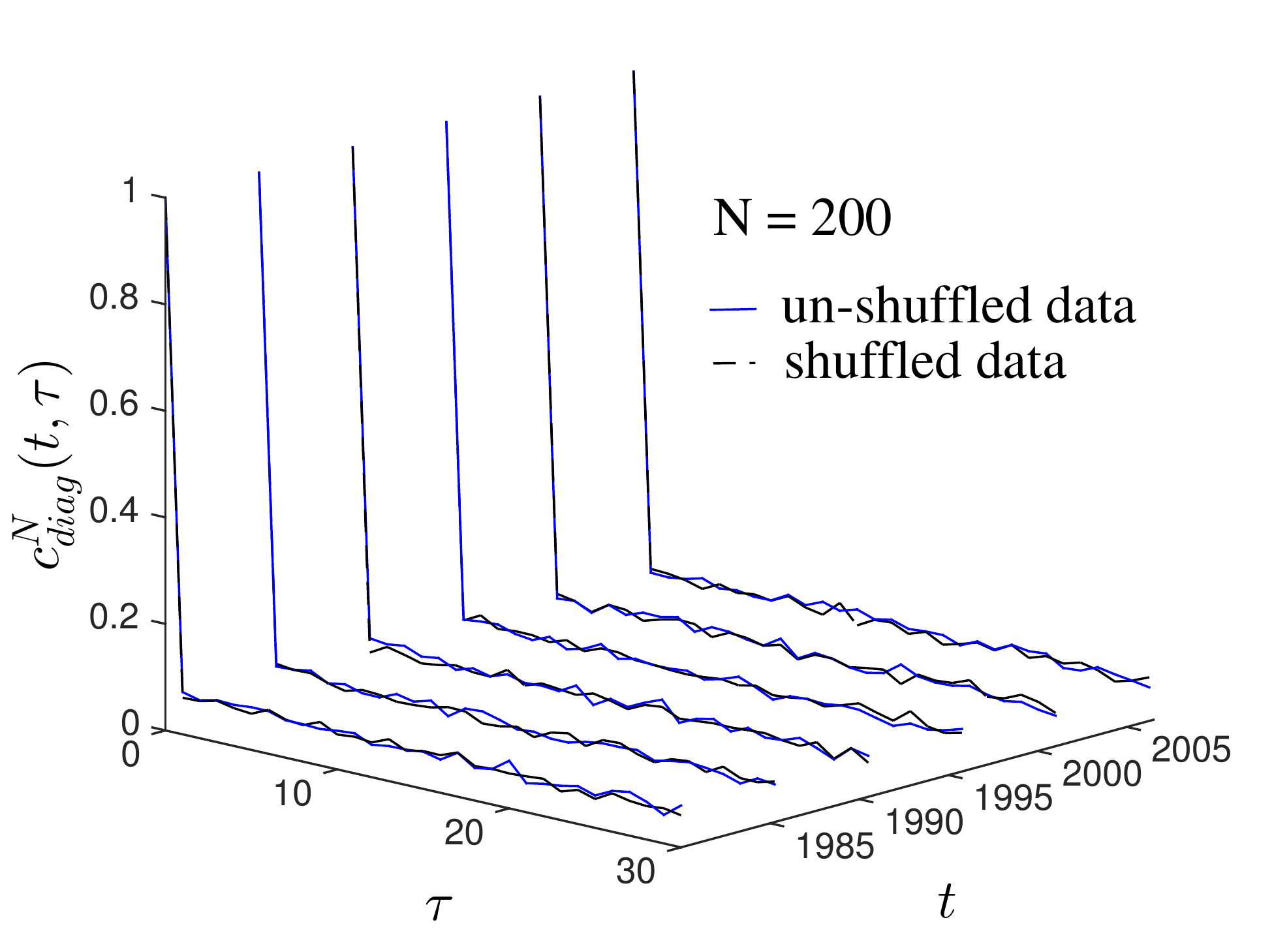}
\includegraphics[width = 0.49\columnwidth]{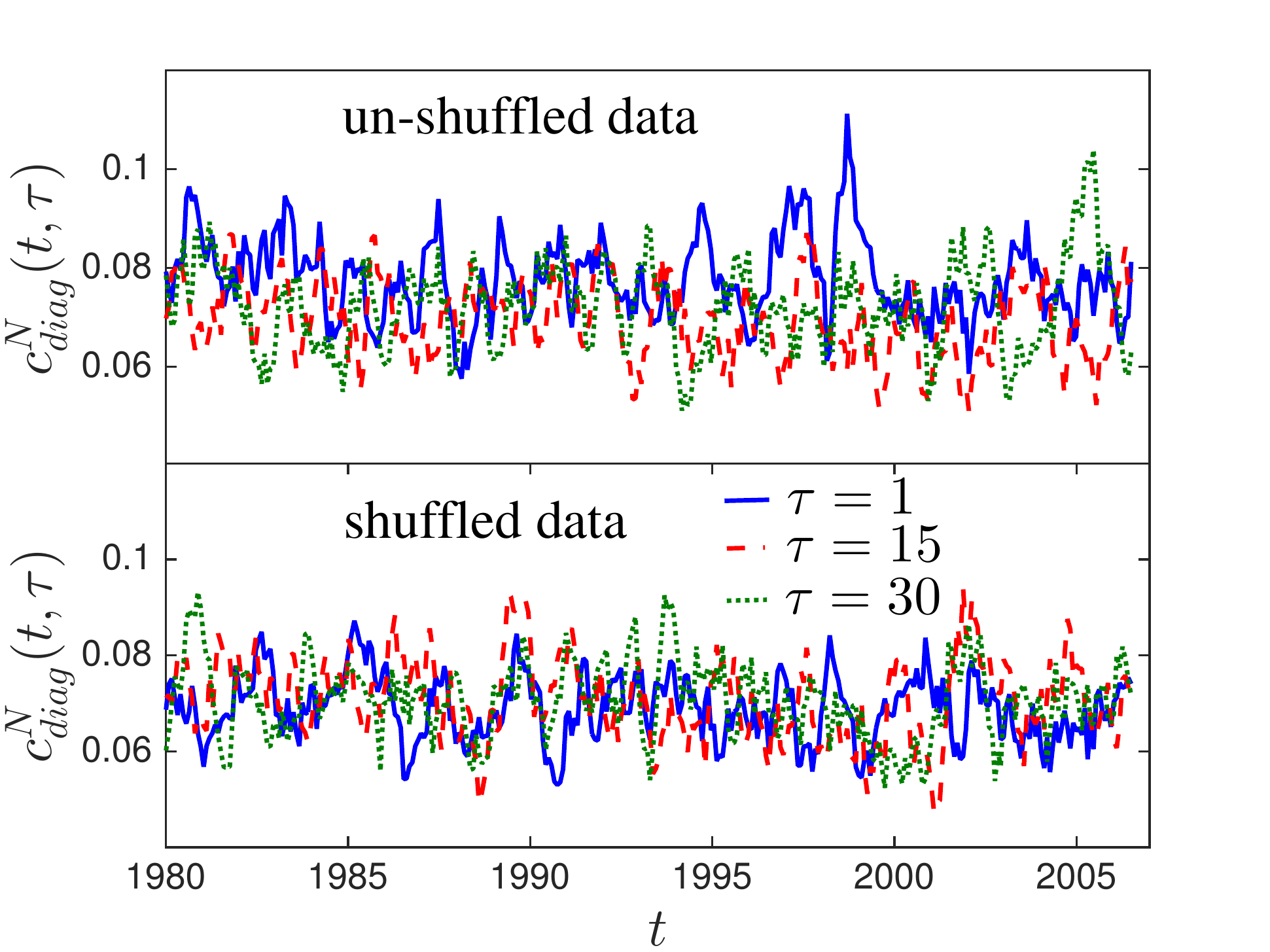}
\caption{$c^N_{diag}(t,\tau)$ versus $\tau$ and $t$ for $N=200$ and comparison with the same quantity calculated from a shuffled version of the data (left); $c^N_{diag}(t,\tau)$ versus $t$ at $N=200$ and $\tau =1,15,30$ (right on the top) and comparison with the same quantity calculated from a shuffled version of the data (right on the bottom)}.
\label{fig:tau_diag}
\end{figure}
\begin{figure}[h!]
\centering
\includegraphics[width = 0.49\columnwidth]{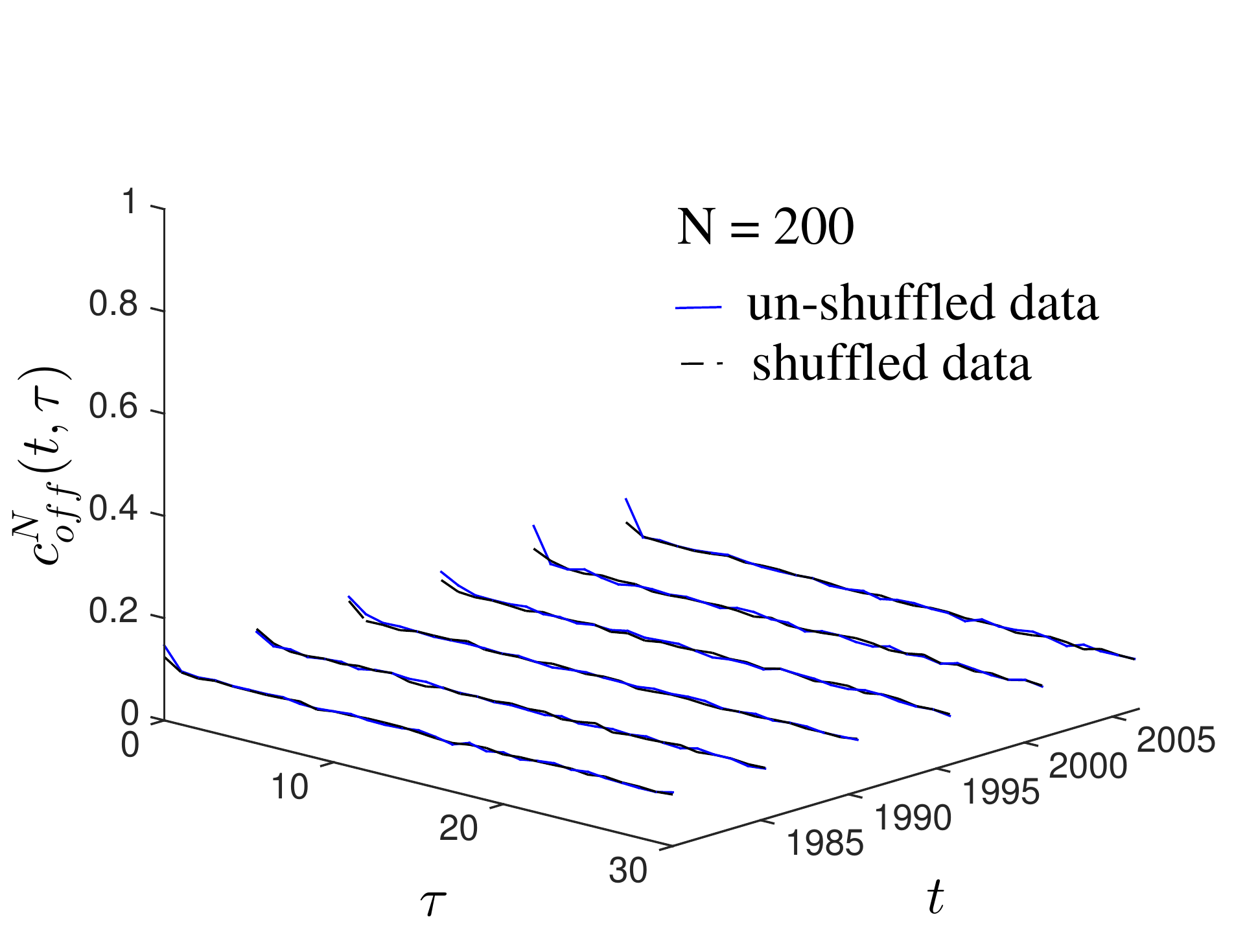}
\includegraphics[width = 0.49\columnwidth]{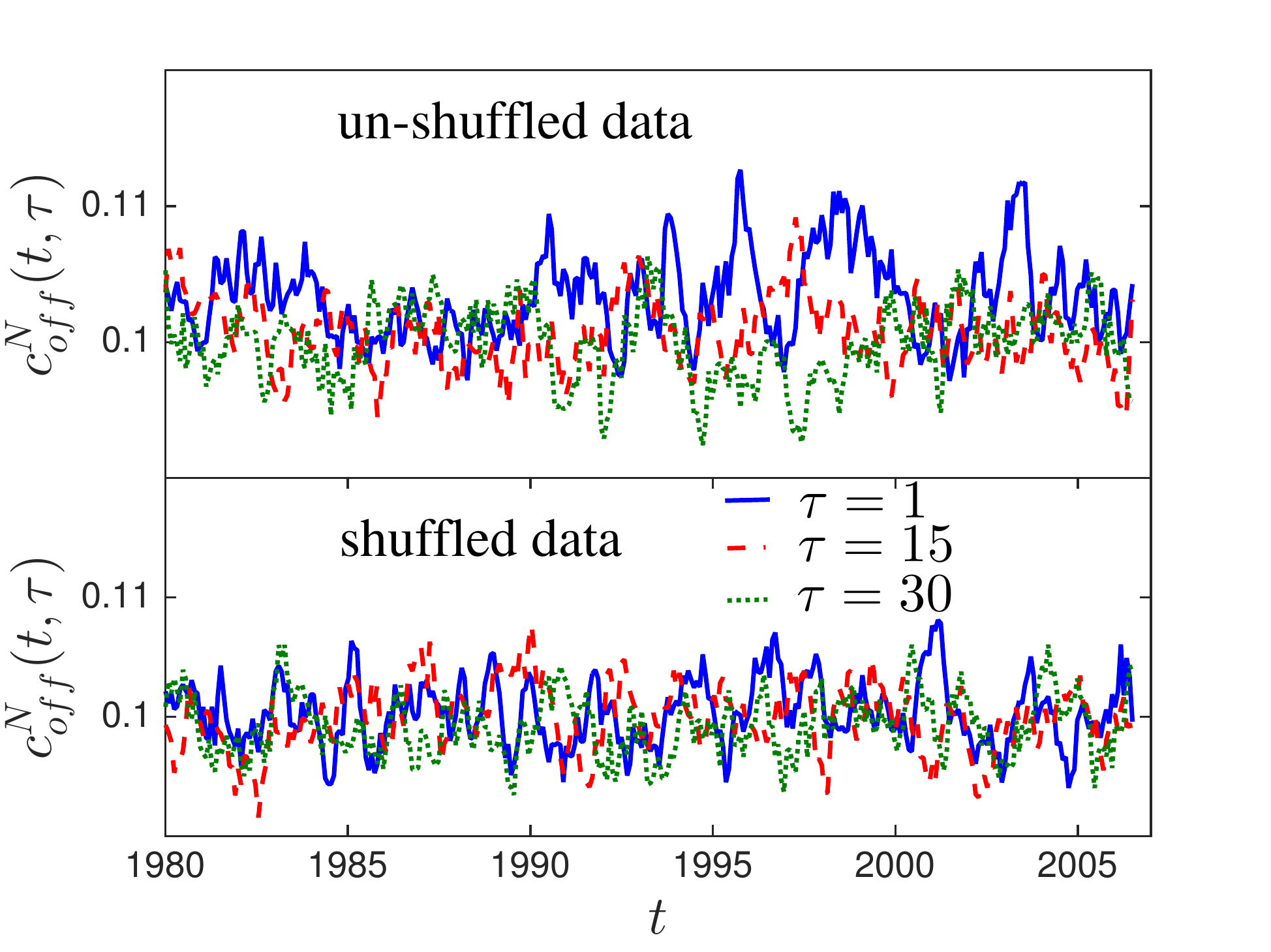}
\caption{$c^N_{off}(t,\tau)$ versus $\tau$ and $t$ for $N=200$ and comparison with the same quantity calculated from a shuffled version of the data (left); $c^N_{off}(t,\tau)$ versus $t$ at $N=200$ and $\tau =1,15,30$ (right on the top) and comparison with the same quantity calculated from a shuffled version of the data (right on the bottom)}.
\label{fig:tau_off}
\end{figure}
\begin{figure}[t!]
\centering
\includegraphics[width = 0.8\columnwidth]{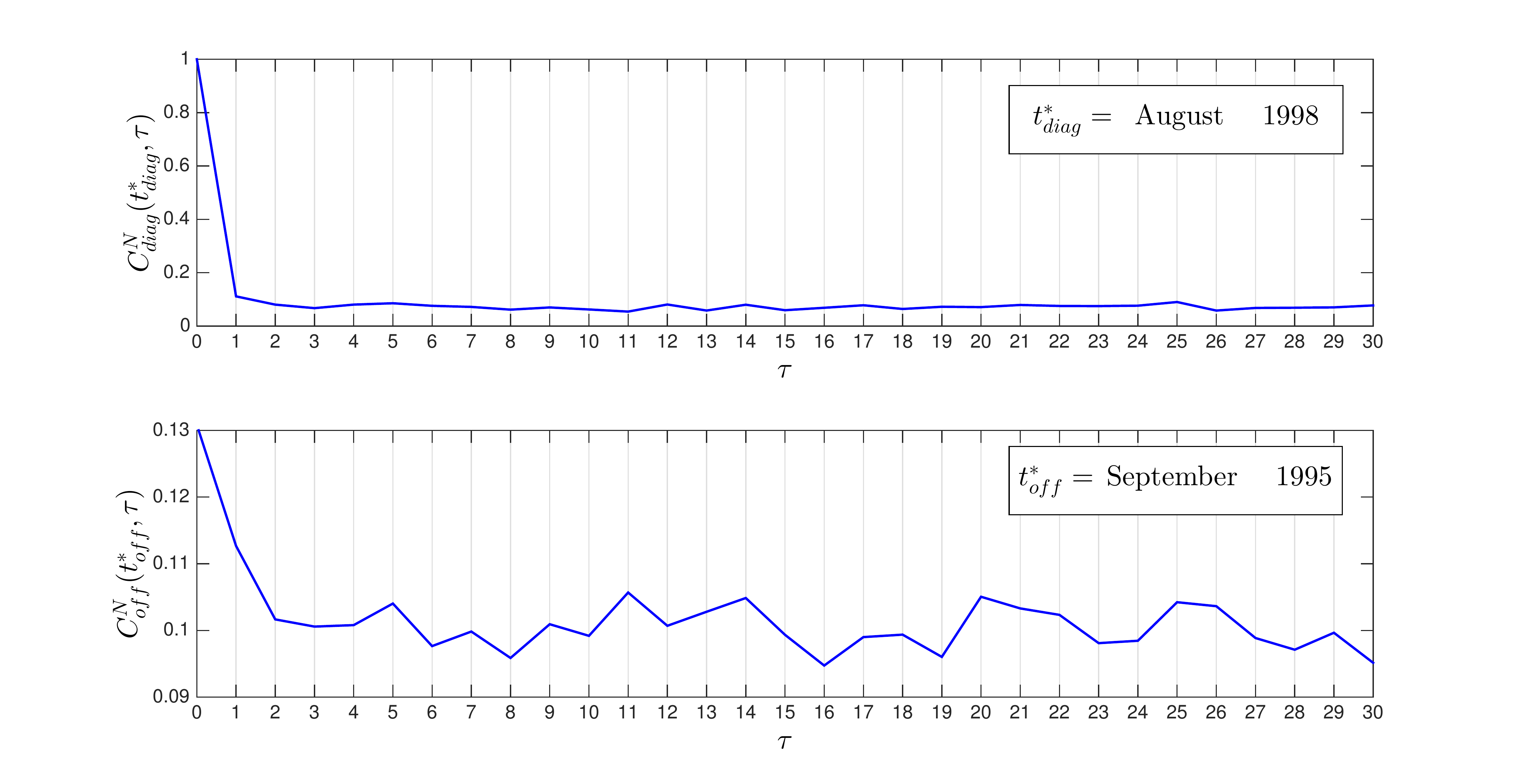}
\caption{Diagonal and off diagonal correlations versus the time delay $\tau$ in the highest correlated scenario}.
\label{fig:tau_example}
\end{figure}
The time delay $\tau$  and the size of the window $N$ are reported in trading days, whereas the reference point $t$ for the time window in years. 
In Figs.\ \ref{fig:tau_diag} and \ref{fig:tau_off}, we show $c^N_{diag}(t,\tau)$ and $c^N_{off}(t,\tau)$ and compare them with the corresponding values for a shuffled version of the data. As one can see from the left panels in these figures, both the diagonal and the off diagonal terms decay very quickly with $\tau$ for any given value of $t$ and at $N=200$: they reach the limiting plateau already for $\tau = 1 \mbox{ or } 2$ and then oscillate around it. The right panels in Figs.\ \ref{fig:tau_diag} and \ref{fig:tau_off}  display the fluctuating behaviour of $c^N_{diag}(t,\tau)$ and $c^N_{off}(t,\tau)$ as a function of $t$ for three increasing values of $\tau$, i.e. $\tau = 1,15\mbox{ and } 30$, showing that the fluctuations for $\tau = 1$ have in general almost the same magnitude as those for $\tau = 15 \mbox{ and } 30$ although they seem to be slightly bigger in some time regions especially in the case of $c^N_{off}(t,\tau)$. This behaviour suggests that on average there is no memory in the data, i.e. the configurations at time $t$ are uncorrelated with those at time $t+\tau$ already for $\tau = 1$, although in certain regions a short memory term seems to be present. This conclusion is further supported by comparing the correlations between the shuffled and un-shuffled data: a shuffling of the dataset provides a new dataset in which the configurations at different times $t$ are uncorrelated. This comparison shows that the average magnitude of the fluctuations for the shuffled and un-shuffled datasets are almost the same and that, as already said above, exceptions, i.e. bigger fluctuations in the un-shuffled dataset for $\tau = 1$, are limited to some time windows. In the latter cases the configurations of returns at time $t$ are weakly correlated with those at time $t+1$  but uncorrelated with the configurations at time $t+\tau$ for $\tau \geq 2$. As an example of the highest time correlated scenarios, we plot in Fig. \ref{fig:tau_example} $c^N_{diag}(t_{diag}^{*},\tau)$ and $c^N_{off}(t_{off}^{*},\tau)$ as a function of $\tau$ for two values of $t$, $t_{diag}^{*}$ and $t_{off}^{*}$ that represent the times at which $c^N_{diag}(t,\tau = 1)$ and $c^N_{off}(t,\tau = 1)$ assume their maximum values, namely the values corresponding to the biggest fluctuations of the blue line in the right top panels of Figs.\ \ref{fig:tau_diag} and \ref{fig:tau_off}. As one can see in this extreme case, both diagonal and off diagonal terms seem to be correlated for $\tau = 1$ but then the correlations decay and fluctuate around a plateau for $\tau > 1$. In Fig.\ref{fig:tau_T}, we show the correlations for three different values of the time window $N=100,200,300$ noting the same same general behaviour for the fluctuations: the magnitude of fluctuations for $\tau = 1$ are almost the same or just a little bit bigger in certain regions than the ones for $\tau = 15$ and $\tau = 30$ meaning that the configurations decorrelate very quickly in time. Therefore, this trend seems to be preserved when varying the size of the window.

\begin{figure}[t!]
\centering
\includegraphics[width = \columnwidth]{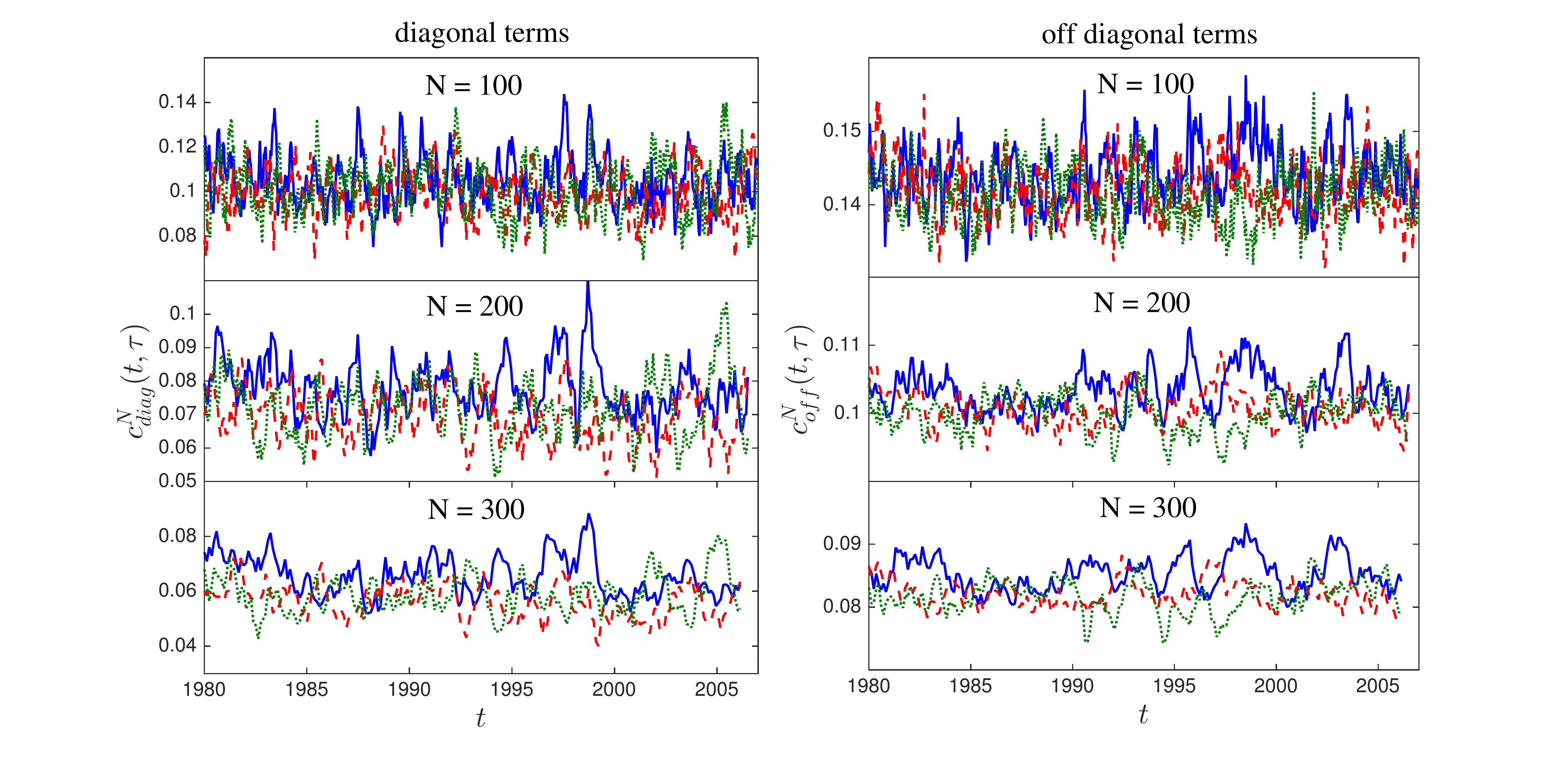}
\caption{Same figures as the right insets of Fig.\ \ref{fig:tau_diag} and \ref{fig:tau_off} for different values of the time window $N = 100,200,300$}.
\label{fig:tau_T}
\end{figure}

\begin{figure}[t!]
\centering
\includegraphics[width = 0.32\columnwidth]{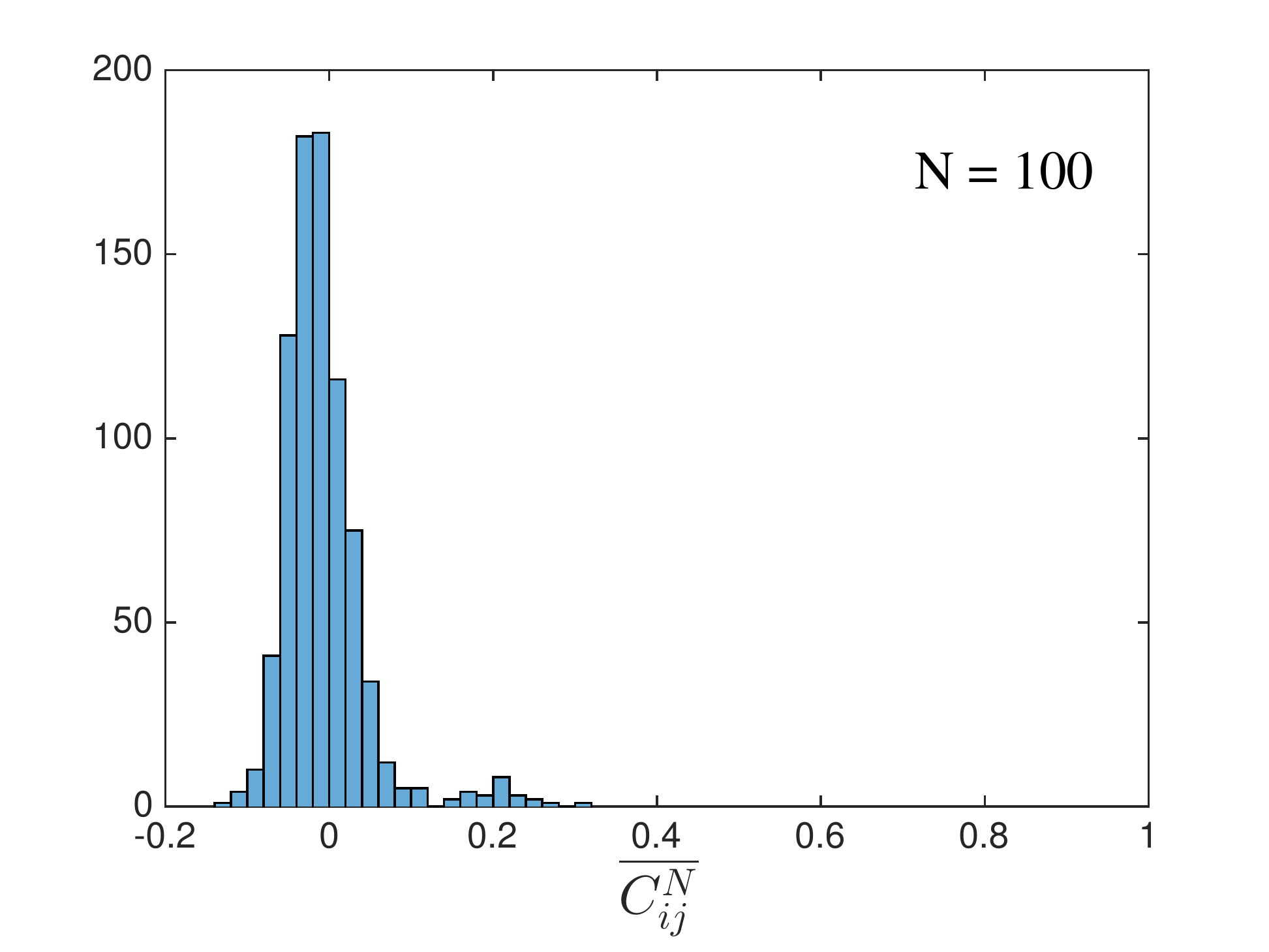}
\includegraphics[width = 0.32\columnwidth]{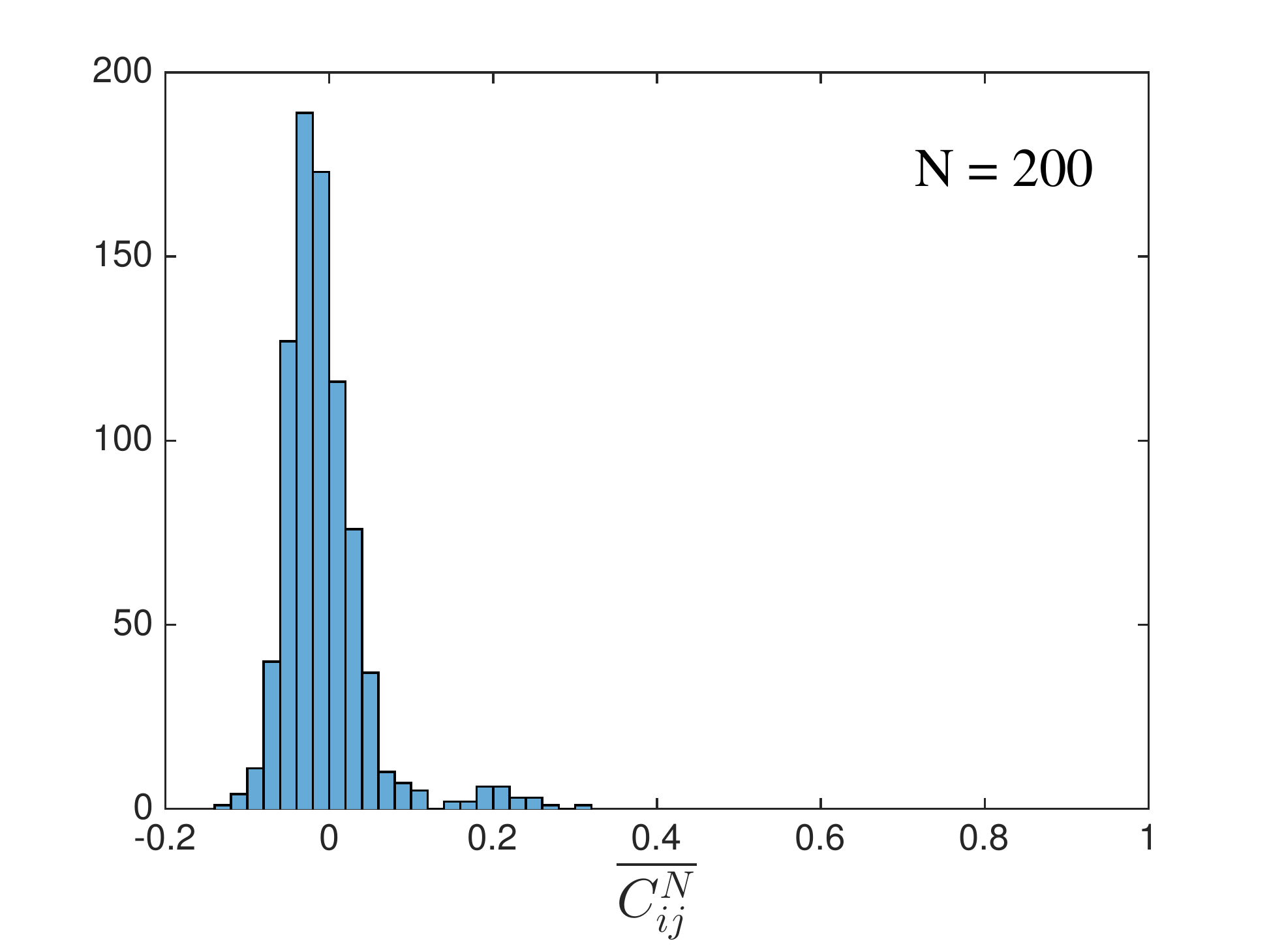}
\includegraphics[width = 0.32\columnwidth]{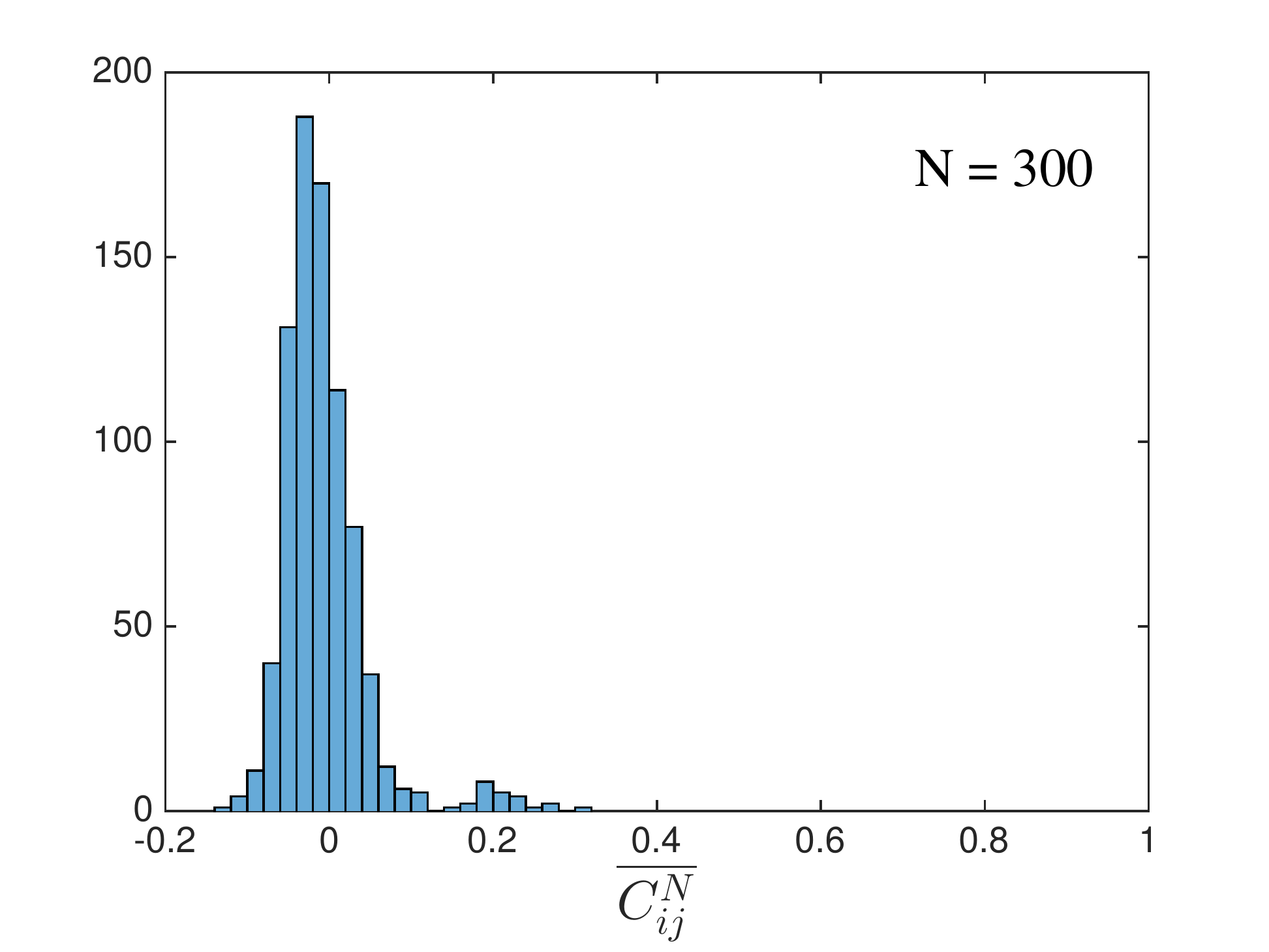}
\caption{Histograms of the time averaged connected correlation functions $\overline{C^N_{ij}}$ for each couple for three values of $N = 100,200,300$ showing a big peak around zero and a small one a bit separated representing couples which tend to be more correlated than others}.
\label{fig:histograms}
\end{figure}

\begin{figure}[h!]
\centering
\includegraphics[width=0.49\columnwidth] {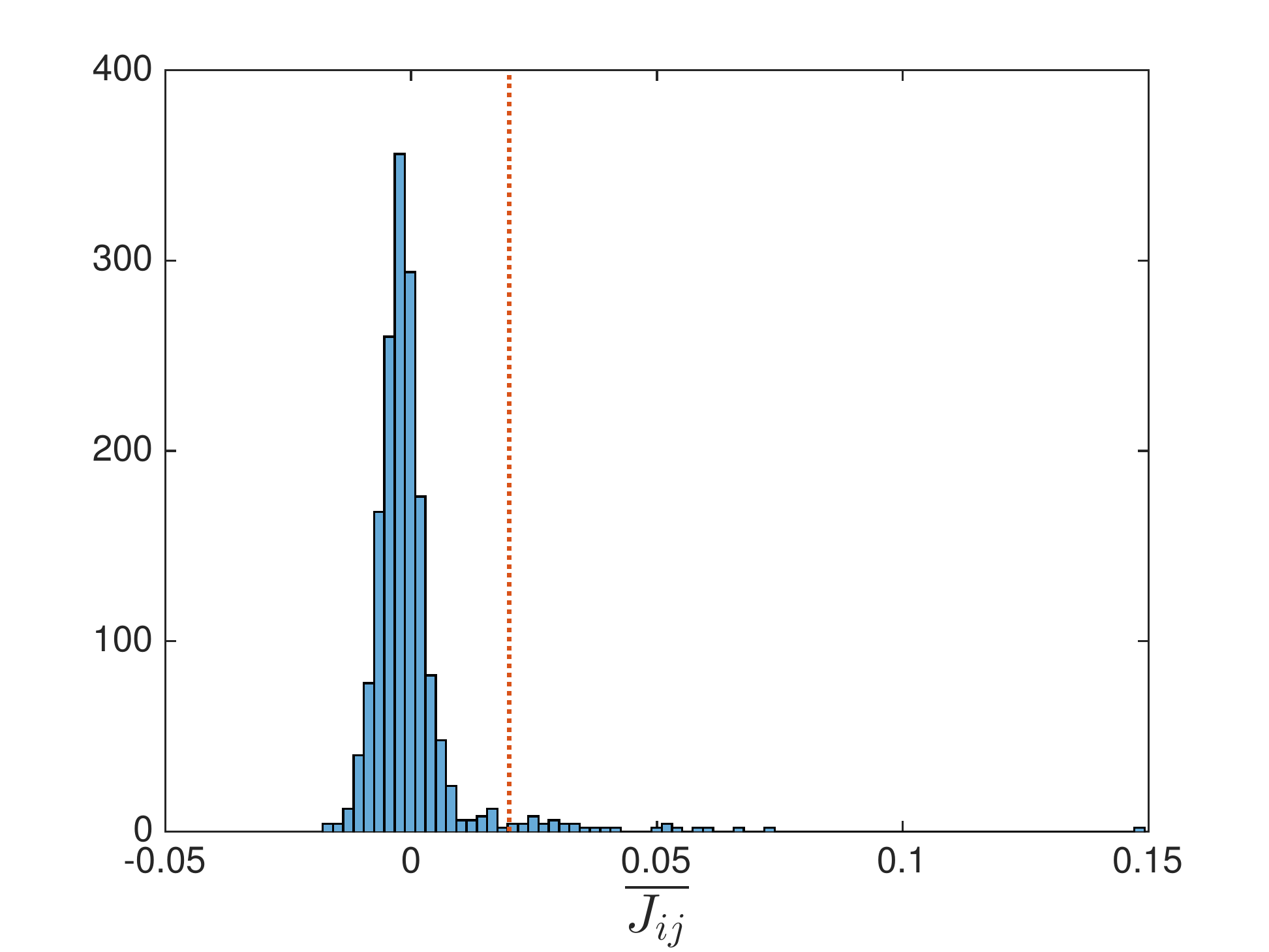}
\includegraphics[width = 0.49\columnwidth]{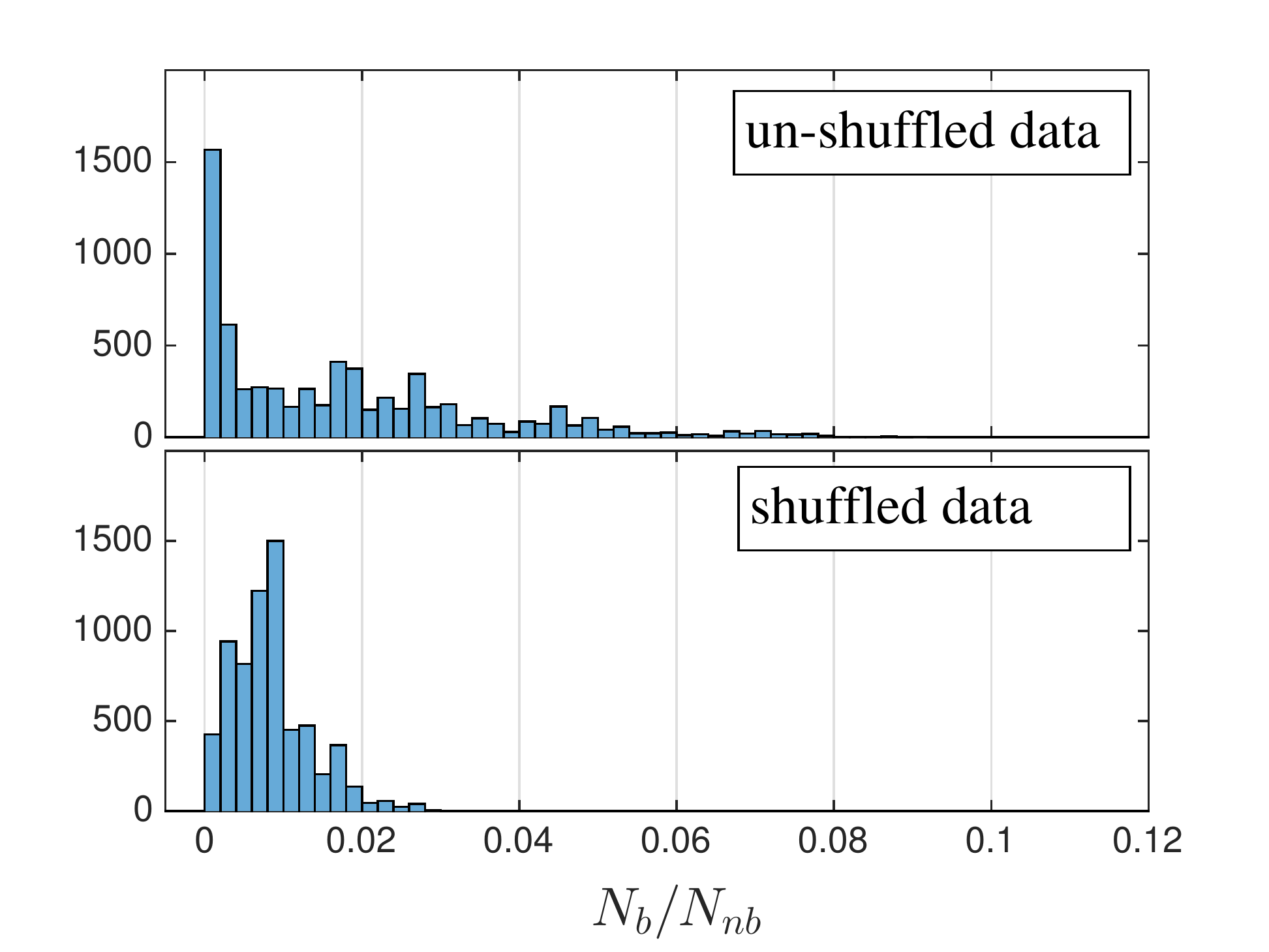}
\caption{Histogram of time averaged inferred couplings $\overline{J_{ij}}$ showing a long tail for positive values of $\overline{J_{ij}}$ (left). Since a clear gap is not evident here and in order to obtain an `equilibrium graph' as in the other cases, we thresholded $\overline{J_{ij}}$ at 0.02 (red dotted line) since qualitatively below that value the distribution seems to be symmetric and centered around zero; histograms of the `bond'-`no-bond' ratio with a self consistent approach (right top) and its counterpart from a time shuffled version of the data (right bottom)}.
\label{fig:J_L1}
\end{figure}

Given the finding that the returns are largely uncorrelated for $\tau \neq 0$, we then focused on equal time connected correlations $C^N_{ij}(t)$. Looking at the evolution of $C^N_{ij}(t)$ for each pair of spins $i,j$ we noticed that, for the majority of the pairs, the connected correlation fluctuates in time around some small value close to zero while for a few of them the connected correlations take on average values different from zero. Indeed looking at the histogram of the time averaged connected correlation functions $\overline{C^N_{ij}}$ for each pair shown in Fig.\ \ref{fig:histograms}, one can distinguish a big peak around zero and a small but separate one centered at a positive value. Performing a shuffling in time and building again the previous histograms does not change their shape \footnote{This is not obvious since a shuffling in time changes data points inside each time window. Therefore the fact that we found the same histograms enforces our finding of pairs more correlated than others in time}. These couples, exhibiting a bigger average values of the connected correlations, correspond exactly to those which are recognised by MS as the ones which are most often linked by a direct connection (see paper). The same connections are also recognised by PLM+$\ell_1$ and make up the long tail in the histogram of the average values of the inferred couplings (see left inset of Fig.\ \ref{fig:J_L1}). Finally changing the size of the window does not influence the `equilibrium' graph we got: in other words, this division between more and less correlated pairs holds for the different investigated time scales. Above this `equilibrium' graph, time varying fluctuations occur and determine the temporal evolution. An effect of non stationarity in the data is highlighted in Fig.\ \ref{fig:J_L1} (right) for $N = 200$: the long tail in the histogram of the `bond'-`no-bond' ratio is suppressed after a shuffling of the dataset which spreads uniformly the time variability along the entire dataset. The clear modification of the histogram after the shuffling is a strong evidence of the presence of time dependent fluctuations in the data.
 
\begin{figure}[t!]
\centering
\includegraphics[width = 0.32\columnwidth]{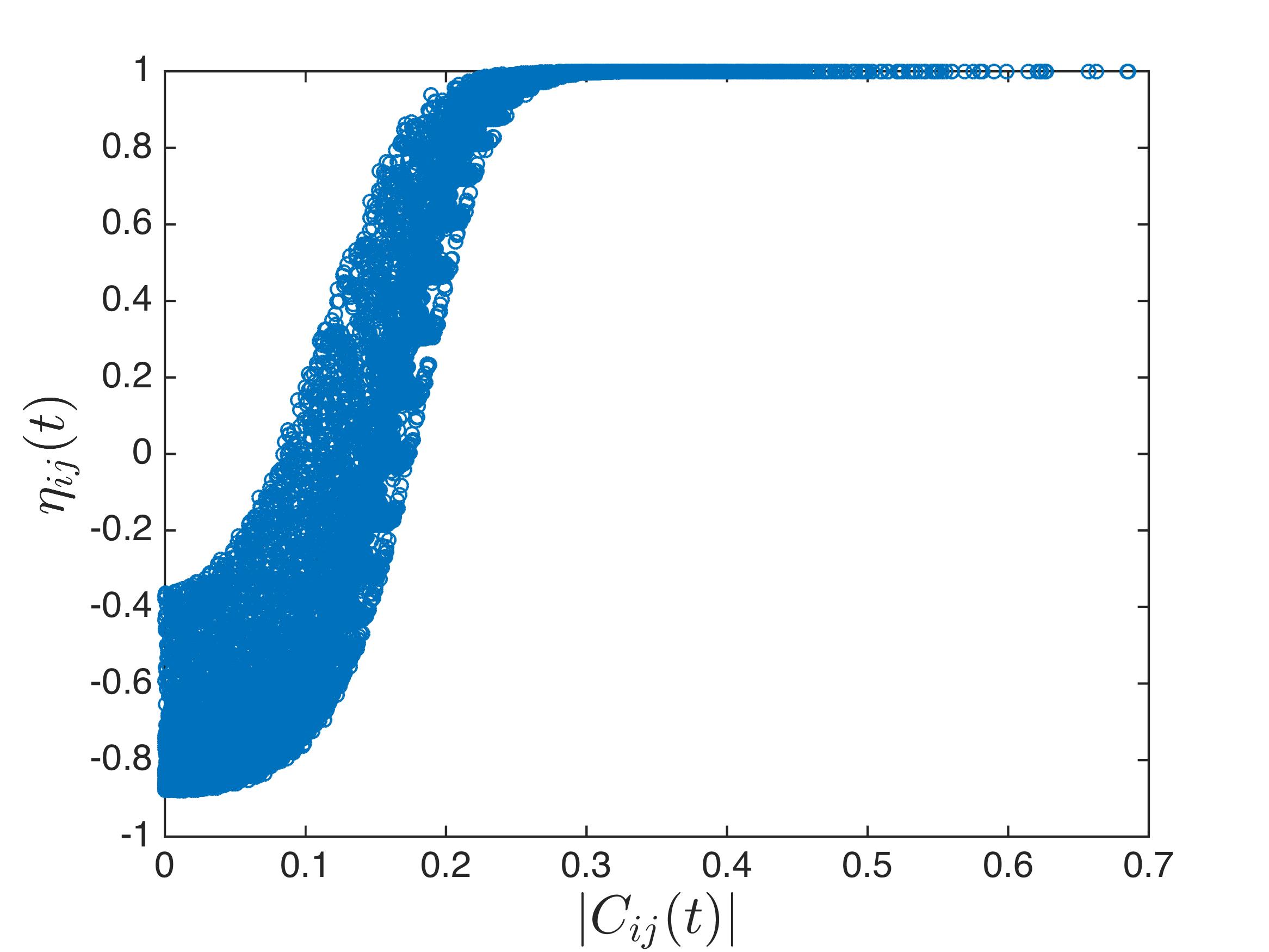}
\includegraphics[width = 0.32\columnwidth]{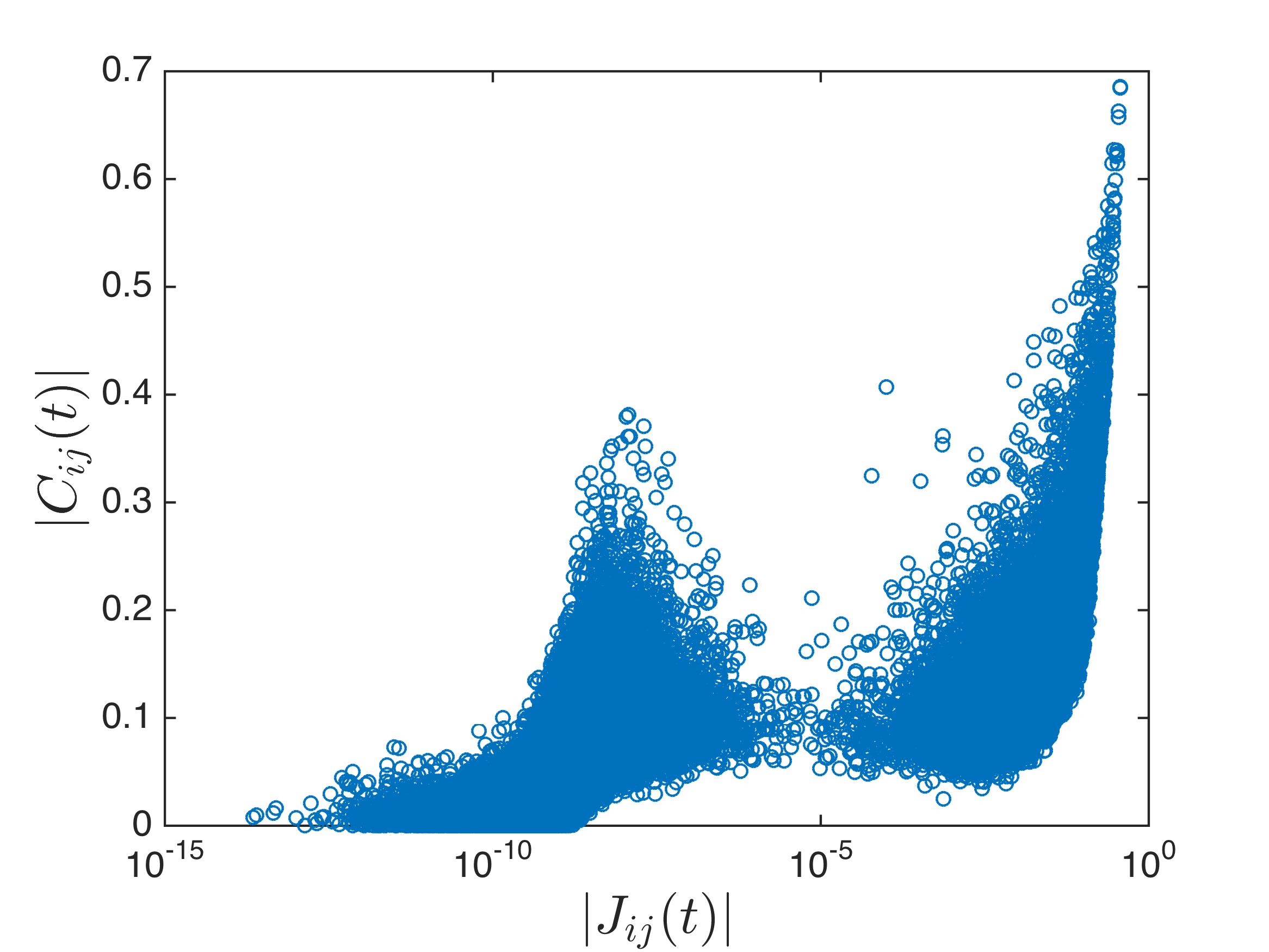}
\includegraphics[width = 0.32\columnwidth]{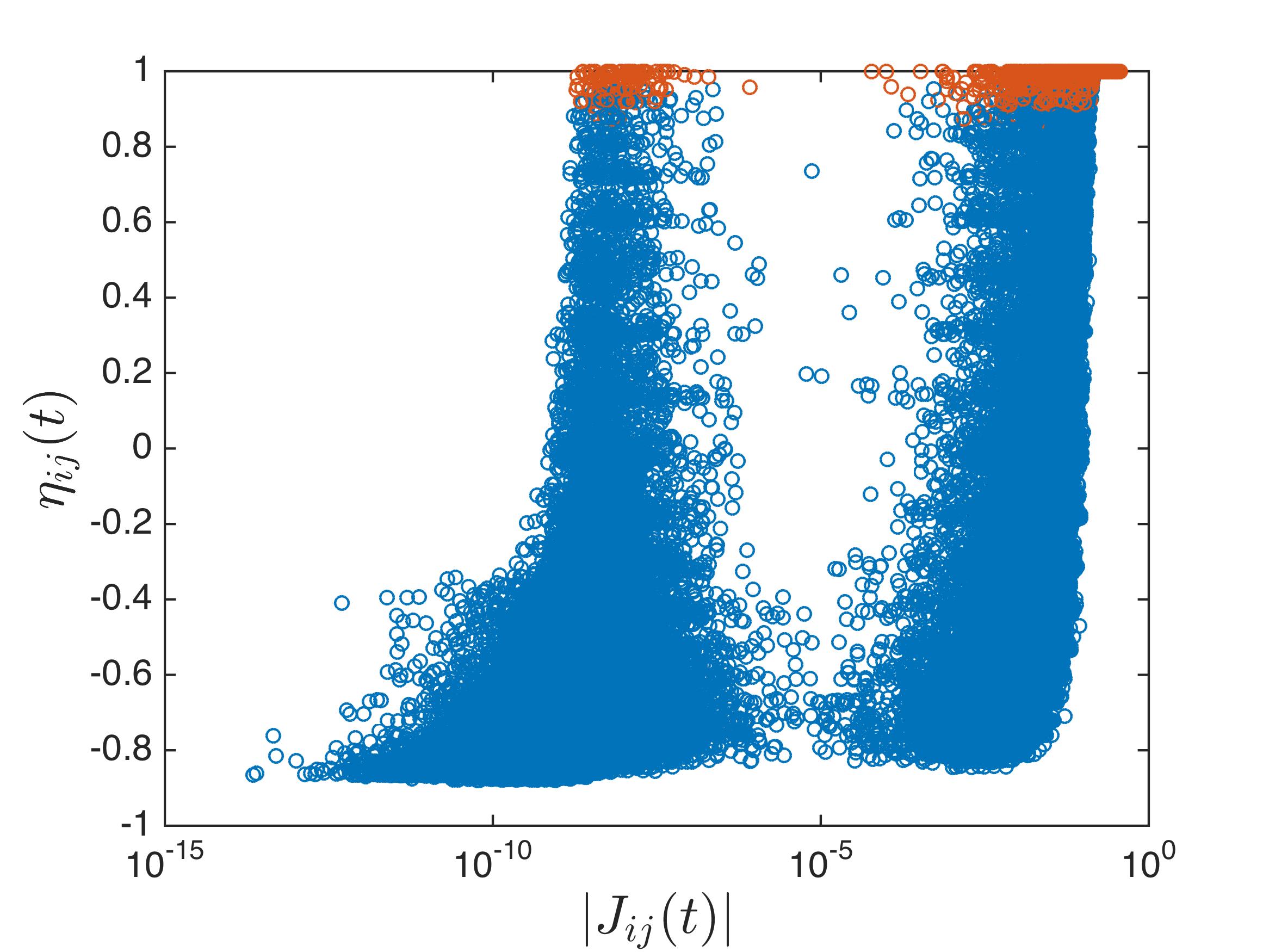}
\caption{Relation between $\eta_{ij}(t)$ and the absolute values of connected correlations $C^N_{ij}(t)$ while varying $i,j$ and $t$ (left); same plot showing the relation between $|C_{ij}(t)|$ and $|J_{ij}(t)|$ (centre); A comparison between MS reconstructions and PLM+$\ell_1$ reconstructions for several values of $t$ along the dataset. The red circles are pairs found to be directly connected by the self consistent approach. The $94\%$ of them are located beyond the gap along the $|J_{ij}(t)|$ axis meaning that they are also recognised as direct connections by the PLM+$\ell_1$ approach (right).}.
\label{fig:correlations_versus_others}
\end{figure}

A closer comparison between the absolute values of the connected correlations and the outcomes of the two methods under investigation is shown in Fig.\ \ref{fig:correlations_versus_others} for $N=200$. In particular, as one can appreciate by looking at left panel of Fig.\ \ref{fig:correlations_versus_others}, MS approach can be understood as a quantitative and unambiguous way of thresholding connected correlations. The relation between $J_{ij}$ and the connected correlations is also depicted in Fig.\ \ref{fig:correlations_versus_others} (central panel). PLM+$\ell_1$ technique creates a gap along the $|J_{ij}|$ axis and the pairs with a value of $|J_{ij}|$ beyond that gap are assigned a bond, while the others are considered to be disconnected. Therefore looking at the figure, not surprisingly, it turns out that higher values of correlations $|C_{ij}(t)|$ mean direct connections, very low ones are assigned to disconnected pairs and values in between belong to a transition region ($0.1\lesssim |C_{ij}(t)| \lesssim 0.3$ ) whose width is more or less the same as the one observed in the MS case. However, in the PLM+$\ell_1$ case the relation between the absolute values of the inferred couplings and the ones of the connected correlations is not as simple as in the MS case, and its complexity reflects the more involved recipe underlying PLM+$\ell_1$ technique. Finally, as already discussed in the paper, PLM+$\ell_1$ typically recovers more bonds than MS, but $94\%$ of the direct connections retrieved with MS using the self consistent approach are also recognized by PLM+$\ell_1$ as direct connections. This is shown in the right panel of Fig.\ \ref{fig:correlations_versus_others} and this further emphasizes the rather conservative and relevant predictions obtained with MS.

Finally in Tab.\ \ref{table:stocks}, we report here the list of the 41 stocks in the Dow Jones index along with their economic sector. Stocks belonging to different connected components of the equilibrium graph inferred with MS are highlighted with different colours.

\begin{table}[t]
\begin{minipage}{\textwidth}   
\begin{center}
\tiny
  \begin{tabular}{ |c |c |c |c|| }
    \hline
    \textbf{Ticker} & \textbf{Name} & \textbf{Sector} & \textbf{Industry} \\ \hline\hline
    AA & Alcoa Inc & Materials & Aluminium \\ \hline
    \rowcolor{green} PEG & Public Serv. Enterprise & Utilities & Electric Utilities \\ \hline
    \rowcolor{green} AEP & American Electric Power & Utilities & Electric Utilities \\ \hline
    AXP & American Expess co & Financials & Consumer Finance \\ \hline  
    CNW & Con-Way Inc & Services & Trucking \\ \hline
    \rowcolor{lightgray} AMR & American Airline Group Inc & Industrials & Airlines \\ \hline
    BA & Boeing Company & Industrials & Aereospace and Defence \\ \hline
    BNI & Burlington Northern Santa Fe & Industrials & Rail Freight \\ \hline
    CAT & Caterpillar Inc & Industrials & Farm and Construction Machinery\\ \hline
    C & Citigroup Inc & Financials & Banks \\ \hline
    \rowcolor{green} CNP & Center Point Energy & Utilities & Multiutilities \\ \hline
    \rowcolor{yellow} CVX & Chevron Corp & Energy & Integrated Oil and Gas \\ \hline
    DD & E. I. du Pont de Nemours and Company & Basic Materials & Agricultural Chemicals \\ \hline
    DIS & The Walt Disney Company & Services & Entertainment - Diversified \\ \hline
    DOW & The Dow Chemical Company & Basic Materials & Chemicals - Major Diversified \\ \hline
    \rowcolor{green} ED & Consolidated Edison, Inc & Utilities & Electric Utilities \\ \hline
    \rowcolor{green} EIX & Edison International  & Utilities & Electric Utilities \\ \hline
    EK & Eastman Kodak Co. & Consumer Goods & Electronic Equipment \\ \hline
    \rowcolor{green} EXC &Exelon Corporation & Utilities & Diversified Utilities \\ \hline
    FDX & FedEx Corporation & Services & Air Delivery and Freight Services \\ \hline
    FO & Fortune Brands, Inc. & Consumer Goods & Home and Security, Spirits, and Golf \\ \hline
    GE & General Electric Company & Industrial Goods & Diversified Machinery \\ \hline
    GM & General Motors Company & Consumer Goods & Auto Manufacturers - Major \\ \hline
    GT & The Goodyear Tire and Rubber Company & Consumer Goods & Rubber and Plastics \\ \hline
    HON & Honeywell International Inc. & Industrial Goods & Diversified Machinery \\ \hline
    HPQ & Hewlett-Packard Company & Technology & Diversified Computer Systems \\ \hline
    IBM & International Business Machines Corporation & Technology & Information Technology Services \\ \hline
    IP & International Paper Company & Consumer Goods & Packaging and Containers \\ \hline
    \rowcolor{cyan} JNJ & Johnson $\&$ Johnson & Healthcare & Drug Manufacturers - Major \\ \hline
    KO & The Coca-Cola Company & Consumer Goods & Beverages - Soft Drinks \\ \hline
    \rowcolor{lightgray} LUV & Southwest Airlines Co. & Services & Regional Airlines \\ \hline
    MCD & McDonald's Corp. & Services & Restaurants \\ \hline
    MMM & 3M Company & Industrial Goods & Diversified Machinery \\ \hline
    MO & Altria Group Inc. & Consumer Goods & Cigarettes \\ \hline
    \rowcolor{cyan} MRK & Merck $\&$ Co. Inc. & Healthcare & Drug Manufacturers - Major \\ \hline
    NAVZ & Navistar International Corporation & Consumer Goods & Trucks and Other Vehicles \\ \hline
    \rowcolor{green} PCG & PG$\&$E Corporation & Utilities & Electric Utilities \\ \hline
    PG & The Procter $\&$ Gamble Company & Consumer Goods & Personal Products \\ \hline
    UTX & United Technologies Corporation & Industrial Goods & Aerospace/Defense Products $\&$ Services \\ \hline
    WMT & Wal-Mart Stores Inc. & Services & Discount, Variety Stores \\ \hline
    \rowcolor{yellow} XOM & Exxon Mobil Corporation & Basic Materials & Major Integrated Oil and Gas \\ \hline
  \end{tabular}
\end{center}
\end{minipage}
\caption{List of the 41 stocks in the Dow Jones index along with their economic sector. Stocks belonging to different component in the equilibrium graph of Fig.\ \ref{fig:supportMS} inferred with MS (thick lines) are highlighted with different colours. }
\label{table:stocks}
\end{table}

\clearpage
\section{A Matlab implementation of the MS method}

Here we include a simple Matlab code for a direct implementation of our approach. The function receives as an input the mean activities of two nodes and their correlation and gives back the value of the confidence.

\begin{lstlisting}
% FUNCTION: MODEL SELECTION DISCRIMINATOR
% INPUT: MEAN ACTIVITY, CORRELATION AND NUMBER OF POINTS IN THE SAMPLE
% OUTPUT: CONFIDENCE

function eta = MS_discriminator(m1,m2,C,N)

A = [pi sqrt(2)*pi 2*pi pi^2];
dlt = [1 1/2 3/2 2];

% SOLVING SADDLE POINT EQUATIONS FOR ALL MODELS
[fh1] = Mind_saddlepoint(m1,N,dlt(1));
[fh2] = Mind_saddlepoint(m2,N,dlt(1));
[fJ] = Mind_saddlepoint(C,N,dlt(1));
[fh] = Mind_saddlepoint((m1+m2)/2,N,dlt(2));
[phi] = M2_saddlepoint(m1,m2,C,N,dlt(3));
[psi] = M3_saddlepoint(m1,m2,C,N,dlt(4));

% MODEL WITH 0 PARAMETERS (M_1)
PM0 = -log(4);

% MODELS WITH 1 PARAMETERS (M_2, M_3, M_4, M_6)
PM1a = fh1 -log(2) + log(2*pi/(N*(1+dlt(1)/N)*A(1)^2))/(2*N);
PM1b = fh2 -log(2) + log(2*pi/(N*(1+dlt(1)/N)*A(1)^2))/(2*N);
PM1c = fJ -log(2) + log(2*pi/(N*(1+dlt(1)/N)*A(1)^2))/(2*N);
PM1d = 2*fh + log(2*pi/(N*(1+dlt(2)/N)*A(2)^2))/(2*N);

% MODELS WITH 2 INDEPENDENT PARAMETERS (M_5, M_7, M_8)
PM2a = fh1 + fh2 + log(2*pi/(N*(1+dlt(1)/N)*A(4)))/N;
PM2b = fh1 + fJ + log(2*pi/(N*(1+dlt(1)/N)*A(4)))/N;
PM2c = fh2 + fJ + log(2*pi/(N*(1+dlt(1)/N)*A(4)))/N;

% MODEL WITH 2 DEPENDENT PARAMETERS (M_9)
PM2d = phi + log(2*pi/(N*(1+dlt(3)/N)*A(3)))/N;

% MODEL WITH 3 PARAMETERS (M_10)
PM3 = psi + 3*log(2*pi/(N*(1+dlt(4)/N)*A(4)^(2/3)))/(2*N);

% VECTOR OF PROBABILITIES
logPM = [PM0 PM1a PM1b PM1d PM2a PM1c PM2b PM2c PM2d PM3];
PM = exp(N*logPM);

% EVALUATING THE CONFIDENCE
NB = PM(1) + PM(2) + PM(3) + PM(4) + PM(5);
B = PM(6) + PM(7) + PM(8) + PM(9) + PM(10);
eta = (B-NB)/(B+NB);
end

% FIND THE SADDLE POINT FOR MODELS WITH 1 AND 2 INDEPENDENT PARAMETERS
function [y] = Mind_saddlepoint(a,N,delta)
xguess = 0;
opts = optimset('Diagnostics','off','Display','off');
B = (1+delta/N);
x = fsolve(@(x) a - B*tanh(x),xguess,opts);
y = a*x - log(2*cosh(x));
end

% FIND THE SADDLE POINT FOR MODEL M_9
function [phi] = M2_saddlepoint(m1,m2,C,N,delta)
xguess = zeros(1,2);
opts = optimset('Diagnostics','off','Display','off');
B = (1+delta/N);
x = fsolve(@(x)M2system(x,m1,m2,C,N,B),xguess,opts);
h = x(1);
J = x(2);
fh = h*(m1+m2) - 2*log(2*cosh(h));
fJ = J*C - log(2*cosh(J));
sigma = (1+tanh(J)*tanh(h)^2)/2;
phi = fh+fJ-log(sigma);
end

%FIND THE SADDLE POINT FOR MODEL M_10
function [psi] = M3_saddlepoint(m1,m2,C,N,delta)
xguess = zeros(1,3);
opts = optimset('Diagnostics','off','Display','off');
B = (1+delta/N);
x = fsolve(@(x)M3system(x,m1,m2,C,B),xguess,opts);
h = [x(1) x(2)];
J = x(3);
fh1 = h(1)*m1 - log(2*cosh(h(1)));
fh2 = h(2)*m2 - log(2*cosh(h(2)));
fJ = J*C - log(2*cosh(J));
sigma = (1+tanh(h(1))*tanh(h(2))*tanh(J))/2;
psi = fh1+fh2+fJ-log(sigma);
end

function y = M2system(v,m1,m2,C,N,B)
y = [m1 + m2 - 2*B*tanh(v(1)) - 2*B*(1-tanh(v(1))^2)*tanh(v(1))*...
tanh(v(2))/(1+tanh(v(1))^2*tanh(v(2)));
    C + 1/(2*N) - B*tanh(v(2)) - B*((1-tanh(v(2))^2)*tanh(v(1))^2)/...
(1+tanh(v(1))^2*tanh(v(2)))];
end

function y = M3system(v,m1,m2,C,B)
y = [m1 - B*tanh(v(1)) - B*((1-(tanh(v(1)))^2)*tanh(v(2))*tanh(v(3)))/...
(1+tanh(v(1))*tanh(v(2))*tanh(v(3)));
    m2 - B*tanh(v(2)) - B*((1-(tanh(v(2)))^2)*tanh(v(1))*tanh(v(3)))/...
(1+tanh(v(1))*tanh(v(2))*tanh(v(3)));
    C - B*tanh(v(3)) - B*((1-(tanh(v(3)))^2)*tanh(v(1))*tanh(v(2)))/...
(1+tanh(v(1))*tanh(v(2))*tanh(v(3)))];
end

\end{lstlisting}

\end{document}